\PassOptionsToPackage{unicode}{hyperref}
\PassOptionsToPackage{hyphens}{url}
\PassOptionsToPackage{dvipsnames,svgnames,x11names}{xcolor}
\documentclass[
  10pt,
]{article}
\usepackage{xcolor}
\usepackage{amsmath,amssymb}
\usepackage{iftex}
\ifPDFTeX
  \usepackage[T1]{fontenc}
  \usepackage[utf8]{inputenc}
  \usepackage{textcomp} 
\else 
  \usepackage{unicode-math} 
  \defaultfontfeatures{Scale=MatchLowercase}
  \defaultfontfeatures[\rmfamily]{Ligatures=TeX,Scale=1}
\fi
\usepackage{lmodern}
\ifPDFTeX\else
\fi
\IfFileExists{upquote.sty}{\usepackage{upquote}}{}
\IfFileExists{microtype.sty}{
  \usepackage[]{microtype}
  \UseMicrotypeSet[protrusion]{basicmath} 
}{}
\makeatletter
\@ifundefined{KOMAClassName}{
  \IfFileExists{parskip.sty}{%
    \usepackage{parskip}
  }{
    \setlength{\parindent}{0pt}
    \setlength{\parskip}{6pt plus 2pt minus 1pt}}
}{
  \KOMAoptions{parskip=half}}
\makeatother
\providecommand{\tightlist}{%
  \setlength{\itemsep}{0pt}\setlength{\parskip}{0pt}}

\usepackage{newunicodechar}
\newunicodechar{−}{\ensuremath{-}}
\newunicodechar{×}{\ensuremath{\times}}
\newunicodechar{→}{\ensuremath{\rightarrow}}
\newunicodechar{±}{\ensuremath{\pm}}
\newunicodechar{·}{\ensuremath{\cdot}}
\newunicodechar{‖}{\ensuremath{\|}}
\newunicodechar{≈}{\ensuremath{\approx}}
\newunicodechar{≥}{\ensuremath{\geq}}
\newunicodechar{≤}{\ensuremath{\leq}}
\newunicodechar{√}{\ensuremath{\surd}}
\newunicodechar{ρ}{\ensuremath{\rho}}
\newunicodechar{κ}{\ensuremath{\kappa}}
\newunicodechar{μ}{\ensuremath{\mu}}
\newunicodechar{η}{\ensuremath{\eta}}
\newunicodechar{λ}{\ensuremath{\lambda}}
\newunicodechar{⁻}{\textsuperscript{\ensuremath{-}}}
\newunicodechar{¹}{\textsuperscript{1}}
\newunicodechar{²}{\textsuperscript{2}}
\newunicodechar{³}{\textsuperscript{3}}
\newunicodechar{⁴}{\textsuperscript{4}}
\newunicodechar{⁵}{\textsuperscript{5}}
\newunicodechar{⁶}{\textsuperscript{6}}
\newunicodechar{⁷}{\textsuperscript{7}}
\newunicodechar{⁸}{\textsuperscript{8}}
\newunicodechar{⁹}{\textsuperscript{9}}
\newunicodechar{⁰}{\textsuperscript{0}}
\makeatletter
\newcommand*{\section@prefix}{}
\newcommand*{\setappendixprefix}{\renewcommand*{\section@prefix}{Appendix~}}
\renewcommand*{\@seccntformat}[1]{\csname #1@prefix\endcsname\csname the#1\endcsname\quad}
\makeatother
\usepackage{caption}
\usepackage[margin=2.5cm]{geometry}
\usepackage{microtype}
\usepackage{array,booktabs,longtable,graphicx}
\usepackage{needspace}
\usepackage[section]{placeins}
\definecolor{linkblue}{rgb}{0.1,0.2,0.6}
\newcolumntype{P}[1]{>{\raggedright\arraybackslash}p{#1}}

\AtBeginDocument{%
  \bookmarksetup{numbered, open, openlevel=1}%
  \hypersetup{linkcolor=linkblue, citecolor=linkblue, urlcolor=linkblue,
    pdfsubject={Embedding maps, LLM ideation and forecasting}}%
}
\usepackage{bookmark}
\IfFileExists{xurl.sty}{\usepackage{xurl}}{} 
\hypersetup{
  pdftitle={Do Embedding-Map Distances Guide LLM Ideation? Four Tests Against What Came Next},
  pdfauthor={Javier Aguilar Martín},
  pdfkeywords={idea generation, large
language models, embedding
spaces, novelty, forecasting, pre-registration},
  colorlinks=true,
  linkcolor={blue},
  filecolor={Maroon},
  citecolor={Blue},
  urlcolor={blue},
  pdfcreator={LaTeX via pandoc}}

\title{Do Embedding-Map Distances Guide
LLM Ideation? Four Tests Against What
Came Next}
\author{Javier Aguilar Martín\\
AGILabs (javieraguilar.ai)\\
\href{mailto:javiecija96@gmail.com}{\nolinkurl{javiecija96@gmail.com}}}
\date{November 2026}

\begin{document}
\maketitle
\begin{abstract}
A popular ideation recipe treats an
embedding map of existing work as a
guide: pick points far from everything
and decode them into ideas. I test four
ways such distances could guide ideation
against what appeared later. (1) Far
random directions are not text: no later
item reaches a raw cosine above about
0.10 with one (after centring or
whitening, 22 of 24 cells stay out of
reach), and an LLM decoder shown a
target's neighbours, with score-only
feedback, writes those neighbours in all
six target sets. A neighbour-free search
writes coherent text at median cosine
0.57 and 0.47 to interior gaps, beating
the nearest existing item on 15 and 14
of 20, but only 0.07 and 0.06 to far
points. (2) On k-means cells of a
past-only map, the registered hole score
mostly tracks cell size and position;
three density-based gap scores forecast
which cells grow fastest at or below
chance (AUC 0.30 to 0.66) and add
nothing to momentum in Steam and arXiv,
with patents inconclusive; more isolated
new Steam games are slightly better
reviewed. (3) LLM continuations of arXiv
papers stay closer to the source than
papers that cite it, under three user
prompts and against proposal rewrites
(gpt-4o) and April-December 2023 papers
(OLMo). Critique lengthens this step
beyond content-free revision, replicated
on 150 new papers. Revealing a real
follow-up's title turns the ideas by
0.032 along a follow-up-specific
direction beyond revision; reviewer,
oracle and distance feedback add 0.13,
0.08 and 0.34 of that turn (upper bounds
0.46, 0.46, 0.70), and their turn is not
separable from revision or their longer
step. (4) For inspiration material
graded by distance, no inverted U was
detected, on an anticipation margin that
tracks format more than content.
Original tests were registered before
running, follow-ups (thirteen requested
in two AI-agent reviews) after the main
results; all are dated only by local git
commits.

\textbf{Keywords:} idea generation,
large language models, embedding spaces,
novelty, forecasting, pre-registration
\end{abstract}

\section{Introduction}\label{sec:1}

Several lines of work treat distances in
an embedding map of existing work as a
guide to new work. EAGLE (Tennenholtz et
al.,
\hyperlink{ref-Tennenholtz2024b}{2024b})
steers a language model towards
``content gaps'' in a movie embedding;
SciMON (Wang et al.,
\hyperlink{ref-Wang2024}{2024}) sends
ideas back when they are too similar to
the literature; benchmarks score ideas
by their distance to what exists, and
Tang and Yang
(\hyperlink{ref-Tang2026}{2026}) measure
how far research agents step from their
seed papers. A widely circulated post on
X put the premise in its plainest form,
in reply to the remark that AI cannot
create a brand new idea (SyntheticBeef,
\hyperlink{ref-SyntheticBeef2026}{2026}):
``Take all 100,000+ games on steam,
scrape their descriptions and metadata,
semantically plot them in vector space
using embeddings, mathematically choose
a location in vector space far away from
neighbors, convert that back to a
description and metadata,
boom,''\footnote{The first 278
  characters of a longer post, as served
  by X's syndication endpoint on 15
  November 2026; the visible text does
  not say how a point is converted back.}
I call this \textbf{gap-based ideation}.
Its premise is that distances in an
embedding map of existing work carry
information about where good new work
lies, and that a model can be guided by
them.

The premise splits into three testable
claims about the map, plus a fourth
about the material a model is inspired
by. This paper tests each against ground
truth that a past-only map cannot see:
the items released or published after a
frozen date.

\begin{enumerate}
\def\labelenumi{\arabic{enumi}.}
\tightlist
\item
  \textbf{Empty points are ideas}
  (\hyperref[sec:3]{Section~3}). I find
  that far random directions, the
  recipe's literal targets, are not
  text: no later item and no coherent
  text I could construct comes
  meaningfully close to one (in the raw
  space, cosine about 0.10 at most for
  later items and 0.14 for coherent
  text), and after centring or whitening
  they stay out of reach in 22 of 24
  corrected-space cells, which
  random-direction geometry predicts. An
  LLM decoder shown a target's nearest
  items, with score-only feedback,
  writes those items, causally. Interior
  gaps of the map behave differently in
  absolute terms: a search without
  neighbours writes coherent text at
  median cosine 0.57 (Steam) and 0.47
  (arXiv) to them, beating the nearest
  existing item on 15 and 14 of 20
  targets by about 0.04 at the median.
  Against far points the same search
  also beats their near-zero ceiling on
  16 and 11 of 20, but its best coherent
  text stays near-orthogonal (median
  cosine 0.07 and 0.06). The claims
  cover the decoder and searches I ran
  and these two kinds of target; they do
  not cover embedding inversion, which I
  could not test informatively.
\item
  \textbf{Gaps forecast where the future
  grows} (\hyperref[sec:4]{Section~4},
  pre-registered). The test is on cells
  of a k-means partition of the past
  map, so it asks whether sparse cells
  grow faster, not whether empty regions
  fill. I find that the registered hole
  score mostly reflects the size and
  position of a cell, and that three gap
  scores that do detect planted holes
  forecast which cells grow fastest in
  volume at or below chance and add
  nothing to momentum, the naive
  baseline, in Steam and arXiv; in
  patents the result is inconclusive.
  The registered conclusions also pass
  in two spaces whose encoder was not
  trained on post-split text, a
  TF-IDF/LSA space fitted on past text
  only and, for Steam and arXiv, an
  encoder released before the splits,
  although with LSA the Steam result
  becomes inconclusive. The registered
  item-level test, whether saved gap
  targets are approached faster than
  density-matched points, could not be
  run as designed and is inconclusive.
  More isolated new Steam games are
  slightly better reviewed, a failed
  prediction in the direction of the
  premise.
\item
  \textbf{Distances can steer a model
  continuing a paper}
  (\hyperref[sec:5]{Section~5},
  pre-registered with registered
  follow-ups). Real research starts from
  a given paper. I find that LLM
  continuations take a shorter step than
  the papers that later cited the
  source, as Tang and Yang
  (\hyperlink{ref-Tang2026}{2026})
  showed at scale, under three wordings
  of the user prompt, also after those
  papers are rewritten as proposals by
  the same model with or without the
  source in view, and, for OLMo, also
  against citing papers from 2023, just
  after its training data end; and that
  feedback lengthens the step beyond
  what revising alone does, in a
  replication on new papers. Whether
  feedback also orients the ideas is not
  resolved beyond revision: revealing
  one real follow-up's title turns
  gpt-4o's ideas towards the citing
  papers, reviewer, oracle and distance
  feedback add 0.13, 0.08 and 0.34 of
  that turn beyond revision (upper
  bounds 0.46, 0.46 and 0.70), and the
  critique arms' rise in alignment
  cannot be separated from their longer
  step.
\item
  \textbf{The distance of inspiration
  matters} (\hyperref[sec:6]{Section~6},
  pre-registered). Ideas often come from
  material that looks irrelevant. The
  result here is methodological: the
  anticipation margin I registered to
  score such ideas against the games
  released after the model's cutoff
  fails its validation, separating real
  future games from ideas mainly by
  format, so the null on distance (no
  middle optimum detected, a small
  positive slope in one of two
  embedders) is uninformative.
\end{enumerate}

What the four parts share is narrower
than a single mechanism. Momentum, a
statement about where recent items
already are, was the strongest
forecaster of cell-level growth I found,
as earlier work on concept networks
leads one to expect. What a decoder
writes follows the items it is shown. In
the gpt-4o discovery sample,
content-free revision produces about
half or more of the step lengthening
that critique produces and, in Qwen3,
turns the ideas along the
follow-up-specific direction no less
than any critique arm; in the
replication, in Qwen3, it turns them
less (+0.008 against +0.012 to +0.019
along that direction); the BGE-M3
comparisons differ in both samples; and
for gpt-5.4-mini it produces a third to
two fifths of the lengthening. And the
one intervention that moved ideas
towards what came next, revealing a real
later paper, moved them only part of the
way.

\hyperref[tab:1]{Table~1} lists each
claim tested, what I found and what kind
of registration the finding rests on.

\begin{table}[tbp]
\centering\footnotesize\setlength{\tabcolsep}{4pt}
\caption[]{\textbf{Table 1.} Results at a glance. Basis: \emph{original} = original registration, committed before its data; \emph{follow-up} = registered follow-up, written after the main results were known; \emph{exploratory} = not registered.}\label{tab:1}
\begin{tabular}{@{}P{\dimexpr 0.240\linewidth-2\tabcolsep\relax}P{\dimexpr 0.360\linewidth-2\tabcolsep\relax}P{\dimexpr 0.160\linewidth-2\tabcolsep\relax}P{\dimexpr 0.240\linewidth-2\tabcolsep\relax}@{}}
\toprule
Claim (section) & Finding & Verdict & Basis \\
\midrule
Far random directions can be decoded into ideas (3) & no later item reaches a raw cosine above about 0.10 with one; 22 of 24 corrected-space cells stay out of reach & not supported & exploratory (raw future check); follow-up (centred and whitened) \\
A decoder guided by distance follows the target (3) & shown a target's neighbours with score-only feedback, it writes those neighbours in 6 of 6 target sets & not supported for this decoder & original; follow-up (independent target sets) \\
Interior gaps can be written as text (3) & median cosine 0.57 and 0.47; above the nearest existing item on 15 and 14 of 20 targets & supported, by small margins & original (search); exploratory (exact ceiling counts) \\
Sparse cells of the past map grow faster (4) & valid gap scores forecast top-decile growth at or below chance (AUC 0.30 to 0.66); nothing added to momentum in Steam and arXiv; registered criteria also met in two spaces without post-split encoder training (Steam inconclusive in LSA) & not supported; patents inconclusive & original (F2, F3); follow-up (score validity, leak-free spaces) \\
Empty regions fill (4) & new items are slightly more isolated than held-out past ones (AUC 0.61 to 0.64 arXiv, 0.55 to 0.60 Steam, 0.51 patents); preferential filling of the past map's empty points not shown (Steam and arXiv inconclusive, patents contradicting) & not shown by the registered contrast (a graded association in arXiv); new arXiv items more isolated & original (F6, inconclusive); follow-up (item-level density) \\
Isolated new items are more successful (4) & more isolated new Steam games are slightly better reviewed (ρ +0.11 to +0.13), also net of review count, year, reach and genre; more isolated patent applications are slightly less often granted & slightly, in Steam (registered F5 fails); reversed in patents & original (F5); follow-up (controls) \\
LLM continuations step as far as later work (5.1) & shorter step than citing papers (0.149), also with two other user-prompt wordings (0.135, 0.130), against papers rewritten as proposals (0.112 with the context shown, 0.232 without), true continuations (0.048) and, for OLMo, citing papers from 2023 (0.215) & not supported & original; follow-up (wording, rewrites, continuations, 2023 papers) \\
Feedback lengthens the step (5.2) & beyond content-free revision, replicated on 150 new papers & supported & original (critique; revision control committed with its results); follow-up (replication) \\
Feedback turns ideas towards later work (5.3) & revealing a follow-up's title turns them by 0.032; feedback adds 0.08 to 0.34 of that, not separable from revision or from the longer step & not resolved & follow-up \\
Inspiration distance has an optimum (6) & no inverted U detected; the margin fails its validation (it tracks format more than content) & uninformative & original (P1 to P5); follow-up (validation) \\
\bottomrule
\end{tabular}
\end{table}

\section{Setup}\label{sec:2}

\textbf{Corpora and maps.} Steam games
(Bustos,
\hyperlink{ref-Bustos2026}{2026};
released before 2024: 68,222 items after
removing adult-tagged rows; from 2024:
40,989), arXiv cs titles and abstracts
(Cornell University,
\hyperlink{ref-Cornell2026}{2026}; a
seeded sample of 72,000 first versions
before 2023 and 41,000 from 2023-01-01
to 2026-10-15) and US patent
applications (the Harvard USPTO Patent
Dataset, HUPD; Suzgun et al.,
\hyperlink{ref-Suzgun2023}{2023}; filed
2012--2015: 71,521; 2016--2017: 40,342).
Every map is built from past-split items
only; the future split is the ground
truth. Two routes of leakage remain and
I do not close them: the embedders were
trained on text that may postdate the
splits, and an arXiv record's abstract
is its latest version, which can be
later than the first-version date that
assigns it to a split
(\hyperref[sec:8]{Section~8}).
\hyperref[app:A.1]{Appendix~A.1} gives
the details
(\hyperref[tab:A1]{Table~A1}).

\textbf{Embedders and shared measures.}
Qwen3-Embedding-0.6B (Zhang et al.,
\hyperlink{ref-Zhang2025}{2025};
\(d = 1024\), L2-normalised) is the
pre-registered space, and BGE-M3 (Chen
et al., \hyperlink{ref-Chen2024}{2024})
is the replication space. Not every
result was replicated, and not every
replication succeeded; each section says
which (\hyperref[tab:A13]{Table~A13} in
\hyperref[app:C.1]{Appendix~C.1}
summarises). All similarities are
cosines. An item's \textbf{isolation} is
1 − its maximum cosine to the past
corpus. Its \textbf{latitude} is its
cosine with the normalised mean of the
past corpus, its position on the cap
that texts occupy in the raw space
(\hyperref[sec:3]{Section~3}).

\textbf{Generators.}
\hyperref[tab:2]{Table~2} lists every
model and where it is used. The
continuation and inspiration studies use
generators whose training predates their
ground truth; the gaps study uses 2026
models, which postdate its splits, and
reads them only in ways that do not
depend on the future
(\hyperref[sec:8]{Section~8}).

\begin{table}[tbp]
\centering\footnotesize\setlength{\tabcolsep}{4pt}
\caption[]{\textbf{Table 2.} Models. Cutoffs are the providers' published training cutoffs unless stated.}\label{tab:2}
\begin{tabular}{@{}P{\dimexpr 0.270\linewidth-2\tabcolsep\relax}P{\dimexpr 0.270\linewidth-2\tabcolsep\relax}P{\dimexpr 0.300\linewidth-2\tabcolsep\relax}P{\dimexpr 0.160\linewidth-2\tabcolsep\relax}@{}}
\toprule
Model & Role & Training cutoff & Sections \\
\midrule
gpt-4o-2024-08-06 (OpenAI, \hyperlink{ref-OpenAI2024}{2024}), temperature 1 & generator & October 2023 & \hyperref[sec:5]{5}, \hyperref[sec:6]{6}, \hyperref[app:A.7]{\hyperref[app:A.7]{A.7}} \\
OLMo-7B-0424-Instruct (Groeneveld et al., \hyperlink{ref-Groeneveld2024}{2024}), temperature 1 & generator & February--March 2023 (Dolma; Soldaini et al., \hyperlink{ref-Soldaini2024}{2024}) & \hyperref[sec:5]{5} \\
gpt-5.4-mini~(2026-03-17), temperature 1 & generator & 31 August 2025 (third-party listings; none published) & \hyperref[sec:5.4]{5.4} \\
gpt-6-luna & decoder and generator & after all splits & \hyperref[sec:3]{3}, \hyperref[app:A.1]{\hyperref[app:A.1]{A.1}}--\hyperref[app:A.4]{\hyperref[app:A.4]{A.4}} \\
Claude Sonnet 5.5 & decoder and generator & after all splits & \hyperref[sec:3]{3}, \hyperref[app:A.1]{\hyperref[app:A.1]{A.1}}--\hyperref[app:A.4]{\hyperref[app:A.4]{A.4}} \\
gpt-6.1-sol, Claude Opus 5.5, Claude Haiku 5.5 & stress-test generators, coherence and novelty judges, follow-up labellers & after all splits & \hyperref[sec:6]{6}, \hyperref[sec:8]{8}, \hyperref[app:A.1]{\hyperref[app:A.1]{A.1}}, \hyperref[app:A.7]{\hyperref[app:A.7]{A.7}} \\
\bottomrule
\end{tabular}
\end{table}

\textbf{Pre-registration and reporting.}
Two kinds of registration appear in this
paper, and I name them differently. The
\textbf{original registrations}
(\hyperref[sec:3]{Sections~3} to
\hyperref[sec:6]{6}) wrote down each
confirmatory test, its predictions and
its analysis rules in a document
committed to the project's git
repository before the data it is tested
on were generated or inspected (the one
exception, the neutral control of
\hyperref[sec:5.2]{Section~5.2}, was by
my record written before its first call
but committed with its results).
\textbf{Registered follow-ups} were
written after the main results were
known: fifteen strengthening studies,
and thirteen studies requested in two
reviews of the preprint by AI agents
(the ``Use of AI tools'' section); each
was committed before its own data were
generated or analysed, and the paper
calls them registered follow-ups, not
pre-registrations in the original sense.
Commit timestamps are set locally and
the public repository is a later
anonymised copy, so nothing dates the
registrations independently of the
repository; \hyperref[app:B]{Appendix~B}
lists every document with its commit.
Earlier work had used the same corpora
and splits: the decoding runs of
\hyperref[sec:3]{Section~3} never touch
the future split, and the memory probe
of \hyperref[app:A.7]{Appendix~A.7} asks
only whether generated items land nearer
later items. No analysis before its
pre-registration looked at where future
items sit on the past map or how its
cells grow, and the continuation and
inspiration data were collected after
theirs. Deviations are stated where they
occur. Holm correction is applied within
each registered family (a study, an
embedder and, where registered, a set of
predictions); no correction is applied
across families, so the headline claims
rest on effects that replicate across
samples or embedders. Analyses that were
not registered are marked as exploratory
where they appear, and their
\emph{p}-values are uncorrected. A
failed prediction is reported as failed.

\section{Far random directions are not text, and a neighbour-prompted decoder writes its neighbours}\label{sec:3}

The recipe's decoder shows an LLM the
eight nearest past items of a target
point, asks for a new item, embeds it
and reports back the cosine reached and
the best so far (score-only
``hot-and-cold'' feedback), over four
rounds of three candidates. If the map's
empty points were ideas, the text should
be driven by the target. Three results
bear on this.

\begin{figure}[!htb]
\centering
\includegraphics[width=\linewidth]{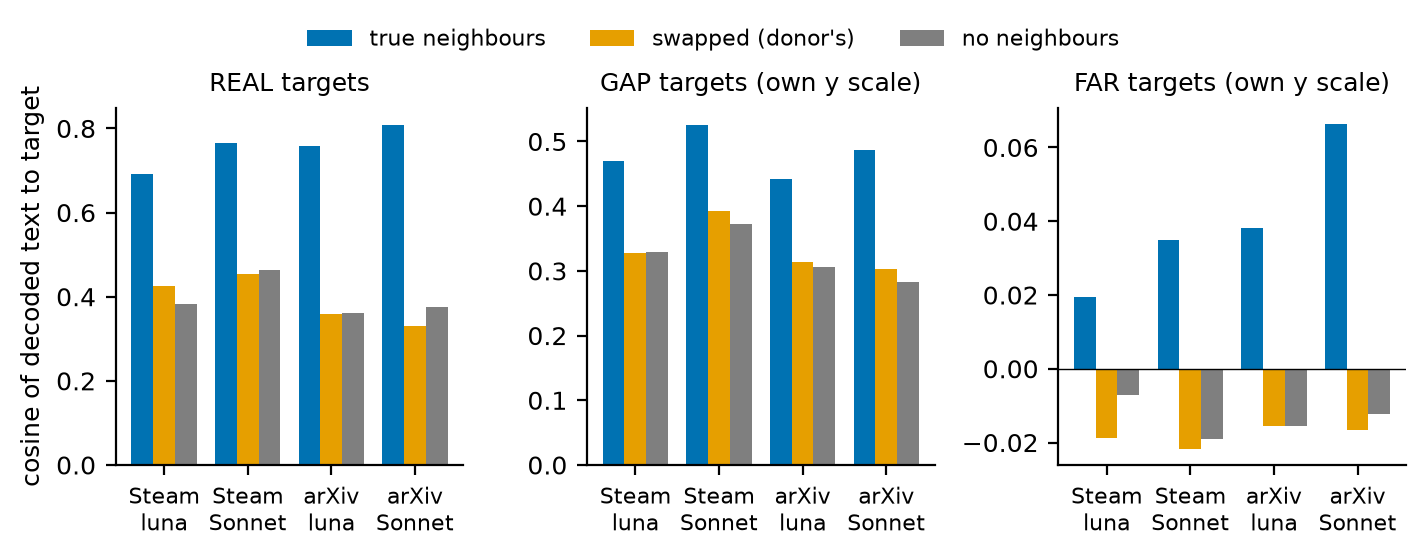}
\caption[]{\textbf{Figure 1.} Causal anchoring. Mean cosine of the decoded text to its target when the decoder is shown the target's true neighbours, another target's neighbours (swapped) or none, with the target and the score feedback unchanged; 20 targets per bar (luna = gpt-6-luna, Sonnet = Claude Sonnet 5.5). Swapping the neighbours lowers the landing on real and gap targets by 0.13 to 0.48 and removes it for far targets. The three panels have different y-axis scales (up to 0.8, 0.5 and 0.06).}\label{fig:1}
\end{figure}

\textbf{What the decoder writes follows
the neighbours it is shown.} I
pre-registered a randomised intervention
on 20 saved targets per arm in Steam and
arXiv, with two decoders (gpt-6-luna and
Claude Sonnet 5.5): REAL (hidden
held-out items, the positive control),
GAP (interior holes, the most isolated
draws from the corpus Gaussian) and FAR
(the recipe's far points, the most
isolated uniform random directions). The
target and the score-feedback procedure
were held fixed; only the shown
neighbours changed, between the true
ones, another target's (SWAP) and none.
Counting targets as the units
(registered follow-up), the 12 decoder
cells are 6 independent sets of 20
targets each, shared by the two
decoders: in all 6 the decoded text
moves towards the swapped-in neighbours
and away from the true ones (Holm
\emph{p} = 0.0006 in each set, the floor
of the resampling test: none of 10,000
resamples had a nonpositive effect,
times six sets;
\hyperref[fig:1]{Figure~1}, Table A4).
Pooled over decoders, swapping lowers
the landing on the target by 0.29 and
0.44 for REAL targets (Steam, arXiv),
0.14 and 0.16 for GAP targets, and 0.047
(CI 0.039 to 0.055) and 0.068 (CI 0.059
to 0.078) for FAR targets. The two
decoders largely agree target by target
for REAL and GAP (Spearman 0.67 to
0.96), so those cells are close to one
confirmation each, not two. The
attraction to the shown neighbours also
holds when the decoded texts are
re-embedded in BGE-M3 and in every
centred or whitened version of both
embedders. The one-coefficient law that
the pre-registration froze for its size
holds in 5 of 12 cells only
(\hyperref[app:A.3]{Appendix~A.3}).

This test shows that the decoder reads
its prompt; it cannot show that a target
is undecodable, because the neighbours
are the channel through which the target
reaches the decoder. What remains is the
score. Without neighbours, from round 0
(no feedback yet) to round 3 the text
moves towards its own FAR target by
−0.003 (CI −0.010 to +0.006; Steam) and
+0.004 (CI −0.004 to +0.013; arXiv) more
than towards the other targets of its
arm, so the score alone does not
detectably steer towards a far point
(bound 0.02, registered follow-up); the
final margin of about 0.035 over other
targets comes from choosing the best of
12 candidates, of which the round-0
choice among three already gives 0.015
and 0.013. For REAL and GAP targets the
score does steer: the text moves towards
its own target by +0.024 to +0.027 more
than towards the others in three of the
four sets (Holm), and by +0.013 (not
significant) in the fourth
(\hyperref[app:A.4]{Appendix~A.4}).

\textbf{Far points are out of reach.} In
the raw space texts occupy a narrow cap
of the sphere (two Steam games have
median cosine 0.41; two random
directions 0 ± 0.031), so a far
direction is far from every text, not
only from existing ones. From corpus
statistics alone I wrote down the median
corpus ceiling of the arXiv FAR targets
before building them: 0.062; it was
0.058
(\hyperref[app:A.3]{Appendix~A.3}). That
prediction concerns the past corpus.
Whether later items come closer I
checked only afterwards (exploratory):
across the 50 saved FAR targets of each
corpus, the best cosine of any later
item exceeds the past ceiling on 58 \%
(Steam, patents) to 86 \% (arXiv) of
targets, but stays at 0.046 to 0.066 at
the median and about 0.10 at most,
against a past ceiling of 0.046 to
0.062. That the future stays far is what
the geometry of random directions
predicts, so it is a check of that
geometry rather than a finding about the
future: text from other domains never
enters the far region of these maps
either (\hyperref[sec:4]{Section~4}).
The cap is a property of the uncentred
space, so I checked whether the result
survives removing it (registered
follow-up). After mean-centring,
all-but-the-top (Mu and Viswanath,
\hyperlink{ref-Mu2018}{2018}),
BERT-whitening (Su et al.,
\hyperlink{ref-Su2021}{2021}) and a
shrunk whitening of both embedders, the
median pairwise cosine of texts falls
from 0.25--0.59 to −0.012 to −0.001, but
far targets rebuilt in each space stay
farther from every past text than 99 \%
of real items are from their nearest
neighbour in all 30 corpus × embedder ×
transform cells, and from every future
text in 28 (the exceptions:
BERT-whitened BGE-M3, where one of 50
Steam and two of 50 arXiv far targets
have a future item inside that range,
while Qwen3 does not). In the corrected
spaces the explanation is no longer a
cap: texts are spread out, and a far
target is simply the most isolated of
20,000 random directions, whose best
cosine to \(N\) items in \(d'\)
dimensions stays near the maximum of
\(N\) random cosines
(\hyperref[app:A.3]{Appendix~A.3}). The
small landing of decoded text on FAR
targets with the true neighbours (cosine
0.02 to 0.07, gone with swapped or no
neighbours) is a raw-cosine result: in
the corrected spaces it holds in 1 or 2
of the 4 FAR cells.

\textbf{Interior gaps can be decoded.}
Searching for coherent text directly,
without neighbours, separates the two
kinds of target. A seed-and-rewrite
search, gated by a positive control that
passed on both corpora, has an LLM
rewrite the best seeds of a gradient
search into coherent items, with the
cosine as its only clue, and two models
must accept the result as coherent
(\hyperref[tab:A5]{Table~A5}). For GAP
targets it finds coherent text beyond
the 99th percentile of corpus items on
19 of 20 targets in both corpora (the
registered count), and closer to the
target than the nearest existing item,
the corpus ceiling, on 15 of 20 (Steam)
and 14 of 20 (arXiv; one target got no
accepted text), a post-hoc recount. The
median best text reaches 0.567 against a
median ceiling of 0.533 on Steam and
0.472 against 0.428 on arXiv, a gain of
about +0.035 and +0.042 at the median,
with the target at cosine 1. For FAR
targets the registered verdict uses
ceiling*, the larger of the visible
corpus ceiling and the ceiling of a
held-out slice
(\hyperref[app:A.4]{Appendix~A.4}). The
search reaches coherent text at least
0.05 above ceiling* on 5 of 20 Steam
targets, which refutes the
pre-registered strong form of the
objection on Steam (25 \% against a 20
\% threshold; 15 \% if a third,
independent model must also accept the
text), and on none of 20 arXiv targets
(one against the visible ceiling alone).
FAR targets thus also beat their visible
ceiling often (16 and 11 of 20, median
gain +0.030 and +0.007), but that
ceiling is near zero: the best coherent
text reaches a median cosine of 0.069
(Steam) and 0.057 (arXiv) to the target,
at most 0.138 and 0.109. The difference
between the two kinds of target is one
of absolute distance, not of beating the
corpus: coherent text reaches a cosine
of about 0.5 with an interior gap and
stays near-orthogonal to a far point. No
search I ran wrote coherent text
meaningfully close to a far point, and
no finite set of coherence labels could
certify that none exists
(\hyperref[app:A.5]{Appendix~A.5}).

\textbf{What the claim does not cover.}
Embedding inversion is the natural
decoder for ``convert that point back
into a description''. No inverter exists
for the two encoders used here; a
registered attempt with a trained
vec2text inverter for gtr-t5-base, the
closest supported encoder, with targets
rebuilt in that space, was uninformative
by its own gate: the inverter recovers
hidden real items (median cosine 0.97),
but the two-judge coherence gate
accepted only 12 \% (Steam) and 28 \%
(arXiv) of real 32-token fragments, so
no inversion could be judged coherent or
not (\hyperref[app:A.4]{Appendix~A.4}).
The claims of this section are about the
neighbour-prompted decoder and the
searches I ran, not about inversion.

Asking a generator for new items while
showing it the titles it had already
proposed, with no map at all, was judged
more novel by LLM judges than every
map-guided arm in all three corpora,
though my own blind ratings agree with
the judges only weakly
(\hyperref[app:A.7]{Appendix~A.7}).

\section{Gap scores do not forecast which cells grow}\label{sec:4}

\textbf{Design (pre-registered).} I
cluster each past map into 256 cells
(k-means, refitted in each embedder, a
deviation from the pre-registration) and
score each cell by \textbf{growth} (the
future share over the past share),
\textbf{hole} (the share of 200,000
draws from the corpus Gaussian that fall
in the cell over its share of past
items) and \textbf{momentum} (the share
of the cell's past items from the last
two past years). The forecasting target
is whether a cell is in the top decile
of growth. The claim tested is therefore
about where the volume of new work
grows, at the level of map cells, not
about single items. Each future item is
also checked for whether it lands in the
\textbf{far region} (isolation ≥ 0.85,
that is, maximum cosine to the past ≤
0.15, where the recipe's far targets
sit). The 256 cells, the top decile and
the two-year window were fixed in the
registration before any look at growth.
\hyperref[tab:A2]{Table~A2} has every
prediction (F1--F6) in every cell.

\textbf{The registered hole score
measures cell geometry more than
emptiness} (registered follow-up). Draws
from a full-covariance Gaussian in
\(d = 1024\) lie far from every real
item, so which centroid they fall
nearest to may say more about the
centroid than about the cell's
emptiness. It does. Across cells, the
registered hole score correlates with
the cosine of the cell's centroid to the
corpus mean at +0.74 to +0.80 and with
the cell's size at +0.34 to +0.53;
position and size together explain 59 \%
to 72 \% of its rank variance, and net
of them it is \emph{higher} in denser
cells (partial ρ with mean density +0.19
to +0.37). It does not see holes planted
by removing a ball of items (AUC 0.45 to
0.59, every interval including 0.5), at
128, 256 or 512 cells and in raw,
centred or whitened space. Some of these
correlations (with size and latitude)
were seen while checking feasibility,
before the thresholds were registered.
The pre-registered tests below therefore
test a weak operationalisation of
``gap''. Three other scores do measure
emptiness: the kNN density at the cell's
centroid, the mean sparseness of its
items, and the share of probe points in
the cell that are not near any item. The
registered rule calls a score valid if
it detects planted holes or tracks
emptiness without being dominated by
position and size. All three detect
planted holes (AUC 0.74 to 0.99, in 5 or
6 of 6 corpus × embedder cells), so with
128, 256 and 512 cells and in centred or
whitened space they give 15 valid
configurations; 13 of them also track
emptiness without geometric domination,
while centroid density passes on the
planted-hole branch alone with 128 cells
(where it is dominated by geometry) and
with 256 cells
(\hyperref[app:A.2]{Appendix~A.2},
\hyperref[tab:A3]{Table~A3}). The ranges
below are the same on the 13.

\begin{figure}[!htb]
\centering
\includegraphics[width=0.72\linewidth]{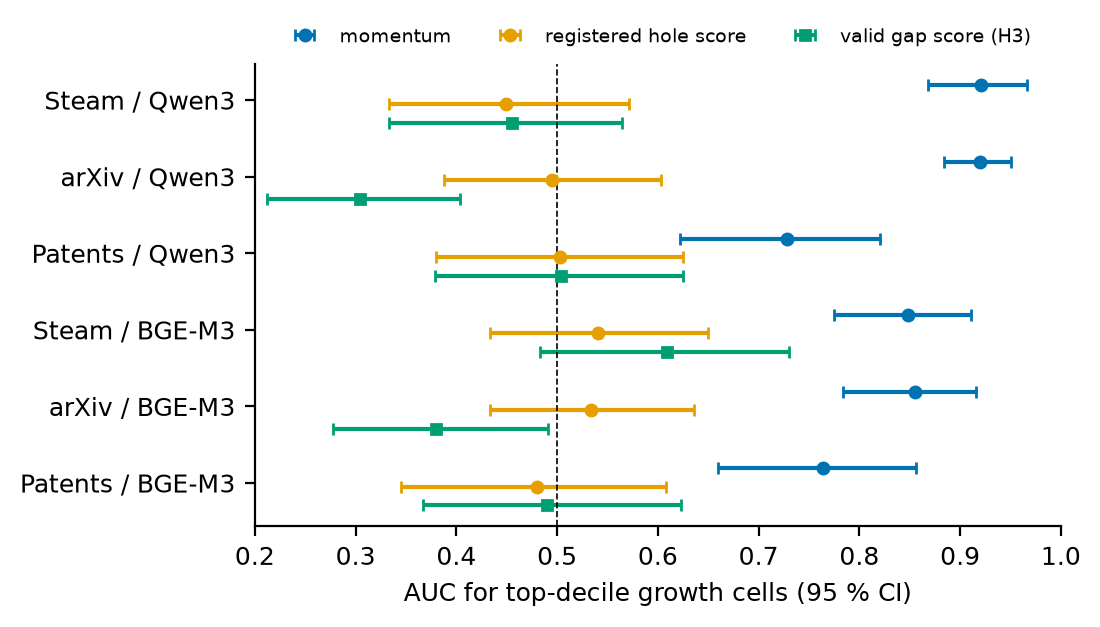}
\caption[]{\textbf{Figure 2.} Forecasting which past-map cells grow fastest. AUC for the top decile of growth, scored directly from each cell's momentum, its registered hole score, or one valid gap score (H3, the share of uncovered probe points; raw space, from the validity study's refit of the 256 cells), with bootstrap 95 \% intervals; 26 positives per row. The registered hole score is shown in \hyperref[sec:4]{Section~4} to track cell position and size more than emptiness; the valid score, which detects planted holes, gives the same picture, and so do the other valid configurations (\hyperref[tab:A3]{Table~A3}). The dashed line is chance.}\label{fig:2}
\end{figure}

\textbf{Valid gap scores forecast at or
below chance; momentum does not.} Scored
directly, the valid configurations give
AUCs of 0.30 to 0.66 for top-decile
growth, none with an interval above 0.5
in both embedders. In arXiv four of them
are \emph{below} chance in both
embedders (for example 0.33 and 0.37 for
centroid density with 256 cells), and at
a 12-month horizon all three are (0.27
to 0.32): there, emptier cells are less
likely to be the fastest-growing ones.
This direction was not a registered
outcome. The registered score itself
gives 0.45 to 0.54
(\hyperref[fig:2]{Figure~2}). Momentum,
scored the same way, reaches 0.73 to
0.92, and its strength is
corpus-specific: at a common 12-month
horizon it is 0.90 and 0.92 in Steam,
0.96 and 0.92 in arXiv and 0.75 and 0.72
in patents (Qwen3, BGE-M3), whose
interval does not overlap arXiv's. These
AUCs are computed on each score
directly, an exploratory choice, because
a cross-validated single-feature model
with 26 positives can fall below 0.5 by
the fold procedure alone.

\textbf{No gap score adds to momentum in
Steam and arXiv; patents are
inconclusive.} The pre-registered test
(F3) adds hole and latitude to momentum
in a cross-validated logistic model. In
Steam and arXiv the gain is −0.016 to
+0.006, and with a cluster bootstrap
that keeps the copies of a resampled
cell in one fold the intervals end at
+0.008 to +0.033
(\hyperref[app:A.2]{Appendix~A.2}):
gains above the registered margin of
0.03 are ruled out in three of the four
cells (arXiv/BGE-M3 reaches +0.033). In
patents the BGE-M3 gain is +0.031, above
the margin, with intervals reaching
+0.051 (Qwen3) and +0.114 (BGE-M3); with
cells of at least 100 items it is
−0.016, and with shrunken growth +0.003,
both with intervals still above +0.1, so
the registered patents decision rested
on noise in small cells and patents are
inconclusive. The valid gap scores
behave the same way: their gains range
from −0.040 to +0.055 over the 15
configurations and three corpora, none
threatening the registered conclusion,
and the result does not depend on the
definition, the resolution or the space.
Momentum is a naive baseline; the point
is that no gain from the map's gaps is
detectable on top of it.

\textbf{Emptier cells grow more once
momentum is held fixed, for most valid
definitions (F2 fails).} I predicted no
association between growth and hole
after controlling for momentum and log
past count. With the registered score
the partial Spearman ρ is 0.17 to 0.20
(Qwen3) and 0.33 to 0.34 (BGE-M3) in
Steam and arXiv, CIs excluding zero, and
0.07 in patents. With the valid scores
the association is definition-dependent:
positive with intervals excluding zero
in both Steam and arXiv and both
embedders for 12 of the 15
configurations (0.20 to 0.34 with 256
cells), in none in patents, and in none
at a 12-month horizon. This agrees, with
that qualification, with Luo and Evans
(\hyperlink{ref-Luo2026}{2026}), who
find in concept space that the future
fills small, nearby holes. It is still
not a usable forecast: the partial
correlation runs over the whole range of
growth, whereas the forecasting question
is which cells grow most, and there no
gap score adds to momentum.

\textbf{More isolated new Steam games
are slightly better reviewed (F5
fails).} The one registered test that
relates distance to quality predicted
that more isolated future items are not
better received. Among future Steam
games with at least 20 reviews (n =
6,697), the share of positive reviews
rises with isolation (Spearman ρ =
+0.107, CI 0.085 to 0.132, \emph{p} = 1
× 10⁻¹⁸, Qwen3; +0.132, CI 0.108 to
0.155, \emph{p} = 2 × 10⁻²⁷, BGE-M3).
Patents go the other way: more isolated
applications are slightly less often
granted (ρ = −0.081 and −0.084, n =
9,477 decided, 30,865 pending excluded).
Both associations survive controls
(registered follow-up,
\texttt{review\_\allowbreak{}f5\_\allowbreak{}controls.md}):
net of log review count, release year,
owners bracket and primary genre the
Steam partial ρ is +0.102 (CI 0.078 to
0.125) and +0.126 (CI 0.103 to 0.149),
positive in each band of review count
(20--99, 100--999, ≥ 1,000), and net of
filing year and CPC section the patents
partial ρ is −0.059 (CI −0.079 to
−0.040) and −0.055 (CI −0.074 to
−0.035); CPC section absorbs about a
third of the patents association. The 1
\% tails (67 and 95 items) are not
distinguishable from the rest in either
domain, but their intervals are wide:
the granted share in the most isolated 1
\% of patents is −0.042 below the rest
(CI −0.137 to +0.028; −0.074 in BGE-M3),
as large as in the most isolated 10 \%
(−0.033 and −0.076), so this is a
non-detection, not an absence. By the
registered tail reading (interval
excluding zero in both embedders) all
four 10 \% tails differ in the expected
direction; the stricter Holm correction
across the 16 tail tests, which the
registration lists but does not use for
the reading, keeps the least isolated 10
\% in every cell (Steam −0.026 and
−0.032 in positive-review share, patents
+0.063 and +0.057 in granted share) and
the most isolated 10 \% in BGE-M3 only.
The effects are small, but the Steam one
is a failed prediction in the direction
of the recipe's premise: among games
that were released, being far from the
catalogue went with slightly better
reception, an association not explained
by review volume, year, reach or genre.
The samples are selected (games with at
least 20 reviews, 6,697 of 40,989;
decided patent applications, 9,477 of
40,342), and the controls do not address
that selection.

\textbf{The item-level test is
inconclusive (F6).} The registered test
closest to the premise, whether the
saved GAP targets are approached by
future items faster than density-matched
past points, could not be run as
designed: no target had the 200
density-matched controls it required,
and the exploratory fallback is biased
in an unknown direction
(\hyperref[app:A.2]{Appendix~A.2}).

\textbf{Nothing lands in the far region
(F1 holds).} None of 122,331 future
items has maximum cosine ≤ 0.15 to the
past; the most isolated has 0.41 to
0.57. Items of one corpus measured
against another corpus's past never land
there either (lowest maximum cosine 0.17
to 0.49, exploratory), so F1 could not
have failed: in these corpora the far
region is empty of other domains' text
too, not only of future text. The
informative quantity is the size of the
shift: the future is only 0.002 to 0.029
more isolated (median) than a held-out
slice of the past, while text from
another domain sits 0.08 to 0.32 further
out.

\textbf{Without cells, new items are
slightly more isolated, but empty points
are not shown to fill first} (registered
follow-up,
\texttt{review\_\allowbreak{}item\_\allowbreak{}density.md}).
Two item-level tests ask the premise
directly. First, how well isolation from
the past map separates future items from
held-out past items: AUC 0.61 to 0.62 in
arXiv, 0.56 to 0.59 in Steam and 0.51 to
0.52 in patents (95 \% intervals about
±0.008), and the same or larger against
held-out items from the last past years
only (arXiv 0.64, Steam 0.55 to 0.60,
patents 0.51), so the shift is not just
later items being later. Inside the
past, the same design gives 0.50 to
0.55, except in Steam with Qwen3, where
earlier past items already separate as
well (0.551, CI 0.536 to 0.566), so
there the shift is no larger than the
past's own. By the registered thresholds
arXiv supports the prediction that new
items are more isolated in both
embedders, patents contradicts it, and
Steam is weak. This is isolation
(distance to the nearest past item), not
position on the map: the latitude test
F4 does not replicate, and in the spaces
without post-split encoder training
below the future moves towards the
centre. In those spaces the future is
also more isolated than held-out past
items (median difference +0.019 and
+0.007 in Steam, +0.025 and +0.023 in
arXiv, +0.004 in patents), so the
isolation shift is not an artefact of
the encoders' training data. Second,
whether empty points of the past map are
covered by the future faster than the
past would predict: for 20,000 points
drawn from the past corpus's Gaussian
per space, the ratio of future to
held-out items within each point's own
tenth-neighbour radius is no higher in
the emptiest decile than in the middle
deciles by the registered margin of 1.25
in any domain (ratio of ratios 1.06 to
1.09 in Steam, 1.10 to 1.15 in arXiv,
with Holm \emph{p} ≥ 0.12, inconclusive,
since arXiv/Qwen3's interval, 1.02 to
1.29, still reaches the margin; 0.92 to
0.97 in patents, which contradicts it;
short of support, the registered rule
calls a result inconclusive whenever the
interval's upper bound reaches 1.10, as
it does in all four Steam and arXiv
cells). The contrast of extreme against
middle deciles misses most of a gradient
that arXiv does show: there the rate
ratio rises with emptiness from 0.48 in
the densest decile to 0.72 and 0.77 in
the emptiest, and the registered
secondary correlation is detected (ρ =
+0.08 and +0.10, Holm \emph{p}
\textless{} 0.006, none of 2,000
resamples crossing zero), while in Steam
it is not (Holm \emph{p} ≥ 0.08), and in
patents it is null in Qwen3 and slightly
negative in BGE-M3 (−0.05, Holm \emph{p}
= 0.04). So new arXiv papers, and to a
lesser degree new games, are somewhat
more isolated than old ones;
preferential filling of the map's empty
points is not shown by the registered
contrast (inconclusive in Steam and
arXiv, contradicted in patents),
although in arXiv emptier points do draw
somewhat more of the future; the probes
catch few items (median one held-out
item each), so the second test is coarse
(\hyperref[app:A.2]{Appendix~A.2}).

\textbf{Cheap combination signals add
nothing to momentum} (pre-registered).
The Alien Space of Science (Artiles et
al., \hyperlink{ref-Artiles2026}{2026})
recovers future idea-atom triples far
above uniform chance but leaves open how
a momentum or popularity baseline would
do. On all 1,313,400 triples of 200 NMF
topics of arXiv, momentum forecasts
which triples appear in 2023--2026
papers (29.1× chance in the top 0.875
\%, AUC 0.969), and static popularity is
nearly as good (28.2×). Within strictly
novel candidates momentum still ranks
the ones that appear (30.9×, AUC 0.962),
although on the full ranking unfitted
momentum puts none of them in its top
0.875 \%. A fitted model free to use
coherence and rarity gains +0.0002 AUC
over fitted momentum, and the embedding
hole adds −0.0003. My coherence proxy,
smoothed PMI, is inverted on its own
(AUC 0.164), because it rewards pairs of
rare topics, so this says that cheap
combination signals add nothing to
momentum, not that Alien Space's trained
model would add nothing
(\hyperref[app:A.6]{Appendix~A.6},
\hyperref[tab:A6]{Table~A6}).

\textbf{In spaces without post-split
encoder training, the conclusions hold,
except that Steam becomes inconclusive
in LSA} (registered follow-up,
\texttt{review\_\allowbreak{}leakfree\_\allowbreak{}encoder.md}).
The encoders were trained on text that
may postdate the splits, and an encoder
that already places later trends near
each other would help momentum and
density alike. I therefore reran the
Section-4 tests and the gap-score
validity study in two spaces without
that leak; the arXiv abstracts are still
latest versions, so these spaces close
the encoder route of leakage, not the
revision route
(\hyperref[sec:2]{Section~2}). In a
TF-IDF/LSA space fitted on past text
only (256 dimensions, all three
corpora), no future item lands in the
far region; the registered hole score is
again invalid while H1 to H3 detect
planted holes; momentum forecasts
top-decile growth (AUC 0.77, 0.89 and
0.69 in Steam, arXiv and patents) better
than every gap score by point estimate,
the registered rule (the intervals
overlap in Steam and patents); the valid
scores stay at chance (0.43 to 0.55, no
interval above 0.5); and none adds to
momentum with an interval above zero. In
Steam their gains are 0.04 to 0.07 with
intervals including zero, so there, as
in patents in the main spaces, the
result is inconclusive. In patents the
gains stay below the registered margin
(at most +0.029), but with cluster
intervals about ±0.15 wide the test
there says little. In all-mpnet-base-v2,
released in 2021 before the Steam and
arXiv splits (not before the patent
split, so patents are covered by LSA
only), every registered conclusion holds
in both corpora: no future item in the
far region, the same validity verdicts,
momentum at AUC 0.85 and 0.89, the valid
scores at chance in Steam (0.53 to 0.57)
and below it in arXiv (0.24, intervals
below 0.34: denser cells grow more, more
clearly than in the main spaces), and
gains over momentum of −0.01 in both, so
in this space Steam is no longer
inconclusive. Two differences from the
main spaces are descriptive: in LSA the
hole score's partial association with
growth (F2) is also positive in patents
(0.21, CI 0.10 to 0.33), and the future
shifts towards the centre of the map in
arXiv and patents in LSA and in both
corpora in mpnet.

\section{Feedback lengthens a continuation's step; the turn it adds is not separable from revision or from the longer step}\label{sec:5}

\textbf{Data.} 150 past arXiv cs papers
dated 2021--2022 (the \emph{contexts}),
sampled with a fixed seed among those
with at least five \emph{follow-ups}:
arXiv papers that cite the context
according to Semantic Scholar (Kinney et
al., \hyperlink{ref-Kinney2023}{2023})
and appeared on or after 2024-01-01,
after every generator's cutoff; at most
50 per context, 2,883 in all. The
threshold selects well-cited papers: in
the frame of 1,000 seeded 2021--2022
papers the median context has 2 such
citing papers (interquartile range 0 to
8) and 35 \% have five or more, against
a median of 12 (8 to 26) in the sample.
Within the sample I detect no gradient
of the step gap with citation count
(Spearman 0.05, CI −0.12 to +0.21),
which does not show the absence of one
below the threshold. Throughout,
``follow-up'' means a later citing
paper, nothing more: labelling all 2,883
context/follow-up pairs blind from the
two abstracts, two LLM labellers
(gpt-6.1-sol and Claude Opus 5.5;
Cohen's κ = 0.75 on continues-or-not)
agree that 112 (3.9 \%) continue the
context's own work (95 \% CI 2.4 to 5.6
\% for similar contexts, cluster
bootstrap over contexts), and the rest
mostly cite it as background or
comparison (Section 8). What this
section measures is whether the ideas
move where the literature that cites a
paper later went; Section 5.1 checks the
step against the continuations alone.
For robustness I also use the subset of
follow-ups that Semantic Scholar marks
as influential (Valenzuela et al.,
\hyperlink{ref-Valenzuela2015}{2015}) or
with a methodology or result intent
(Cohan et al.,
\hyperlink{ref-Cohan2019}{2019}),
keeping contexts with at least three
(the \textbf{S2-flagged subset}, 68
contexts). It is a robustness check on a
different set, not a cleaner one: of its
403 follow-ups, 40 (10 \%) are among the
112 both labellers call continuations,
against 2.5 \% of the follow-ups
Semantic Scholar does not flag. A few
later papers cite two sampled contexts
(24 of 2,859 distinct follow-ups,
touching 32 contexts); resampling
connected groups of contexts instead of
single contexts changes no reading in
this sample.

\textbf{Measures.} For an idea or
follow-up \(\mathbf{x}\) of context
\(\mathbf{c}\), \textbf{step} =
\(\cos(\mathbf{c},
\mathbf{x})\), so a lower value is a
longer step, and ``lengthening the
step'' means lowering it. \textbf{Fit}
is the maximum cosine of an idea to the
context's follow-ups. The \textbf{step
gap} is a context's median idea step
minus its median follow-up step. The
unit of analysis is the context, with
bootstrap CIs over contexts (2,000
resamples unless stated) and Holm
correction within each registered
family.

\subsection{The short step replicates against citing papers}\label{sec:5.1}

Asked once to ``propose the paper that
should come next'' from a context's
title and abstract, gpt-4o writes ideas
that stay closer to the context than the
real follow-ups: the mean step gap is
0.149 (CI 0.138 to 0.160), and 0.140
with the eight nearest past items shown
(pre-registered B1, Holm \emph{p}
\textless{} 10⁻²⁴). The gap holds in
BGE-M3 (0.168), in the S2-flagged subset
(0.129 and 0.159), and with OLMo (0.179
to 0.218), with \emph{p} \textless{}
10⁻¹¹ in all eight generator × embedder
× follow-up-set cells. This is Tang and
Yang's (\hyperlink{ref-Tang2026}{2026})
finding, not a new one; my design adds a
single-paper context and generators
whose training predates every follow-up:
querying the arXiv API, no follow-up's
first version is earlier than 2024-01-01
(2025-09-01 for
\hyperref[sec:5.4]{Section~5.4}), and
where Semantic Scholar's date differs
from the first version it is the earlier
one. Abstracts are latest versions, not
first ones (42 \% of follow-ups and 51
\% of contexts have more than one
version; 8 of the 150 contexts were last
revised after gpt-4o's cutoff), which
changes the text the model saw, not
whether it could have seen a follow-up.
Other one-shot variants (memory of
earlier titles, a verbal hint of where
the area is moving) narrow the gap, but
by at most a fifth (memory: 0.022 of
0.149, CI 0.016 to 0.028); they are in
\hyperref[app:A.8]{Appendix~A.8}.

\textbf{The step does not hinge on the
user prompt's wording} (registered
follow-up). ``The paper that should come
next'' could itself ask for a near step.
With two other user prompts, a paper
``that will cite this one, written about
three years from now'' and ``a research
idea that builds on this paper'', 750
new ideas each on the same contexts, the
gap is 0.135 (CI 0.123 to 0.146) and
0.130 (CI 0.118 to 0.142) (Holm \emph{p}
\textless{} 10⁻²⁵ for both). All three
keep the system prompt ``You are a
researcher proposing the next paper'',
so the phrase is varied in the request
only, not removed. Both wordings narrow
the gap slightly but detectably: by
0.015 (CI 0.010 to 0.020) and 0.020 (CI
0.014 to 0.026), Holm \emph{p} = 1 ×
10⁻⁶ and 1 × 10⁻⁷, so they keep 0.90 (CI
0.87 to 0.94) and 0.87 (CI 0.83 to 0.91)
of the gap; in BGE-M3 the gaps are 0.164
(CI 0.156 to 0.171) and 0.159 (CI 0.151
to 0.167) against 0.168, narrower by
0.005 (CI 0.0002 to 0.009) and 0.009 (CI
0.005 to 0.013), Holm \emph{p} = 0.02
and 2 × 10⁻⁵, keeping 0.97 and 0.95. A
budget rule fixed in advance limited
this test to the discovery sample.

\textbf{The gap survives a
genre-and-echo control} (registered
follow-up). The ideas are proposals
written with the context's abstract in
the prompt; the follow-ups are real
abstracts. Both the shared proposal
genre and reuse of the context's wording
would raise an idea's cosine to its
context without its going less far. The
ideas do reuse the wording more: 3-gram
containment with the context's abstract
is 0.038 for gpt-4o's one-shot ideas
against 0.007 for follow-ups and 0.018
for continuations, and at equal overlap
the gap shrinks by about a third (echo
share 0.35, CI 0.29 to 0.41, against all
follow-ups; 0.41, CI 0.25 to 0.72,
against continuations). The direct test
has gpt-4o rewrite every follow-up as a
proposal for the paper that should come
next, with the context shown and its
content kept, and compares the ideas
with the rewrites. The gap survives in
every cell: 0.112 (CI 0.102 to 0.122)
against the rewritten follow-ups (0.149
against the originals), 0.108 in the
replication sample of
\hyperref[sec:5.3]{Section~5.3} (0.161
originally), 0.118 for OLMo's ideas
against OLMo's rewrites (0.198), and
0.097 to 0.102 in BGE-M3; Holm over the
two primary tests \emph{p} = 6 × 10⁻²⁶.
Rewriting a real paper this way moves it
0.037 closer to the context (CI 0.027 to
0.048), but that move comes from the 22
\% of rewrites that drift from the
paper's content (cosine to the original
below 0.80): without them rewriting
moves the papers by 0.001 and the gap is
0.129. Showing the context in the
rewrite prompt can itself add echo, so I
registered a cleaner control: the same
rewrite without the context, for all
2,883 follow-ups (the discovery sample
only, under a budget rule fixed in
advance). These rewrites keep the
papers' content better (6.9 \% below
0.80, against 22 \%), and the gap
against them is 0.232 (CI 0.221 to
0.242) in Qwen3 and 0.167 (CI 0.159 to
0.175) in BGE-M3, larger than against
the context-shown rewrites in both (by
0.119 and 0.071, intervals excluding
zero): 1.55 times the original gap (CI
1.50 to 1.61) and 2.06 times the
context-shown one (CI 1.92 to 2.22) in
Qwen3, 0.99 (CI 0.98 to 1.01) and 1.73
(CI 1.63 to 1.85) in BGE-M3. Dropping
the 199 no-context rewrites that drift
from the paper's content leaves the gap
at 0.228 (0.166 in BGE-M3). Part of the
shrinkage to 0.112 was therefore echo of
the shown context, and the
genre-controlled gap is at least the
original one in Qwen3 and equal to it in
BGE-M3: without the context, rewriting
into the proposal genre leaves the
papers' step unchanged in BGE-M3
(+0.001, \emph{p} = 0.31) and moves them
0.082 away from the context in Qwen3.
The drift account above is therefore
incomplete: in Qwen3 even
content-preserving rewrites that saw the
context sit about 0.08 closer to it than
rewrites that did not. Why a
content-preserving rewrite moves a real
paper 0.082 away in one encoder and not
at all in the other I do not know; it
may come from the rewriting itself
(length normalisation, a proposal with
no antecedent) rather than from genre,
so I do not interpret the Qwen3 excess
over 0.149. The registration named the
no-context gap as the genre-controlled
figure; I keep 0.112 as the lower,
conservative one and report both. With
rewrites in the echo model, wording
accounts for a quarter of the gap (0.24,
CI 0.20 to 0.28). Removing the mean
difference between generated proposals
and real abstracts from every embedding
does not shrink the gap either; it
widens it (0.149 to 0.220).

\textbf{Against the papers that continue
the work, the gap is smaller but real.}
I registered this check in two steps,
after the main results were known. A
first sample of 400 pairs, extended to
every follow-up of the 25 contexts where
it found a continuation, gave 18
contexts with a continuation both
labellers agree on: an adaptively
ascertained set, in which gpt-4o's gap
against continuations was not detected
(0.023, CI −0.025 to +0.069). I then
registered labelling the rest of the
census, so that the test covers every
context with a continuation: 47
contexts. In them continuations sit 0.12
closer to their context than the
follow-ups both labellers call
background (CI 0.10 to 0.15; closer in
45 of the 47). gpt-4o's one-shot ideas
still stay closer to the context than
the continuations do: the gap is 0.048
(CI 0.019 to 0.076, one-sided Wilcoxon
\emph{p} = 0.0015), positive in 30 of
the 47 contexts, against 0.144 for all
follow-ups of the same contexts. The
registered test passes, and it passes on
the 29 contexts never analysed before
(0.064, CI 0.028 to 0.095, \emph{p}
\textless{} 0.001), while the 18 earlier
ones reproduce the earlier non-detection
(0.023). The gap is larger in BGE-M3
(0.086) and for OLMo (0.119 and 0.102),
so against all follow-ups of the same
contexts the ratio is 0.33 in the
primary cell and 0.52 to 0.56 in the
others. Against continuations rewritten
as proposals the gap is 0.081 (CI 0.054
to 0.108; Holm \emph{p} = 3 × 10⁻⁷),
larger than against the originals; for
OLMo it is 0.078. Against continuations
rewritten without the context it is
0.134 (CI 0.105 to 0.164; 0.093 in
BGE-M3). The short step is, then, a step
short of the citing literature and, by a
third to a half as much, of the papers
that continue the work, and neither is
explained by genre or wording. Nor,
within the follow-ups I have, by how
much later they appeared (exploratory):
the gap is 0.153, 0.155 and 0.155
against follow-ups first posted on arXiv
in 2024, 2025 and 2026 (Qwen3; 0.164,
0.170 and 0.177 in BGE-M3), so later
follow-ups are not closer to the
context; within contexts they are
slightly farther (−0.004 and −0.006 per
year from 2023 to 2026, intervals
excluding zero;
\hyperref[app:A.8]{Appendix~A.8}).
Earlier citing papers do not close it
either (registered follow-up): OLMo's
training data stop in February--March
2023, and against 1,204 citing papers
first posted from April to December 2023
(143 contexts) its gap is 0.215 (CI
0.200 to 0.231); on the 139 contexts
with both years it is 0.216 against
0.221 for 2024 (paired difference
−0.006, CI −0.020 to +0.009; BGE-M3
0.185 against 0.182). The idea step
cancels in this paired difference, so
the test compares the follow-ups
themselves: no difference was detected
between citing papers from just after
OLMo's training data and those of 2024
(interval above); OLMo matters because
it could not have seen them. The
restriction to follow-ups from 2024 on
does not create the short step; citing
papers from before April 2023 remain
untested
(\hyperref[app:A.8]{Appendix~A.8}). The
continuations are labelled by LLMs from
abstracts; in a registered blind check
of 40 pairs I confirmed 24 of the 25
consensus continuations sampled, about
half of them with some doubt
(\hyperref[sec:8]{Section~8}).

\subsection{Iterative critique (pre-registered)}\label{sec:5.2}

\textbf{Design.} Same 150 contexts and
generator. Per context and arm, three
independent ideas, each refined over
three rounds (the one-shot prompt, then
two revisions that see the context, the
previous proposal and the round's
feedback). I call all three feedback
arms \textbf{critique}:

\begin{itemize}
\tightlist
\item
  \textbf{GEO} (rule-based): the
  proposal's cosine to the context and
  the interquartile range of real
  follow-up steps (0.574 to 0.661), with
  ``take a bigger step'', ``keep a
  clearer link'' or ``keep the step''.
  The band is an aggregate of the same
  150 contexts, a stated mild leakage;
  the replication of
  \hyperref[sec:5.3]{Section~5.3} reuses
  it on new contexts, where it does not
  leak.
\item
  \textbf{Reviewer} (verbal): a separate
  gpt-4o call says whether the proposal
  is a genuine follow-up or a
  rephrasing, and what would make it a
  stronger next paper, in at most 80
  words. It never sees a follow-up.
\item
  \textbf{ORACLE} (leaks the future):
  the proposal's maximum cosine to the
  context's real follow-ups, ``try to
  get closer''. This is a
  \textbf{single-score oracle with no
  history or best-so-far}: each revision
  sees one number about its previous
  proposal.
  \hyperref[sec:5.3]{Section~5.3} tests
  a version with a fixed target and
  \hyperref[sec:5.4]{Section~5.4} one
  with the history of scores.
\end{itemize}

Predictions: C1, GEO lengthens the step
and closes at least 0.07 (in cosine) of
the step gap; C2, GEO does not lower the
hit rate by more than 0.03; C3, the
reviewer lengthens the step; C4,
two-sided hit change under the reviewer;
C5, ORACLE raises fit, by less than
0.07. \textbf{Hit} is whether an idea's
nearest neighbour among the follow-ups
and the same-category later papers is a
follow-up; it depends on near ties and
on the distractor set, so C2 and C4 are
reported in
\hyperref[app:A.8]{Appendix~A.8} (both
find a drop) and no claim rests on them.

\begin{table}[tbp]
\centering\footnotesize\setlength{\tabcolsep}{4pt}
\caption[]{\textbf{Table 3.} Iterative critique (gpt-4o, Qwen3), round 0 → round 2: means over 150 contexts of each context's mean over its three ideas. Real follow-ups: median step 0.620.}\label{tab:3}
\begin{tabular}{@{}llll@{}}
\toprule
Arm & step & fit & hit \\
\midrule
GEO & 0.766~→~0.679 & 0.677~→~0.636 & 0.627~→~0.518 \\
Reviewer & 0.766~→~0.703 & 0.679~→~0.647 & 0.644~→~0.591 \\
ORACLE & 0.761~→~0.729 & 0.677~→~0.666 & 0.664~→~0.611 \\
NEUTRAL (added later) & 0.766~→~0.722 & 0.677~→~0.662 & 0.627~→~0.613 \\
\bottomrule
\end{tabular}
\end{table}

\begin{figure}[!htb]
\centering
\includegraphics[width=\linewidth]{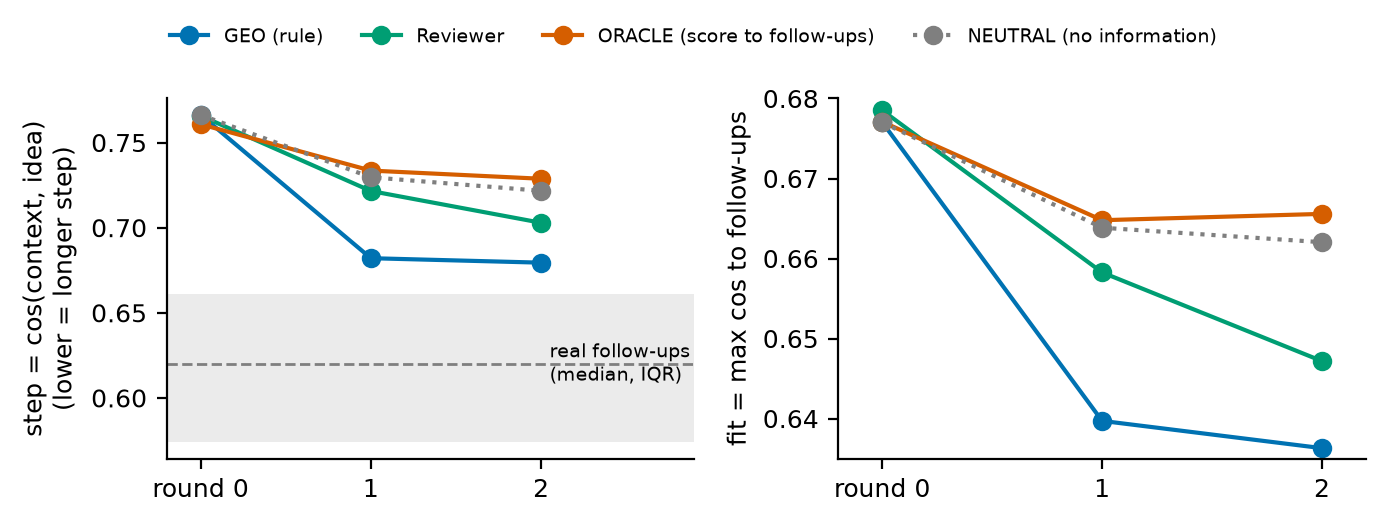}
\caption[]{\textbf{Figure 3.} Iterative critique (gpt-4o, Qwen3, 150 contexts, three ideas per context and arm). Left: critique lengthens the step towards that of the real follow-ups (grey band: their interquartile range); revising with no information (NEUTRAL, dotted) already does part of it. Right: fit to those follow-ups falls in every arm, including the one whose feedback is the similarity to them. The right panel's y-axis spans only 0.635 to 0.680, which magnifies these changes.}\label{fig:3}
\end{figure}

\textbf{Distance feedback closes most of
the step gap (C1 holds).} GEO lengthens
the step by 0.087 (CI 0.080 to 0.094),
closing 0.087 of the 0.148 gap of these
round-0 ideas (the separate one-shot run
of \hyperref[sec:5.1]{Section~5.1} gives
0.149), and the reviewer by 0.063 (CI
0.056 to 0.069; C3 holds). \textbf{C5
fails:} I predicted that ORACLE would
raise fit. It did not: fit fell by 0.012
(CI −0.016 to −0.007).
\hyperref[fig:3]{Figure~3} shows the
step and fit over rounds.

\textbf{Feedback lengthens the step
beyond content-free revision}
(pre-registered, added later). None of
the arms above separates the feedback
from revising as such. A NEUTRAL arm
runs the same two revisions from the
same round-0 ideas as GEO, each told
only ``Revise your proposal: write an
improved version.'' (900 gpt-4o calls).
Revising with no information already
lengthens the step by 0.045 (BGE-M3
0.034) and lowers fit by 0.015. I
pre-registered six comparisons of each
critique arm's change with NEUTRAL's,
paired by context, with Holm over their
14 tests in each embedder
(\hyperref[tab:4]{Table~4}): N1 and N2
predict that GEO and the reviewer
lengthen the step more than NEUTRAL; N3
to N6 ask, two-sided, whether fit, hit,
alignment and the follow-up-specific
change (\hyperref[sec:5.3]{Section~5.3})
differ. Only GEO shares its round-0
ideas with NEUTRAL, so the reviewer's
and ORACLE's contrasts compare
independent round-0 draws and are
noisier.

\begin{table}[tbp]
\centering\footnotesize\setlength{\tabcolsep}{4pt}
\caption[]{\textbf{Table 4.} Critique arm minus NEUTRAL, change from round 0 to round 2 (gpt-4o; Qwen3, with BGE-M3 in brackets; Holm-adjusted \emph{p}). The last row divides N6 by the gap G between the follow-ups' own projection on the follow-up-specific direction and the round-0 ideas' (\hyperref[sec:5.3]{Section~5.3}), with bootstrap 95 \% intervals; ±0.015 is 0.16 of G (0.14 in BGE-M3).}\label{tab:4}
\begin{tabular}{@{}P{\dimexpr 0.235\linewidth-2\tabcolsep\relax}P{\dimexpr 0.255\linewidth-2\tabcolsep\relax}P{\dimexpr 0.255\linewidth-2\tabcolsep\relax}P{\dimexpr 0.255\linewidth-2\tabcolsep\relax}@{}}
\toprule
Test & GEO − NEUTRAL & Reviewer − NEUTRAL & ORACLE − NEUTRAL \\
\midrule
N1, N2: step lengthened (one-sided) & +0.042, \emph{p}~2~×~10⁻¹⁶\newline {[}+0.013,~2~×~10⁻⁵{]} & +0.018, \emph{p}~2~×~10⁻⁴\newline {[}+0.014,~6~×~10⁻⁶{]} & not tested \\
N3: fit & −0.026, \emph{p}~6~×~10⁻¹¹\newline {[}−0.002,~1{]} & −0.016, \emph{p}~0.001\newline {[}−0.009,~0.001{]} & +0.0035, \emph{p}~1\newline {[}−0.003,~0.75{]} \\
N4: hit & −0.096, \emph{p}~0.001\newline {[}−0.033,~0.61{]} & −0.038, \emph{p}~0.74\newline {[}−0.033,~1{]} & −0.040, \emph{p}~0.86\newline {[}+0.002,~1{]} \\
N5: alignment & +0.003, \emph{p}~1\newline {[}+0.012,~0.21{]} & +0.000, \emph{p}~1\newline {[}−0.003,~1{]} & −0.016, \emph{p}~0.12\newline {[}−0.014,~0.22{]} \\
N6: follow-up-specific & −0.003, \emph{p}~1\newline {[}+0.010,~0.48{]} & −0.013, \emph{p}~0.14\newline {[}−0.008,~0.94{]} & −0.022, \emph{p}~0.018\newline {[}−0.014,~0.27{]} \\
N6 as a share of the gap G (\(\psi\)) & −0.04, CI −0.15~to~+0.07\newline {[}+0.10,~CI +0.00~to~+0.19{]} & −0.13, CI −0.27~to~−0.02\newline {[}−0.08,~CI −0.19~to~+0.03{]} & −0.23, CI −0.39~to~−0.09\newline {[}−0.13,~CI −0.24~to~−0.02{]} \\
\bottomrule
\end{tabular}
\end{table}

N1 and N2 hold in both embedders, and
the replication of
\hyperref[sec:5.3]{Section~5.3} repeats
them on 150 new contexts (GEO +0.040,
reviewer +0.023 beyond NEUTRAL, Holm
\emph{p} \textless{} 10⁻⁶; +0.017 and
+0.014 in BGE-M3). Distance feedback
thus closes about 0.04 of the step gap
that revising alone does not. Revising
alone accounts for about half of GEO's
lengthening (0.045 of 0.087) and over
two thirds of the reviewer's (0.045 of
0.063). GEO and the reviewer also lose
more fit than NEUTRAL (N3; in BGE-M3
only the reviewer). On direction, no
critique arm turns the ideas more than
NEUTRAL after Holm (N5, N6), and the
single-score oracle turns them less in
the follow-up-specific direction (N6,
−0.022, CI −0.034 to −0.008, Holm
\emph{p} = 0.018 in Qwen3; −0.014,
uncorrected \emph{p} = 0.03 in BGE-M3).
Exploratory: ORACLE lengthens the step
less than NEUTRAL (−0.013, \emph{p} =
0.003).

\textbf{Falling fit is not, by itself, a
wrong direction.} Moving away from the
context lowers the cosine to follow-ups
that share the context's direction,
whatever the direction of the move. So
the falls in \hyperref[tab:3]{Table~3}
and \hyperref[fig:3]{Figure~3} say that
critique did not bring the ideas closer
to the follow-ups; they do not say where
it turned them. That needs a measure
that separates direction from distance.

\subsection{Direction: a radial-invariant measure and a positive control}\label{sec:5.3}

The obvious baseline, comparing each
idea with the later paper whose distance
to the context is closest, is not
identified here: at round 0, 69 \% to 97
\% of the ideas are closer to their
context than \emph{any} later paper of
the corpus, and 40 \% to 84 \% still are
at round 2, so the matched paper is
systematically further out than the idea
and the comparison mixes direction with
step (\hyperref[app:A.9]{Appendix~A.9}).
I therefore use a measure that ignores
the radial component.

For an idea \(\mathbf{x}\) of context
\(\mathbf{c}\), its \textbf{tangent} is
the unit vector of \(\mathbf{x} -
(\mathbf{x}\cdot\mathbf{c})\,\mathbf{c}\),
the part of the move that is not towards
or away from \(\mathbf{c}\). Let
\(\mathbf{f}\) be the unit mean tangent
of the context's follow-ups,
\(\mathbf{p}\) the unit mean tangent of
its nearest later papers outside the
sampled follow-up set (as many as it has
follow-ups), and \(\mathbf{s}\) the unit
vector of
\(\mathbf{f} - (\mathbf{f}\cdot\mathbf{p})\,\mathbf{p}\),
the \textbf{follow-up-specific}
direction. That pool may contain citing
papers, because 18 of the 150 contexts
hit the cap of 50 saved follow-ups and
citing papers from 2023 were not
excluded; removing every citing arXiv
paper of any date (9,921, fetched
without a cap) changes no contribution
in Table 4 by more than 0.0011. An
idea's \textbf{alignment} is the cosine
of its tangent with \(\mathbf{f}\).
Because \(\mathbf{f} =
(\mathbf{f}\cdot\mathbf{p})\,\mathbf{p} + \sqrt{1 - (\mathbf{f}\cdot\mathbf{p})^2}\,\mathbf{s}\),
the change in alignment from round 0 to
round 2 splits exactly into a
contribution along \(\mathbf{p}\) and
one along \(\mathbf{s}\), each weighted
by its context's coefficient (mean
weights 0.71 and 0.67 in Qwen3, 0.76 and
0.63 in BGE-M3). The real follow-ups'
own leave-one-out alignment (each
follow-up against the mean of the
others) is 0.468 in Qwen3 and 0.525 in
BGE-M3, so the measure detects a shared
orientation when there is one. The
tangent is invariant to radial scaling
by construction, but that does not make
it independent of step length: an idea
close to a paraphrase of its context has
a residual dominated by genre and
wording, and as it moves away topical
content enters the residual, which
raises its cosine with any direction
built from real later papers. Tests in
\hyperref[tab:5]{Table~5} are paired
Wilcoxon over contexts, with Holm over
its 36 tests (4 cells × 3 arms ×
alignment, along \(\mathbf{p}\), along
\(\mathbf{s}\)). OLMo runs only the arms
its 2,048-token window allows and one
idea per context and arm
(\hyperref[app:A.8]{Appendix~A.8}).

\begin{table}[tbp]
\centering\footnotesize\setlength{\tabcolsep}{4pt}
\caption[]{\textbf{Table 5.} Tangent alignment with the follow-ups' direction, round 0 → round 2 (exploratory; means over contexts). Alignment with its Holm-adjusted \emph{p}; then the weighted contributions along \(\mathbf{p}\) and along \(\mathbf{s}\), which add up to the alignment change; then the follow-up-specific change itself (unweighted), with raw and Holm-adjusted \emph{p}. The NEUTRAL rows come from the separately pre-registered control (its own tests are N5 and N6, \hyperref[tab:4]{Table~4}), are outside the 36 tests and have no weighted decomposition. The last column gives the change along \(\mathbf{s}\) as a share of the gap G between the follow-ups' own projection on \(\mathbf{s}\) and the round-0 ideas' (intervals and the feedback-specific shares in \hyperref[tab:A9]{Table~A9}; ±0.015 is 0.16 of G for gpt-4o and 0.12 for OLMo in Qwen3, 0.14 and 0.12 in BGE-M3). The rise in alignment is descriptive: it is not separable from the longer step (see text).}\label{tab:5}
\begin{tabular}{@{}P{\dimexpr 0.100\linewidth-2\tabcolsep\relax}P{\dimexpr 0.120\linewidth-2\tabcolsep\relax}P{\dimexpr 0.235\linewidth-2\tabcolsep\relax}P{\dimexpr 0.105\linewidth-2\tabcolsep\relax}P{\dimexpr 0.105\linewidth-2\tabcolsep\relax}P{\dimexpr 0.245\linewidth-2\tabcolsep\relax}P{\dimexpr 0.090\linewidth-2\tabcolsep\relax}@{}}
\toprule
Generator / embedder & Arm & alignment r0 → r2 (Holm \emph{p}) & contrib. along \(\mathbf{p}\) & contrib. along \(\mathbf{s}\) & change along \(\mathbf{s}\) (\emph{p}; Holm \emph{p}) & share of G (\(\varphi\)) \\
\midrule
gpt-4o / Qwen3 & GEO & 0.232~→~0.265~(1~×~10⁻⁷) & +0.022 & +0.012 & +0.015~(0.01; 0.11) & +0.16 \\
 & Reviewer & 0.233~→~0.264~(9~×~10⁻⁹) & +0.026 & +0.004 & +0.006~(0.09; 0.47) & +0.07 \\
 & ORACLE & 0.238~→~0.252~(0.06) & +0.017 & −0.002 & −0.003~(0.54; 1) & −0.03 \\
 & NEUTRAL (\hyperref[sec:5.2]{Section~5.2}) & 0.232~→~0.262 & --- & --- & +0.019~(5~×~10⁻⁴; see N6) & +0.20 \\
gpt-4o / BGE-M3 & GEO & 0.393~→~0.435~(2~×~10⁻⁹) & +0.024 & +0.017 & +0.024~(3~×~10⁻⁴; 0.005) & +0.23 \\
 & Reviewer & 0.394~→~0.420~(4~×~10⁻⁹) & +0.023 & +0.004 & +0.006~(0.16; 0.62) & +0.06 \\
 & ORACLE & 0.397~→~0.412~(3~×~10⁻⁴) & +0.015 & +0.001 & +0.001~(0.91; 1) & +0.01 \\
 & NEUTRAL (\hyperref[sec:5.2]{Section~5.2}) & 0.393~→~0.423 & --- & --- & +0.014~(0.007; see N6) & +0.14 \\
OLMo / Qwen3 & GEO & 0.221~→~0.297~(9~×~10⁻⁹) & +0.059 & +0.018 & +0.024~(0.01; 0.11) & +0.19 \\
 & Reviewer & 0.210~→~0.263~(1~×~10⁻⁶) & +0.033 & +0.020 & +0.032~(5~×~10⁻⁵; 8~×~10⁻⁴) & +0.25 \\
 & ORACLE & 0.215~→~0.253~(0.001) & +0.022 & +0.017 & +0.025~(0.02; 0.11) & +0.19 \\
OLMo / BGE-M3 & GEO & 0.360~→~0.436~(3~×~10⁻¹²) & +0.057 & +0.019 & +0.028~(0.007; 0.07) & +0.24 \\
 & Reviewer & 0.351~→~0.405~(3~×~10⁻⁹) & +0.044 & +0.010 & +0.018~(0.02; 0.11) & +0.15 \\
 & ORACLE & 0.360~→~0.391~(0.002) & +0.023 & +0.008 & +0.012~(0.19; 0.62) & +0.10 \\
\bottomrule
\end{tabular}
\end{table}

\begin{figure}[!htb]
\centering
\includegraphics[width=0.85\linewidth]{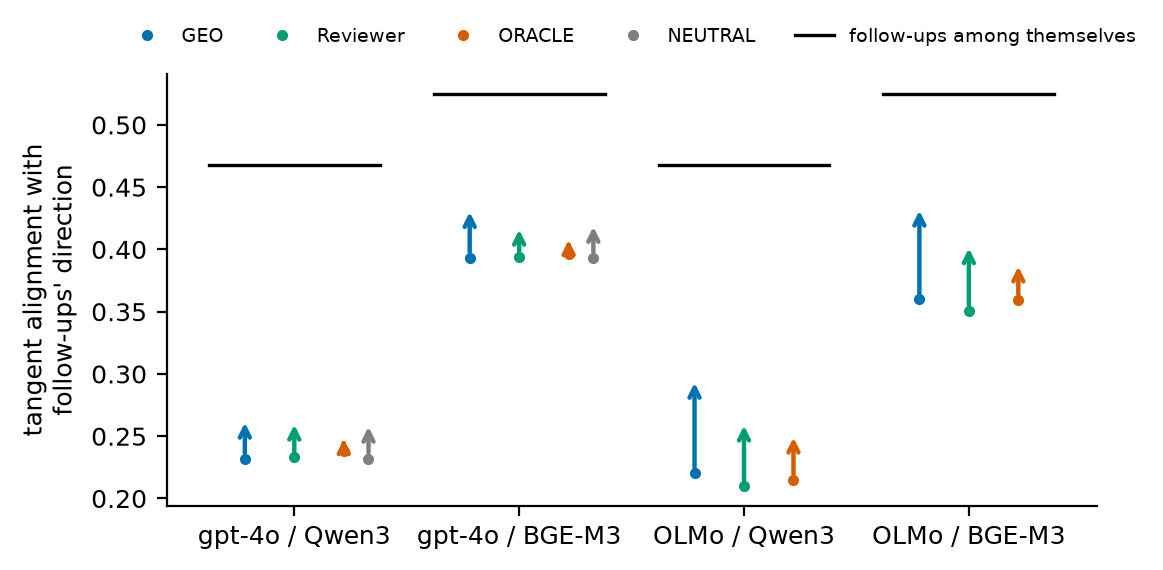}
\caption[]{\textbf{Figure 4.} Radially invariant direction (exploratory). Arrows go from round 0 to round 2 of each arm; the black bar is the real follow-ups' leave-one-out alignment with each other. Every arm turns towards the follow-ups' direction, including NEUTRAL (grey, gpt-4o only), which revises with no information; no arm gets near the alignment the follow-ups share. The rise co-varies with the step and is not separable from it (\hyperref[sec:5.3]{Section~5.3}).}\label{fig:4}
\end{figure}

\textbf{The rise in alignment is not
evidence of direction.} Alignment with
the follow-ups' direction rises after
critique in all 12 generator × embedder
× arm cells (11 of 12 after Holm), the
contribution towards nearby later papers
in general carries 56 \% to 95 \% of it,
and the ideas stay well below the
follow-ups' own alignment (at most 0.297
against 0.468 in Qwen3;
\hyperref[fig:4]{Figure~4}). For gpt-4o,
NEUTRAL, which revises with no
information, turns the ideas about as
much, follow-up-specific part included
(+0.019 in Qwen3, +0.014 in BGE-M3).
Three registered follow-ups then show
that the rise cannot be read as a turn
towards the follow-ups. (i) Within each
arm, contexts whose ideas lengthen the
step more also gain more alignment
(common slope 0.33, CI 0.18 to 0.48, in
this sample; 0.27 in the replication;
0.15 in the OLMo test), and among
round-0 ideas, before any feedback,
those further from their context are
already more aligned, while among real
follow-ups the slope is negative; the
rise is therefore partly a step effect,
and adjusted for step change the
reviewer turns \emph{less} than NEUTRAL
in this sample (−0.017, CI −0.028 to
−0.005, Holm \emph{p} = 0.024), as
ORACLE already did. (ii) Texts with no
follow-up information give a null:
paraphrases of the context and gpt-4o
rewrites of nearby later papers that do
not cite it. At the same step the ideas
project a little more on \(\mathbf{s}\)
than those rewrites (+0.036, CI 0.021 to
0.052; +0.014 in the replication; −0.006
for OLMo) but less on the follow-ups'
direction overall (−0.024 and −0.029),
and the dose-response fitted on the null
texts predicts at least the observed
rise in alignment for most critique arms
in both gpt-4o samples (for GEO 0.057
predicted against 0.033 observed). By
the registered rule, the rise is not
separable from a step effect with these
controls, in every cell and both
embedders. (iii) Removing a genre
direction (mean idea minus mean real
abstract) leaves
\hyperref[tab:5]{Table~5} standing
(alignment Holm-significant in 11 of 12
cells, the same two follow-up-specific
changes significant) but changes four
contrast readings, which the next
paragraphs report. The table is
therefore descriptive. The only evidence
of direction in this section is the
positive control below.

\textbf{Scale.} Reading changes along
\(\mathbf{s}\) needs a yardstick. The
real follow-ups' own leave-one-out
projection on \(\mathbf{s}\) is 0.25 to
0.27 in the confirmatory samples, and
the gap between it and the round-0
ideas' projection is 0.09 to 0.15 in the
gpt-4o and OLMo samples; ±0.015 is about
a tenth to a sixth of that gap (0.11 in
the gpt-4o replication, 0.10 in the OLMo
test, 0.16 in the discovery sample) and
a third for gpt-5.4-mini, whose ideas
start closer
(\hyperref[sec:5.4]{Section~5.4}). In
the replication, feedback beyond
revision closes +0.02 to +0.08 of the
gap along \(\mathbf{s}\) across the
three arms, with upper bounds of at most
0.16; \hyperref[tab:A9]{Table~A9}
(\hyperref[app:A.8]{Appendix~A.8}) gives
each arm's total and feedback-specific
fractions in every sample and embedder.

\textbf{A confirmatory replication with
gpt-4o} (registered follow-up). The
first 150 contexts were read against a
bound of ±0.015 chosen after their
results, so I treat them as a discovery
sample. I repeated the four arms (GEO,
reviewer, ORACLE, and NEUTRAL from GEO's
round-0 ideas; 3 ideas, 3 rounds; 5,850
gpt-4o calls) on 150 new contexts: the
first 150 with at least five follow-ups
in a new frame of 2021--2022 arXiv cs
papers, none of them among the original
150, with 2,419 follow-ups dated by
arXiv first version from 2024 on. The
bound, eight tests and a reading rule
were fixed before any generation:
equivalence within ±0.015 when the 95 \%
t interval lies inside it after Holm
over the eight; a change detected when
that interval excludes zero without
equivalence; otherwise insufficient.
\hyperref[tab:6]{Table~6} gives the
results with a Holm correction over the
12 detection tests of both embedders,
which the registered rule did not apply
and which I add here post hoc.

\begin{table}[tbp]
\centering\footnotesize\setlength{\tabcolsep}{4pt}
\caption[]{\textbf{Table 6.} gpt-4o critique replication on 150 new contexts: change from round 0 to round 2 along \(\mathbf{s}\) (E1--E3), the same minus NEUTRAL's (E4--E6), and the step lengthened beyond NEUTRAL (R1, R2), with 95 \% t intervals. Holm \emph{p}: registered family for R1 and R2; Holm over the 12 detection tests (post hoc) for E1--E6, with the registered equivalence test for BGE-M3 E3.}\label{tab:6}
\begin{tabular}{@{}P{\dimexpr 0.240\linewidth-2\tabcolsep\relax}P{\dimexpr 0.210\linewidth-2\tabcolsep\relax}P{\dimexpr 0.110\linewidth-2\tabcolsep\relax}P{\dimexpr 0.210\linewidth-2\tabcolsep\relax}P{\dimexpr 0.230\linewidth-2\tabcolsep\relax}@{}}
\toprule
Test & Qwen3 & Holm \emph{p} & BGE-M3 & Holm \emph{p} \\
\midrule
R1: GEO − NEUTRAL, step & +0.040 {[}0.032,~0.048{]} & \textless{} 10⁻¹⁴ & +0.017 {[}0.012,~0.022{]} & \textless{} 10⁻⁴ \\
R2: Reviewer − NEUTRAL, step & +0.023 {[}0.016,~0.030{]} & \textless{} 10⁻⁶ & +0.014 {[}0.008,~0.019{]} & \textless{} 10⁻⁴ \\
E1: GEO & +0.019 {[}0.009,~0.029{]} & 0.003 & +0.020 {[}0.010,~0.031{]} & 0.002 \\
E2: Reviewer & +0.012 {[}0.004,~0.020{]} & 0.046 & +0.010 {[}0.001,~0.019{]} & 0.20 \\
E3: ORACLE & +0.012 {[}0.003,~0.021{]} & 0.10 & +0.004 {[}−0.003,~+0.011{]} & 1 (within ±0.015: 0.022) \\
E4: GEO − NEUTRAL & +0.011 {[}−0.000,~+0.022{]} & 0.36 & +0.011 {[}0.002,~0.021{]} & 0.20 \\
E5: Reviewer − NEUTRAL & +0.004 {[}−0.009,~+0.016{]} & 1 & +0.001 {[}−0.010,~+0.012{]} & 1 \\
E6: ORACLE − NEUTRAL & +0.003 {[}−0.010,~+0.016{]} & 1 & −0.005 {[}−0.016,~+0.007{]} & 1 \\
\bottomrule
\end{tabular}
\end{table}

The step result replicates. Each arm's
own change along \(\mathbf{s}\) is
positive, up to about 0.02, and is not
confirmed within ±0.015 except ORACLE's
in BGE-M3, so the discovery sample's
``at most about 0.015'' is withdrawn for
gpt-4o. After correction, GEO's turn is
detected in both embedders and the
reviewer's in Qwen3; the reviewer's in
BGE-M3 (+0.010, CI +0.001 to +0.019) and
ORACLE's in Qwen3 (+0.012, CI +0.003 to
+0.021) are estimated but not
significant after correction. NEUTRAL's
own change is positive too (+0.0085 and
+0.0090). What feedback adds beyond
revision (E4) is about +0.01 for
distance feedback in both embedders and
not significant after correction (Qwen3
+0.011, CI −0.000 to +0.022; resampling
connected groups of contexts gives
+0.0001 to +0.021; BGE-M3 +0.011, CI
+0.002 to +0.021); the embedders do not
differ (+0.001, CI −0.009 to +0.011),
and in the discovery sample the same
contrast was −0.003 in Qwen3 and +0.010
in BGE-M3. Pooling both samples
(exploratory) gives +0.004 (CI −0.004 to
+0.011) in Qwen3 and +0.011 (CI +0.004
to +0.018) in BGE-M3. Adjusted for step
change, after removing the genre
direction, or in units of the gap, the
reading does not move towards detection:
E4 is +0.008 (CI −0.004 to +0.019)
adjusted for step, and +0.010 (CI −0.001
to +0.020) in BGE-M3 without the genre
direction. Seventeen of these 150
contexts are also in the OLMo test below
(an earlier version of this paper
reported none); on the 133 others only
one reading changes, ORACLE's in BGE-M3,
from within ±0.015 to insufficient
(+0.006, CI −0.002 to +0.014).

\textbf{A positive control} (registered
follow-up). Without a positive control
the nulls above could come from a
measure that registers no steering at
all. HINT reveals, at each revision, the
title of one real follow-up per context,
held out at random and removed from the
construction of \(\mathbf{s}\) (gpt-4o,
the 150 replication contexts, 900
calls); the other arms are rescored on
the same held-out direction.
\hyperref[tab:7]{Table~7} gives the
results.

\begin{table}[tbp]
\centering\footnotesize\setlength{\tabcolsep}{4pt}
\caption[]{\textbf{Table 7.} Positive control on the 150 replication contexts (gpt-4o): change from round 0 to round 2, arm minus NEUTRAL, along the direction \(\mathbf{s}\) rebuilt without the hinted paper (95 \% t intervals; Holm over the three PC tests), and the share of HINT's turn that each feedback arm achieves beyond revision (bootstrap intervals). ORACLE-C scores the proposal against the fixed centroid of the follow-ups instead of the nearest one.}\label{tab:7}
\begin{tabular}{@{}P{\dimexpr 0.420\linewidth-2\tabcolsep\relax}P{\dimexpr 0.290\linewidth-2\tabcolsep\relax}P{\dimexpr 0.290\linewidth-2\tabcolsep\relax}@{}}
\toprule
Quantity & Qwen3 & BGE-M3 \\
\midrule
Manipulation check: HINT − NEUTRAL, cosine to the hinted paper & +0.227 {[}0.206,~0.248{]} & +0.183 {[}0.165,~0.200{]} \\
PC1: HINT − NEUTRAL along \(\mathbf{s}\) & +0.032 {[}0.018,~0.046{]}, Holm 0.0003 & +0.024 {[}0.012,~0.036{]}, Holm 0.0003 \\
PC2: HINT − GEO along \(\mathbf{s}\) & +0.021 {[}0.007,~0.036{]}, Holm 0.018 & +0.014 {[}0.001,~0.026{]}, Holm 0.074 \\
PC3: HINT − ORACLE along \(\mathbf{s}\) & +0.029 {[}0.014,~0.045{]}, Holm 0.002 & +0.031 {[}0.017,~0.044{]}, Holm 0.0001 \\
GEO beyond revision, share of HINT's turn & 0.34 {[}0.00,~0.70{]} & 0.43 {[}0.04,~0.90{]} \\
Reviewer beyond revision, share of HINT's turn & 0.13 {[}−0.32,~0.46{]} & −0.02 {[}−0.68,~0.40{]} \\
ORACLE beyond revision, share of HINT's turn & 0.08 {[}−0.40,~0.46{]} & −0.28 {[}−1.11,~0.17{]} \\
ORACLE-C − NEUTRAL along \(\mathbf{s}\) & −0.005 {[}−0.014,~+0.005{]} & −0.008 {[}−0.016,~−0.000{]} \\
\bottomrule
\end{tabular}
\end{table}

Revealing the title of one real
follow-up turns gpt-4o's ideas along
\(\mathbf{s}\) by +0.032 beyond revision
in two rounds (+0.024 in BGE-M3), so the
measure registers informed steering;
that turn is only about twice the ±0.015
used to read the nulls, so the design
separates ``no turn'' from about half of
a revealed title's turn and no finer. It
moves them far more towards the hinted
paper itself (+0.227) than along the
direction the other follow-ups share,
and the turn closes about 30 \% of the
hinted paper's own gap. Against that
yardstick, the reviewer and the oracle
turn the ideas, beyond revision, by at
most 0.46 of what the title does (Qwen3;
0.40 and 0.17 in BGE-M3), so ``barely''
holds for them with a denominator;
distance feedback adds 0.34 of the
title's turn (CI 0.00 to 0.70; 0.43, CI
0.04 to 0.90, in BGE-M3), which is not
``barely''. Content beats distance
feedback in Qwen3 (PC2) but not
detectably in BGE-M3, so I do not claim
that content steers more than distance
feedback in general. The oracle's
failure is not an artefact of scoring
against the nearest follow-up, whose
identity changes in about 40 \% of
revisions (slightly more often than
under plain revision): ORACLE-C, scored
against a fixed centroid, does not raise
its own score beyond NEUTRAL (−0.003, CI
−0.008 to +0.002) and turns the ideas no
further along \(\mathbf{s}\) (−0.005; in
BGE-M3 −0.008, CI −0.016 to −0.000, Holm
\emph{p} = 0.017). The failure is of two
rounds of scalar feedback, not of the
maximum.

\textbf{A confirmatory test with OLMo}
(registered follow-up). In the discovery
sample OLMo's turns were larger (+0.012
to +0.032 along \(\mathbf{s}\), the
reviewer's surviving Holm) and its own
NEUTRAL arm, run afterwards on the same
round-0 ideas, reproduced a third to a
half of them in Qwen3 and two thirds or
more in BGE-M3 (+0.012 and +0.019), with
no critique arm differing from NEUTRAL
after Holm but no interval inside ±0.015
(\hyperref[app:A.8]{Appendix~A.8}).
OLMo's feedback-specific turn therefore
got its own study, with two changes of
design: the direction is built from the
papers that \emph{continue} the source
rather than from all citing papers, and
each arm gets three ideas per context.
The contexts come from a new frame of
2021--2022 arXiv cs papers with at least
five arXiv follow-ups from 2024 on,
disjoint from the original 150; 17 of
them are also in the gpt-4o replication.
The two LLM labellers labelled follow-up
pairs by a registered hybrid rule (pairs
that Semantic Scholar flags first, then
the rest of a context only if one of its
flagged pairs is a consensus
continuation; κ = 0.69 over 13,074
pairs). A context is eligible if it has
a flagged consensus continuation and at
least two consensus continuations in
all: 150 of the 1,800 labelled (8.3 \%),
so the population is contexts with a
flagged continuation, not all contexts
with continuations (a 10 \% audit
labelled in full found no eligible
context missed). For these,
\(\mathbf{s}\) is rebuilt from the
continuations; 83 contexts have exactly
two. OLMo ran GEO, the reviewer, ORACLE
and NEUTRAL, 3 ideas and 3 rounds each,
in two waves under a stopping rule that
read only interval widths. The primary
contrast is each critique arm's change
along \(\mathbf{s}\) minus NEUTRAL's, in
Qwen3, read against ±0.015
(\hyperref[tab:8]{Table~8}).

\begin{table}[tbp]
\centering\footnotesize\setlength{\tabcolsep}{4pt}
\caption[]{\textbf{Table 8.} OLMo confirmatory test on 150 contexts with continuations: critique arm minus NEUTRAL along the continuations' direction \(\mathbf{s}\), change from round 0 to round 2 (Qwen3 unless stated; registered bootstrap intervals in the first two rows, 95 \% intervals otherwise). The last row gives each arm's own change; NEUTRAL's is +0.006 {[}−0.004, +0.016{]}.}\label{tab:8}
\begin{tabular}{@{}P{\dimexpr 0.310\linewidth-2\tabcolsep\relax}P{\dimexpr 0.230\linewidth-2\tabcolsep\relax}P{\dimexpr 0.230\linewidth-2\tabcolsep\relax}P{\dimexpr 0.230\linewidth-2\tabcolsep\relax}@{}}
\toprule
Contrast & GEO & Reviewer & ORACLE \\
\midrule
Registered estimand & −0.002 {[}−0.013,~+0.009{]} & −0.003 {[}−0.014,~+0.008{]} & +0.010 {[}−0.004,~+0.024{]} \\
BGE-M3 & −0.000 {[}−0.010,~+0.009{]} & −0.000 {[}−0.011,~+0.010{]} & +0.001 {[}−0.010,~+0.013{]} \\
133 contexts not in the gpt-4o replication & −0.003 {[}−0.015,~+0.009{]} & −0.003 {[}−0.015,~+0.010{]} & +0.006 {[}−0.008,~+0.021{]} \\
Adjusted for step change & −0.006 {[}−0.018,~+0.006{]} & −0.004 {[}−0.016,~+0.007{]} & +0.010 {[}−0.004,~+0.024{]} \\
Genre direction removed & −0.004 {[}−0.015,~+0.008{]} & −0.004 {[}−0.015,~+0.008{]} & +0.009 {[}−0.006,~+0.023{]} \\
Corrected for the reliability of \(\mathbf{s}\) & −0.004 {[}−0.021,~+0.013{]} & −0.007 {[}−0.024,~+0.011{]} & +0.014 {[}−0.007,~+0.035{]} \\
Arm's own change along \(\mathbf{s}\) (Holm over four) & +0.004 {[}−0.008,~+0.016{]}~(1) & +0.003 {[}−0.008,~+0.014{]}~(1) & +0.016 {[}+0.004,~+0.027{]}~(0.076) \\
\bottomrule
\end{tabular}
\end{table}

On the registered estimand GEO and the
reviewer show no feedback-specific turn
of 0.015 or more (the reviewer's t
interval, −0.0146 to +0.0087, clears the
bound by 0.0004), and ORACLE's is
insufficient (Holm \emph{p} = 0.69).
That reading does not survive the
registered checks. A direction built
from a median of two continuations is
noisy: its split-half reliability on the
46 contexts with at least four is λ =
0.30 (CI 0.25 to 0.36), and corrected
for it the intervals exclude only turns
of about 0.024 or more. Adjusted for
step change, after removing the genre
direction, or on the 133 contexts not
shared with the gpt-4o replication,
GEO's and the reviewer's contrasts
become insufficient. The bound itself is
small against the continuations' own
leave-one-out projection on
\(\mathbf{s}\) (0.267; ±0.015 is 0.056
of it). Of the four per-arm changes,
ORACLE's, the arm with future
information, is the largest and the only
one whose interval excludes zero
(uncorrected \emph{p} = 0.019, Holm over
four 0.076). And the positive control
fails on these contexts: revealing a
held-out follow-up's title moves OLMo's
ideas towards that paper (+0.122, CI
0.102 to 0.141) but not detectably along
\(\mathbf{s}\) (+0.007, CI −0.004 to
+0.019; +0.006 in BGE-M3). On OLMo the
measure does not register even a
title-driven turn, so the OLMo result
cannot be read as ``feedback does not
steer OLMo''; no feedback-specific turn
was detected, and none is bounded.
Critique still lengthens OLMo's step
beyond revision (GEO +0.040, reviewer
+0.015, Holm \emph{p} \textless{} 0.001
and 0.003), GEO's extra move goes
towards the nearest later papers in
general (+0.031 along \(\mathbf{p}\)),
and the step gap is larger for
more-cited contexts in this sample
(Spearman 0.36, CI 0.21 to 0.50), so
these estimates describe well-cited
papers.

\subsection{A 2026 generator, and score feedback with history (pre-registered)}\label{sec:5.4}

Two questions were left open: whether
the pattern belongs to 2023-era models,
and whether the single-score oracle
fails only because it has no memory. I
pre-registered both before their test
runs (\hyperref[app:B]{Appendix~B} lists
the smoke calls made beforehand). For
the first, gpt-5.4-mini (released March
2026; training cutoff reported as 31
August 2025 by third-party listings,
with no official date published)
continues 150 arXiv papers from January
to June 2025, the first 150 in a seeded
order with at least five follow-ups
citing them from September 2025 on
(2,331 follow-ups, at most 50 per
context). For the second, a new arm,
ORACLE-H, shows the model all its
earlier proposals with their similarity
to the real follow-ups and revises from
the best one so far; it runs with gpt-4o
on the original contexts and with
gpt-5.4-mini on the new ones.

\begin{table}[tbp]
\centering\footnotesize\setlength{\tabcolsep}{4pt}
\caption[]{\textbf{Table 9.} Modern generator and ORACLE with history (Qwen3; change from round 0 to round 2, mean over 150 contexts, Wilcoxon \emph{p} in parentheses).}\label{tab:9}
\begin{tabular}{@{}P{\dimexpr 0.280\linewidth-2\tabcolsep\relax}P{\dimexpr 0.130\linewidth-2\tabcolsep\relax}P{\dimexpr 0.130\linewidth-2\tabcolsep\relax}P{\dimexpr 0.240\linewidth-2\tabcolsep\relax}P{\dimexpr 0.220\linewidth-2\tabcolsep\relax}@{}}
\toprule
Generator / arm & Step lengthened & Fit change & Alignment change & Follow-up-specific \\
\midrule
gpt-4o / ORACLE & 0.032 & −0.012 & +0.014~(0.0052) & −0.003~(0.52) \\
gpt-4o / ORACLE-H & 0.023 & −0.008 & +0.018~(5~×~10⁻⁶) & +0.005~(0.13) \\
gpt-5.4-mini / GEO & 0.061 & −0.021 & +0.025~(6~×~10⁻⁵) & −0.003~(0.37) \\
gpt-5.4-mini / Reviewer & 0.050 & −0.022 & +0.0035~(0.40) & −0.004~(0.59) \\
gpt-5.4-mini / ORACLE & 0.008 & −0.008 & −0.022~(2~×~10⁻⁵) & −0.006~(0.19) \\
gpt-5.4-mini / ORACLE-H & 0.012 & −0.007 & −0.012~(0.03) & +0.001~(0.96) \\
\bottomrule
\end{tabular}
\end{table}

The short step replicates with the 2026
model and is larger: its one-shot ideas
sit 0.211 closer to the context than the
real follow-ups (0.198 with memory of
its earlier titles; \emph{p} ≈ 10⁻²⁶),
against 0.149 for gpt-4o (a difference
of 0.062, CI 0.046 to 0.077, on
different contexts; M1 holds). Critique
lengthens the step (GEO +0.061; M2
holds). The single-score oracle does not
detectably change the follow-up-specific
direction (−0.006, CI −0.016 to +0.004,
\emph{p} = 0.19; M3, which predicted no
rise, holds) and lowers the overall
alignment. M1 to M3 replicate in BGE-M3
(step gap 0.171; GEO +0.041; ORACLE
−0.006, CI −0.013 to +0.002). History
adds little: ORACLE-H's fit gain over
ORACLE is +0.004 for gpt-4o (CI −0.003
to +0.011, one-sided \emph{p} = 0.20, so
the pre-registered H1 fails in Qwen3; in
BGE-M3 the same gain, +0.0045, is
significant, CI +0.0002 to +0.0086, Holm
\emph{p} = 0.027) and +0.001 for
gpt-5.4-mini (CI −0.004 to +0.005,
\emph{p} = 0.48; +0.002 in BGE-M3), so
any gain from the history is below 0.011
and 0.005 in cosine, well below the 0.07
ceiling that C5 set for ORACLE's own
gain. With gpt-5.4-mini, in Qwen3, even
GEO moves the direction only towards
nearby papers in general (+0.025), not
towards the follow-ups specifically
(−0.003, CI −0.015 to +0.010); in BGE-M3
GEO's follow-up-specific change is
+0.016 (CI +0.005 to +0.027, uncorrected
\emph{p} = 0.009).

The 95 \% intervals of every null cell
in \hyperref[tab:9]{Table~9} lie within
±0.016, but for gpt-5.4-mini ±0.015 is a
third of the gap between its round-0
ideas and the follow-ups along
\(\mathbf{s}\) (0.34 in Qwen3, 0.25 in
BGE-M3), because its ideas start closer;
its nulls exclude only turns larger than
about a third of that gap. No positive
control was run for this generator.

\textbf{A neutral control for the 2026
generator} (registered follow-up). The
NEUTRAL arm of
\hyperref[sec:5.2]{Section~5.2}, run
with gpt-5.4-mini from the same round-0
ideas as GEO (900 calls) and tested with
the same 14 comparisons, Holm in each
embedder. Revising with no information
lengthens gpt-5.4-mini's step by 0.021
(BGE-M3 0.016) and produces no detected
follow-up-specific change (−0.007, CI
−0.015 to +0.001, \emph{p} = 0.17;
+0.0004, CI −0.007 to +0.008, in
BGE-M3). GEO and the reviewer lengthen
the step beyond it (N1, N2: +0.040 and
+0.029, Holm \emph{p} \textless{} 0.001;
+0.025 and +0.026 in BGE-M3) and lose
more fit (N3, both in Qwen3, the
reviewer only in BGE-M3); GEO also loses
more hits (N4, −0.12, Qwen3 only). GEO's
ideas raise tangent alignment more than
NEUTRAL's (N5, +0.029; +0.034 in BGE-M3;
Holm \emph{p} \textless{} 0.001), mostly
towards nearby later papers in general
(+0.036, exploratory), while ORACLE
raises it less (−0.018, Holm \emph{p} =
0.011; −0.034). Along the
follow-up-specific direction no critique
arm differs from NEUTRAL after Holm (N6,
Qwen3, registered bootstrap intervals:
GEO +0.0045, CI −0.007 to +0.017;
reviewer +0.003, CI −0.009 to +0.014;
ORACLE +0.001, CI −0.009 to +0.012); in
BGE-M3 GEO's extra change is +0.016 (CI
+0.005 to +0.026, uncorrected \emph{p} =
0.02, Holm \emph{p} = 0.16), so distance
feedback may close about a quarter of
this generator's gap beyond revision
there (0.27, CI 0.10 to 0.44). Critique
lengthens the step beyond revision in
all three generators; a
follow-up-specific turn beyond revision
was not detected after correction in any
of them.

\section{Inspiration distance: the anticipation margin fails its validation (pre-registered)}\label{sec:6}

\hyperref[sec:3]{Sections~3} to
\hyperref[sec:5]{5} show the model
material of the same kind as the target.
Someone with ideas also has around them
material they never think of as
relevant: a recipe, the rules of a
sport, a random encyclopaedia page.
Design cognition has long disagreed
about how far such inspiration should be
(\hyperref[sec:7]{Section~7}), and none
of that work scores the ideas against
what was released later. I registered
such a test; its main result is
methodological. The measure I registered
for anticipation separates real future
games from generated ideas mainly by
their format, so the test cannot say
whether distance helps. The heading
departs from the one registered for this
outcome (``No effect of inspiration
distance detected on a margin that
tracks format more than content''),
because \hyperref[tab:10]{Table~10}
shows small positive slopes in Qwen3 and
because the finding that stands is the
measure's failure.
\hyperref[app:C.2]{Appendix~C.2} gives
the design and every result in full.

\textbf{Design in brief.} From a pool of
49,531 items dated before 2024 from nine
sources (past Steam games, board games,
film plots, sports rules, arXiv papers,
patents, news, recipes and random
Wikipedia articles) I drew 40 stimuli
per source and quintile of distance,
where distance is isolation from the
Steam past (1,800 stimuli). gpt-4o
proposed one Steam game per stimulus
under an incidental and an instructed
framing; 200 ideas with no material
(NONE) and 200 real 2023 games from
high-momentum cells (MOMENTUM) are the
baselines. An idea's \textbf{margin} is
its maximum cosine to 32,322 Steam games
whose store pages were probably created
after the cutoff, minus its maximum
cosine to an equal-sized sample of past
games, so a positive margin means nearer
to what came next than to what existed;
its \textbf{uptake} is its cosine to its
own stimulus minus its mean cosine to
other stimuli of the same source and
quintile. Five predictions were
registered
(\hyperref[tab:10]{Table~10}): P1, the
within-source slope of margin on
distance is not positive, with an
inverted U tested by the two-lines test
(Simonsohn,
\hyperlink{ref-Simonsohn2018}{2018});
P2, source explains more variance in
margin than distance; P3, isolation
rises with distance (the manipulation
check); P4, instruction raises uptake;
P5, no source × quintile cell beats both
baselines.

\textbf{The margin fails its validation}
(registered follow-up). Every effect in
\hyperref[tab:10]{Table~10} is a few
thousandths of margin, so the margin
needs a positive control. Held-out real
post-cutoff games score 0.021 above no
material (CI 0.013 to 0.030; 0.017 in
BGE-M3). When gpt-4o rewrites each of
them in the idea format, keeping its
content, the difference falls to 0.009
in Qwen3 (CI 0.004 to 0.014, Holm
\emph{p} = 0.002) and to 0.002 in BGE-M3
(CI −0.001 to +0.006, Holm \emph{p} =
0.094). Under the registered rule this
is outcome F: the margin separates real
future games from ideas mainly by
format, and in BGE-M3 it does not
separate a future game's concept,
written as an idea, from no material at
all. Ten of the 400 rewrites were
blocked by the provider's content filter
and are dropped as registered;
\hyperref[app:C.2]{Appendix~C.2} gives
the sensitivity with them filled in by
another model, which is not a neutral
substitute. Every distance result below
is therefore on a proxy not shown to
measure anticipation of content.

\begin{figure}[!htb]
\centering
\includegraphics[width=0.92\linewidth]{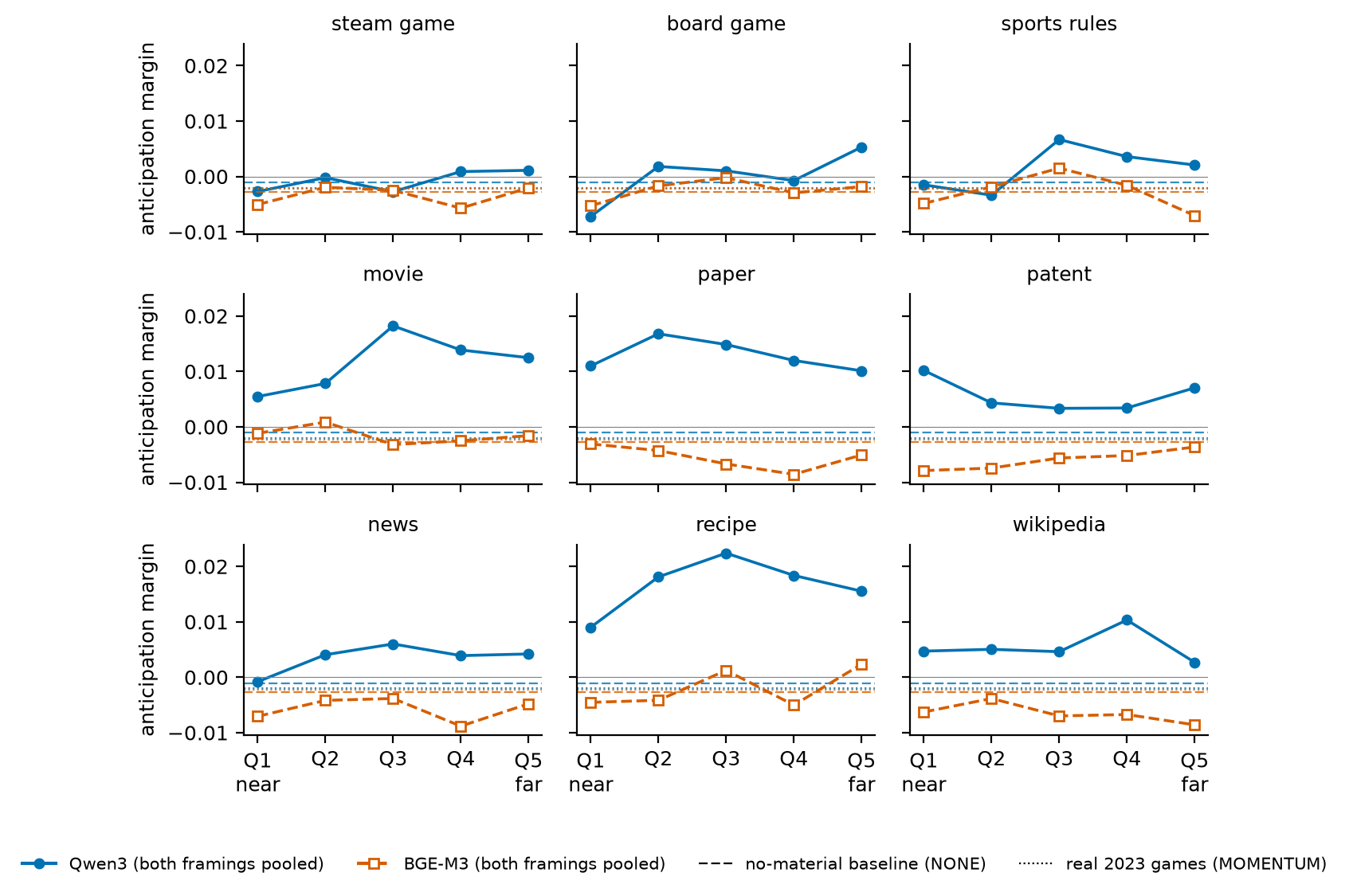}
\caption[]{\textbf{Figure 5.} Anticipation margin by source and within-source distance quintile (gpt-4o, 3,600 ideas from stimuli; 80 per point, both framings pooled; quintiles re-ranked among the drawn stimuli in each embedder, which in BGE-M3 differs from the frozen P2/P5 estimand). One panel per source; solid line with circles: Qwen3, dashed line with squares: BGE-M3; thin horizontal lines in the same style: the no-material baseline (long dashes) and the real-2023-games baseline (dots) of each embedder. Some sources peak in the middle in Qwen3 (recipes, films), but the pre-registered two-lines test finds no inverted U, and neither those peaks nor the ordering of sources survives in BGE-M3.}\label{fig:5}
\end{figure}

\textbf{What is left.} No inverted U was
detected in either embedder or framing
(\hyperref[fig:5]{Figure~5}). That is a
failure to detect, not a bound: the best
bound on a broad middle optimum is 1.1
to 3.9 times the whole margin of a
rewritten future game, so it excludes
nothing the margin can be shown to
measure. P1 fails in Qwen3, where a
farther stimulus gives a margin 0.004 to
0.005 higher (Holm one-sided \emph{p}
0.034 and 0.008; the incidental slope's
two-sided interval includes zero), and
holds in BGE-M3; the Qwen3 slopes lose
significance against a format-matched
reference set and when the idea's own
isolation is controlled. What distance
reliably does is the manipulation it is:
farther material yields more isolated
ideas (P3) and, under instruction, less
coherent ones, and being told to use the
material brings the idea closer to it
(P4). One within-measure contrast holds
in a registered run with new ideas: told
to build on a past game, gpt-4o scores
below no material (by 0.0052 in Qwen3,
CI 0.0032 to 0.0073, and by 0.0040 in
BGE-M3, CI 0.0025 to 0.0055). Both arms
are ideas written by the same model in
the same format, which is what the
margin can still compare. A stress test
with two later command-line agents is in
\hyperref[app:C.3]{Appendix~C.3}.

\begin{table}[tbp]
\centering\footnotesize\setlength{\tabcolsep}{4pt}
\caption[]{\textbf{Table 10.} Inspiration from anything (gpt-4o, 3,600 ideas from stimuli and 200 without). The margin failed its validation (see text), so these are outcomes on a proxy. Within-source slopes on the distance rank (0 = nearest, 1 = farthest), with Holm-adjusted one-sided \emph{p} in parentheses, and outcome of each prediction; 95 \% intervals of the P1 slope in brackets. The isolation-controlled row is exploratory.}\label{tab:10}
\begin{tabular}{@{}P{\dimexpr 0.200\linewidth-2\tabcolsep\relax}P{\dimexpr 0.200\linewidth-2\tabcolsep\relax}P{\dimexpr 0.200\linewidth-2\tabcolsep\relax}P{\dimexpr 0.200\linewidth-2\tabcolsep\relax}P{\dimexpr 0.200\linewidth-2\tabcolsep\relax}@{}}
\toprule
 & Qwen3 INCIDENTAL & Qwen3 INSTRUCTED & BGE-M3 INCIDENTAL & BGE-M3 INSTRUCTED \\
\midrule
P1 slope of margin & +0.0038~(0.034): \textbf{fails} & +0.0053~(0.008): \textbf{fails} & +0.0015~(0.13): holds & −0.0008~(0.72): holds \\
P1, two-sided 95 \% interval & {[}−0.0003,~+0.0079{]} & {[}+0.0010,~+0.0096{]} & {[}−0.0011,~+0.0042{]} & {[}−0.0033,~+0.0018{]} \\
P1 with isolation controlled (exploratory) & +0.0015 {[}−0.0025,~+0.0056{]} & +0.0027 {[}−0.0016,~+0.0070{]} & −0.0014 {[}−0.0040,~+0.0012{]} & −0.0033 {[}−0.0058,~−0.0007{]} \\
Inverted U (two-lines) & no & no & no & no \\
P2 η² source vs quintile & 0.041~vs~0.004: holds & 0.060~vs~0.007: holds & 0.017~vs~0.007: fails & 0.016~vs~0.008: fails \\
P3 slope of isolation & +0.022 (\textless{} 10⁻⁸): holds & +0.024 (\textless{} 10⁻⁹): holds & +0.023 (\textless{} 10⁻¹⁵): holds & +0.022 (\textless{} 10⁻¹⁴): holds \\
P4 uptake, instructed vs incidental & 0.218~vs~0.193~(10⁻²⁶): holds &  & 0.163~vs~0.149~(10⁻¹⁵): holds &  \\
P5 cells above both baselines & 6 of 45: fails &  & 0 of 45: holds &  \\
\bottomrule
\end{tabular}
\end{table}

\section{Related work}\label{sec:7}

\textbf{Continuing a given paper.} Tang
and Yang
(\hyperlink{ref-Tang2026}{2026}) showed
at scale that LLM research ideas stay
closer to their seed literature than the
human work that follows: over 219,655
ideas from five agent frameworks and
five LLMs, the cosine distance from an
idea to the centroid of its five seed
papers is 0.322, against 0.410 for the
papers that cite those seeds.
\hyperref[sec:5.1]{Section~5.1} is an
independent replication with a
single-paper context and generators
whose training precedes every
follow-up's first arXiv version, which
removes the most direct leakage that
their 2020--2025 seeds and recent models
leave open (earlier dissemination
outside arXiv is not checked), and with
a control for the proposal genre and the
prompt's wording. They also observe,
across frameworks, that the one whose
ideas step furthest (AgentLaboratory,
only 0.027 short of the human
follow-ons) has among the lowest
frontier alignment (0.250 against
0.377), and that literature search does
not substantially improve frontier
alignment: an observational version of
``a longer step is not a better-oriented
one'', which \hyperref[sec:5]{Section~5}
tests with controlled feedback arms, a
content-free control and a positive
control. The closest feedback-loop work
is Liu et
al.~(\hyperlink{ref-Liu2026b}{2026b}): a
five-turn ideator-judge dialogue keeps
pseudo-papers 12--17 \% below real
papers in orthogonal novelty at every
turn, and a ``wild'' judge raises
distance but lowers orthogonal novelty.
They measure against same-topic
literature, without a temporal holdout
or a score towards the realised future.
SciMON's novelty boosting (Wang et al.,
\hyperlink{ref-Wang2024}{2024}) makes
ideas more different from the literature
without testing whether they match a
real target better. MOOSE-Chem (Yang et
al., \hyperlink{ref-Yang2025}{2025}) is
the counterpoint: its
feedback-and-recombination unit raises
the judged match to a rediscovered
chemistry hypothesis (Matched Score
2.578 without it, 2.730 with it), but
its inspirations are retrieved from a
corpus that contains the ground-truth
ones, and the effect mixes feedback with
mutation, recombination and ranking.
Reviewer-loop systems (ResearchAgent,
Baek et al.,
\hyperlink{ref-Baek2025}{2025}; The AI
Scientist, Lu et al.,
\hyperlink{ref-Lu2024}{2024}; Chain of
Ideas, Li et al.,
\hyperlink{ref-Li2025}{2025}) never
check refined ideas against later
papers. I found no inference-time score
feedback towards the realised future in
prior work; Wang et
al.~(\hyperlink{ref-Wang2026}{2026})
train on a future-alignment reward, the
trained counterpart.

\textbf{Benchmarks that score ideas
against the future.} HindSight (Jiang,
\hyperlink{ref-Jiang2026}{2026}) matches
generated ideas to later papers and
finds LLM-judged novelty negatively
correlated with future realisation (ρ =
−0.29). PreScience (Ajith et al.,
\hyperlink{ref-Ajith2026}{2026})
generates a paper from its influential
references, the reverse direction of
mine; IdeaForecastBench (Li et al.,
\hyperlink{ref-Li2026a}{2026a}),
future-aligned proposals (Wang et al.,
\hyperlink{ref-Wang2026}{2026}),
GIScholarBench (Li et al.,
\hyperlink{ref-Li2026b}{2026b}) and
Kumar et
al.~(\hyperlink{ref-Kumar2025}{2025})
score generated directions against later
or future-work text. None measures the
step or conditions on it.

\textbf{Does the future fill the gaps?}
Luo and Evans
(\hyperlink{ref-Luo2026}{2026}) embed
OpenAlex concepts per discipline and
year, find holes with persistent
homology and track which later works
fill them: larger holes are less often
closed, and only null models that
combine concepts already close reproduce
the fill rates. That the future fills
small, nearby holes is their conclusion,
in concept space, and my F2 agrees with
it for most valid gap scores; what I add
is the item-level map the recipe uses,
its literal far region, a validity check
of the gap scores, and a head-to-head
with momentum. A line of work on how
science moves relative to its crowd
bears on the same question from the
other side. Rzhetsky et
al.~(\hyperlink{ref-Rzhetsky2015}{2015})
show that biomedical research
concentrates on already-dense regions of
a knowledge network, more conservatively
than would maximise discovery. Shi and
Evans (\hyperlink{ref-Shi2023}{2023})
find that surprising combinations of
contents and contexts go with higher
impact and tend to come from outsiders
in distant fields, and Sourati and Evans
(\hyperlink{ref-Sourati2023}{2023})
predict future discoveries better when
they model which scientists are
positioned to make them, and generate
plausible ``alien'' hypotheses by
avoiding those regions. Their emptiness
is relative to people and communities,
not to an embedding of texts; my results
are about the latter. Patent ``vacuum''
maps (Son et al.,
\hyperlink{ref-Son2012}{2012}) were not
validated against later filings. Link
prediction on concept networks
(Science4Cast, Krenn et al.,
\hyperlink{ref-Krenn2023}{2023};
Impact4Cast, Gu and Krenn,
\hyperlink{ref-Gu2025}{2025}) already
found degree and its time derivatives
the strongest features, so that momentum
forecasts growth is expected; I use it
as the bar a gap score must clear. The
Alien Space of Science (Artiles et al.,
\hyperlink{ref-Artiles2026}{2026})
reports no momentum or popularity
baseline; \hyperref[sec:4]{Section~4}
runs that comparison with cheap proxies.

\textbf{Filling the empty cells of a
map.} Quality-diversity search is the
nearest methodological family to ``fill
the empty cells of a map'': MAP-Elites
(Mouret and Clune,
\hyperlink{ref-Mouret2015}{2015}) keeps
the best solution in each cell of a
behaviour space, and with language
models as mutation operators (Lehman et
al., \hyperlink{ref-Lehman2022}{2022}),
as judges of quality and diversity
(Bradley et al.,
\hyperlink{ref-Bradley2024}{2024}) or as
models of what humans find interesting
(Zhang et al.,
\hyperlink{ref-Zhang2024}{2024}), it
explores open-ended spaces of text and
tasks. These methods define the cells by
chosen descriptors and judge the result
by quality within each cell; none asks
whether the empty cells of a map built
from past items are where later items
appear, which is the assumption the
recipe adds. Novelty search (Lehman and
Stanley,
\hyperlink{ref-Lehman2011}{2011}) drops
the objective and rewards behaviour
unlike any seen before, and atypical
combinations of prior work mark
high-impact science (Uzzi et al.,
\hyperlink{ref-Uzzi2013}{2013}). Both
measure the novelty of what is produced,
not its relation to what came next.

\textbf{Near and far inspiration.}
Design cognition has tested the distance
of inspiration with human designers for
decades, with three verdicts: far
sources raise novelty (Chan et al.,
\hyperlink{ref-Chan2011}{2011}), near
sources lead to better ideas (Chan, Dow
and Schunn,
\hyperlink{ref-Chan2015}{2015}, on
OpenIDEO, with LDA distance and expert
shortlisting), and moderate distance is
best (Fu et al.,
\hyperlink{ref-Fu2013}{2013}, as a
proposal; Gonçalves et al.,
\hyperlink{ref-Gon�alves2013}{2013},
with three ordinal levels). A
pre-registered test of incidental random
Wikipedia pages found no benefit for
human lateral thinking (Malthouse et
al.,
\hyperlink{ref-Malthouse2022}{2022}).
The closest LLM design is Liu et
al.~(\hyperlink{ref-Liu2026a}{2026a}):
randomly assigned source concepts graded
by embedding distance, gpt-4o among
seven LLMs, with originality rising
monotonically with distance as judged by
lay raters. PaperGym (Choi and Song,
\hyperlink{ref-Choi2026}{2026}) finds
cross-domain seeds no better than random
ones for research ideation. None of them
scores ideas against later releases;
\hyperref[sec:6]{Section~6} does,
crosses source with within-source
distance, and separates incidental from
instructed exposure. Its isolation
result agrees with Liu et
al.~(\hyperlink{ref-Liu2026a}{2026a}) on
originality; its anticipation result
turned out to need a better-validated
measure than the one I used.

\textbf{Decoding embedding points and
the cap.} ELM (Tennenholtz et al.,
\hyperlink{ref-Tennenholtz2024a}{2024a})
and EAGLE (Tennenholtz et al.,
\hyperlink{ref-Tennenholtz2024b}{2024b})
decode or steer towards embedding
points, the second explicitly to fill
``content gaps''; neither evaluates
against later items. Embedding inversion
(vec2text; Morris et al.,
\hyperlink{ref-Morris2023}{2023}) is the
natural stronger decoder; it already
breaks under small Gaussian noise on the
embedding, and my attempt to test it in
a supported encoder was uninformative
because real 32-token fragments do not
pass a coherence gate
(\hyperref[sec:3]{Section~3}). That text
representations occupy a narrow cone is
well established (Ethayarajh,
\hyperlink{ref-Ethayarajh2019}{2019};
Gao et al.,
\hyperlink{ref-Gao2019}{2019}; Mu and
Viswanath, \hyperlink{ref-Mu2018}{2018};
Li et al., \hyperlink{ref-Li2020}{2020};
Godey et al.,
\hyperlink{ref-Godey2024}{2024}), and
post-processing such as removing the
dominant directions (Mu and Viswanath,
\hyperlink{ref-Mu2018}{2018}) or
whitening (Su et al.,
\hyperlink{ref-Su2021}{2021}) removes
most of it; \hyperref[sec:3]{Section~3}
checks that the far-point result
survives both. What I add is a
pre-registered prediction, from corpus
statistics, of the median distance
between far points and the nearest item
of the past corpus. That LLMs return to
a narrow set of ideas is also documented
(Si et al.,
\hyperlink{ref-Si2025}{2025}; Doshi and
Hauser, \hyperlink{ref-Doshi2024}{2024};
Padmakumar and He,
\hyperlink{ref-Padmakumar2024}{2024}).

\section{Limitations}\label{sec:8}

\begin{itemize}
\tightlist
\item
  \textbf{A follow-up is a later citing
  paper, not a continuation.} Labelling
  all 2,883 context/follow-up pairs
  blind from the two abstracts, two LLMs
  (gpt-6.1-sol and Claude Opus 5.5) both
  say the follow-up continues its source
  in 112 (3.9 \%; 4.5 \% and 5.7 \% for
  each labeller alone), spread over 47
  of the 150 contexts. They agree on
  continues-or-not with κ = 0.75 (CI
  0.65 to 0.81), and on the four labels
  with κ = 0.41, most disagreements
  being ``unclear'' against
  ``background''. Semantic Scholar's
  flags enrich for continuations without
  isolating them: 11 \% of flagged
  follow-ups are consensus continuations
  against 2.5 \% of the rest. On the
  original 150 contexts, every
  ``direction towards the follow-ups''
  in \hyperref[sec:5]{Section~5} means
  towards the papers that later cite the
  context; only the OLMo test builds it
  from continuations. Built from the
  continuations alone, that direction
  can be tested in the 17 original
  contexts with at least two consensus
  continuations (registered as
  confirmatory once at least ten had
  them): no gpt-4o critique arm moves
  along it beyond NEUTRAL after Holm
  (ORACLE +0.015, CI −0.016 to +0.047;
  GEO +0.012, CI −0.014 to +0.035;
  reviewer +0.001, CI −0.031 to +0.037),
  and no interval lies within ±0.015.
\item
  \textbf{The labels rest on LLMs and
  one non-expert reader.} In a
  registered blind check I labelled 40
  sampled pairs myself from the same
  abstracts, blind to the LLM labels and
  to the stratum (I am not a domain
  expert in most of these fields): I
  called 24 of 25 consensus
  continuations continuations (96 \%,
  Wilson CI 0.80 to 0.99), none of 7
  consensus background pairs, and 5 of
  the 8 pairs on which the labellers
  split (each labeller matched me on 4).
  I marked 11 of those 24 as less
  certain; counting only the pairs I was
  sure of, the share is 13 of 25 (CI
  0.33 to 0.70; a sensitivity analysis
  not in the registration). The novelty
  judgements of
  \hyperref[app:A.7]{Appendix~A.7}
  likewise rest on LLM judges and my 80
  ratings. Two independent expert
  annotators on a few hundred pairs
  would turn this limitation into a
  result.
\item
  \textbf{One domain and three
  generators for the continuation
  study.} All contexts are arXiv cs
  papers, selected for having at least
  five later arXiv citing papers, which
  makes them better cited than the
  typical paper of their frame. gpt-4o
  and OLMo-7B-0424 predate the
  follow-ups by design; OLMo is smaller
  and runs fewer arms, with one idea per
  context in the discovery sample. The
  2026 generator runs on other contexts,
  and its training cutoff is reported
  only by third parties: if it is later
  than stated, some of its follow-ups
  may be in its training data. The
  one-shot step was tested with three
  wordings of the user prompt, on the
  discovery sample only, and all three
  keep the system prompt that asks for
  ``the next paper''; the genre control
  without the context also ran on the
  discovery sample only.
\item
  \textbf{Direction is measured with a
  radial-invariant measure that
  co-varies with step.} Its rise under
  critique is not separable from the
  longer step with the controls I have
  (\hyperref[sec:5.3]{Section~5.3}); the
  positive control shows that it
  registers an informed turn for gpt-4o,
  and fails on OLMo. The yardsticks are
  a revealed title and the gap to the
  follow-ups' own projection, both
  two-round quantities; a stronger or
  longer intervention could produce a
  larger turn. The equivalence bound of
  ±0.015 was chosen after the discovery
  results and is now read as a fraction
  of the gap.
\item
  \textbf{The oracles are inference-time
  scalar feedback.} ORACLE sees one
  score about the previous proposal,
  ORACLE-C one score against a fixed
  target, ORACLE-H the history and the
  best proposal so far. Their failure
  says nothing about training on a
  future-alignment reward.
\item
  \textbf{Everything is measured in two
  general text encoders.} Two embedders
  give robustness across spaces, not
  independence from embeddings. The
  encoders were trained on text that may
  postdate the splits, and arXiv
  abstracts are latest versions; maps
  are past-only in their items, not in
  their representations. For the
  forecasting study I reran
  \hyperref[sec:4]{Section~4} in a
  TF-IDF/LSA space fitted on past text
  only and, for Steam and arXiv, in
  all-mpnet-base-v2, released before
  those splits, and the registered
  conclusions held, except that with LSA
  the Steam result became inconclusive
  (\hyperref[sec:4]{Section~4}); for
  patents only the LSA space is free of
  this leak, and there the test has
  little power. The other sections were
  not rerun in such spaces.
\item
  \textbf{Replication is partial.}
  \hyperref[tab:A13]{Table~A13}
  (\hyperref[app:C.1]{Appendix~C.1})
  says for each confirmatory result
  whether the second embedder was run
  and whether the result held there; the
  main failures to replicate are F4, P1,
  P2 with the frozen quantiles and P5.
\item
  \textbf{The GEO band leaks an
  aggregate} of the original 150
  contexts; the replication reuses it on
  new contexts.
\item
  \textbf{Inspiration study.} One
  generator for the confirmatory test
  (gpt-4o), one target domain (Steam)
  and one stimulus per idea. The
  store-page date is proxied by the
  appid threshold. The margin was
  validated only after the test and
  tracks format more than content
  (\hyperref[sec:6]{Section~6}). The two
  embedders rank stimuli within a source
  only moderately alike (Spearman
  0.33--0.75), and the content effects
  of the primary space do not replicate
  in the second, which is why they are
  not claimed. The coherence judge is an
  LLM, checked against a second LLM
  only. The stress-test generators were
  trained after the future games
  appeared, so they are read only
  against themselves.
\item
  \textbf{Gaps study.} One decoder
  procedure (eight shown neighbours,
  score-only feedback, 4 × 3 candidates)
  and searches that cannot certify a
  bound; inversion is not covered. LLM
  judges overlap with the generators,
  and their criteria are not independent
  (coherence correlates with feasibility
  at 0.35 to 0.63 and appeal at 0.41 to
  0.79). The decoders were trained after
  the Steam and arXiv splits and the
  patent future is inside every
  generator's training data, so no
  decoding result is read as
  anticipation. The registered hole
  score does not measure emptiness; the
  valid alternatives were defined after
  the main results, with partial
  knowledge of the registered score's
  correlates. The future horizons differ
  by corpus (Steam and arXiv up to the
  2026 snapshot, patents two years). GAP
  and FAR targets exist only in Qwen3.
  The combination test uses proxies (NMF
  topics, PMI, past rarity) on one
  corpus.
\item
  \textbf{Studies reported.} Every
  completed study is reported here,
  including the failed and uninformative
  ones; \hyperref[app:B]{Appendix~B}
  lists all of them with their
  registrations.
\end{itemize}

\section{Conclusion}\label{sec:9}

The recipe's far points are not text: no
later item and no coherent text I could
construct came meaningfully close to one
(raw-space cosine about 0.10 at most for
later items, 0.14 for coherent text, and
out of reach in 22 of 24 centred or
whitened cells), and a decoder prompted
with a target's neighbours writes those
neighbours. Interior gaps differ in
absolute distance: coherent text reaches
them at median cosine 0.57 and 0.47 and
beats the nearest existing item on most
of them; whether such text is a good
idea is not something these tests
measure. As forecasts of where a field,
a catalogue or a patent class grows, gap
scores that do detect empty regions did
no better than chance and added nothing
to momentum in Steam and arXiv, with
patents inconclusive, and the same held
in spaces whose encoder saw no
post-split text (with Steam inconclusive
in one of them), although more isolated
new Steam games were slightly better
reviewed. A model continuing a paper
takes a shorter step than the papers
that later cite it, also with other
wordings of the user prompt, when those
papers are rewritten in its own genre,
and, for OLMo, against citing papers
from just after its training data end;
feedback lengthens that step beyond
content-free revision, but what it adds
in direction is about 0.13 (reviewer)
and 0.08 (oracle) of what revealing one
real follow-up's title does, with upper
bounds of 0.46, and about 0.34 for
distance feedback (upper bound 0.70),
and is not separable from the longer
step. For the distance of inspiration
material, the anticipation margin I
registered failed its validation,
tracking format more than content, so
its null on distance is uninformative;
what remains is that farther material
yields more isolated ideas. These are
results for inference-time methods with
the generators, encoders and corpora I
used, not a proof that no use of a map
can help.

\section*{Use of AI tools}\phantomsection\addcontentsline{toc}{section}{Use of AI tools}

Language models are the object of study
here: they generate, decode, judge and
label, and each is named where it is
used. Separately, the experimental code,
the analyses and the text of this paper
were developed with extensive assistance
from AI coding agents (Anthropic's
Claude Code and OpenAI's Codex), which
also served as adversarial reviewers of
analysis plans, results and drafts; the
two reviews that requested the thirteen
review studies of
\hyperref[sec:3]{Sections~3} to
\hyperref[sec:6]{6} were written by such
agents. The research questions, the
approval of every registration before
its run and the decision of what to
claim are mine. Every number in the
paper is produced by a script in the
repository from saved outputs, and a
provenance file maps each number to the
file it comes from; I am responsible for
all of it.

\section*{Data and code availability}\phantomsection\addcontentsline{toc}{section}{Data and code availability}

Code, pre-registration documents,
prompts and per-experiment results are
available at\\
\url{https://github.com/JaviMaligno/latent-gap-ideas}.
The corpora are public: the Steam Games
Dataset and the arXiv metadata snapshot
on Kaggle, the Harvard USPTO Patent
Dataset, and, for the inspiration pool,
BoardGameGeek game listings and
Wikipedia film plots (both as published
on Kaggle), the English Wikipedia dump
of 2022-03-01, AG News and Food.com
recipes. Follow-up citations, influence
flags and citation intents come from the
Semantic Scholar API. Saved model
outputs are in the repository or its
release.

\section*{References}\phantomsection\addcontentsline{toc}{section}{References}

\begingroup\raggedright\small\setlength{\leftskip}{1.5em}\setlength{\parindent}{-1.5em}\setlength{\parskip}{3pt plus 1pt}

\hspace*{-1.5em}\hypertarget{ref-Ajith2026}{}Ajith,
A., Singh, A., DeYoung, J., Kunievsky,
N., Kozlowski, A. C., Tafjord, O.,
Evans, J., Weld, D. S., Hope, T. and
Downey, D. (2026). PreScience: a dataset
and benchmark for scientific
forecasting.
\href{https://arxiv.org/abs/2602.20459}{arXiv:2602.20459}.

\hypertarget{ref-Artiles2026}{}Artiles,
A. H., Weiss, M., Brinkmann, L., Rahwan,
I., Schölkopf, B., Pal, C., Larochelle,
H., Goyal, A. and Rahaman, N. (2026).
The Alien Space of Science: sampling
coherent but cognitively unavailable
research directions.
\href{https://arxiv.org/abs/2603.01092}{arXiv:2603.01092}.

\hypertarget{ref-Baek2025}{}Baek, J.,
Jauhar, S. K., Cucerzan, S. and Hwang,
S. J. (2025). ResearchAgent: iterative
research idea generation over scientific
literature with large language models.
In \emph{Proceedings of NAACL 2025}.
\href{https://arxiv.org/abs/2404.07738}{arXiv:2404.07738}.

\hypertarget{ref-Bradley2024}{}Bradley,
H., Dai, A., Teufel, H., Zhang, J.,
Oostermeijer, K., Bellagente, M., Clune,
J., Stanley, K., Schott, G. and Lehman,
J. (2024). Quality-diversity through AI
feedback. In \emph{Proceedings of ICLR
2024}.
\href{https://arxiv.org/abs/2310.13032}{arXiv:2310.13032}.

\hypertarget{ref-Bustos2026}{}Bustos, M.
(2026). Steam Games Dataset (Kaggle user
fronkongames). Kaggle.
\url{https://www.kaggle.com/datasets/fronkongames/steam-games-dataset}

\hypertarget{ref-Chan2015}{}Chan, J.,
Dow, S. P. and Schunn, C. D. (2015). Do
the best design ideas (really) come from
conceptually distant sources of
inspiration? \emph{Design Studies} 36,
31--58.
\url{https://doi.org/10.1016/j.destud.2014.08.001}

\hypertarget{ref-Chan2011}{}Chan, J.,
Fu, K., Schunn, C., Cagan, J., Wood, K.
and Kotovsky, K. (2011). On the benefits
and pitfalls of analogies for innovative
design: ideation performance based on
analogical distance, commonness, and
modality of examples. \emph{Journal of
Mechanical Design} 133(8), 081004.
\url{https://doi.org/10.1115/1.4004396}

\hypertarget{ref-Chen2024}{}Chen, J.,
Xiao, S., Zhang, P., Luo, K., Lian, D.
and Liu, Z. (2024). M3-Embedding:
multi-linguality, multi-functionality,
multi-granularity text embeddings
through self-knowledge distillation. In
\emph{Findings of ACL 2024}.
\href{https://arxiv.org/abs/2402.03216}{arXiv:2402.03216}.

\hypertarget{ref-Choi2026}{}Choi, Y. and
Song, M. (2026). Read, grep, and
synthesize: diagnosing cross-domain seed
exposure for LLM research ideation.
\href{https://arxiv.org/abs/2605.11532}{arXiv:2605.11532}.

\hypertarget{ref-Cohan2019}{}Cohan, A.,
Ammar, W., van Zuylen, M. and Cady, F.
(2019). Structural scaffolds for
citation intent classification in
scientific publications. In
\emph{Proceedings of NAACL 2019}.
\href{https://arxiv.org/abs/1904.01608}{arXiv:1904.01608}.

\hypertarget{ref-Cornell2026}{}Cornell
University (2026). arXiv dataset
(metadata snapshot). Kaggle.
\url{https://www.kaggle.com/datasets/Cornell-University/arxiv}

\hypertarget{ref-Doshi2024}{}Doshi, A.
R. and Hauser, O. P. (2024). Generative
AI enhances individual creativity but
reduces the collective diversity of
novel content. \emph{Science Advances}
10(28), eadn5290.
\url{https://doi.org/10.1126/sciadv.adn5290}

\hypertarget{ref-Ethayarajh2019}{}Ethayarajh,
K. (2019). How contextual are
contextualized word representations?
Comparing the geometry of BERT, ELMo,
and GPT-2 embeddings. In
\emph{Proceedings of EMNLP-IJCNLP 2019},
55--65.
\url{https://aclanthology.org/D19-1006/}

\hypertarget{ref-Fu2013}{}Fu, K., Chan,
J., Cagan, J., Kotovsky, K., Schunn, C.
and Wood, K. (2013). The meaning of
``near'' and ``far'': the impact of
structuring design databases and the
effect of distance of analogy on design
output. \emph{Journal of Mechanical
Design} 135(2), 021007.
\url{https://doi.org/10.1115/1.4023158}

\hypertarget{ref-Gao2019}{}Gao, J., He,
D., Tan, X., Qin, T., Wang, L. and Liu,
T.-Y. (2019). Representation
degeneration problem in training natural
language generation models. In
\emph{Proceedings of ICLR 2019}.
\href{https://arxiv.org/abs/1907.12009}{arXiv:1907.12009}.

\hypertarget{ref-Godey2024}{}Godey, N.,
de la Clergerie, É. and Sagot, B.
(2024). Anisotropy is inherent to
self-attention in transformers. In
\emph{Proceedings of EACL 2024}.
\url{https://aclanthology.org/2024.eacl-long.3}

\hypertarget{ref-Gon�alves2013}{}Gonçalves,
M., Cardoso, C. and Badke-Schaub, P.
(2013). Inspiration peak: exploring the
semantic distance between design problem
and textual inspirational stimuli.
\emph{International Journal of Design
Creativity and Innovation} 1(4),
215--232.
\url{https://doi.org/10.1080/21650349.2013.799309}

\hypertarget{ref-Groeneveld2024}{}Groeneveld,
D., Beltagy, I., Walsh, E., Bhagia, A.,
Kinney, R., Tafjord, O., Jha, A. H.,
Ivison, H., et al.~(2024). OLMo:
accelerating the science of language
models. In \emph{Proceedings of ACL
2024}.
\href{https://arxiv.org/abs/2402.00838}{arXiv:2402.00838}.

\hypertarget{ref-Gu2025}{}Gu, X. and
Krenn, M. (2025). Forecasting
high-impact research topics via machine
learning on evolving knowledge graphs.
\emph{Machine Learning: Science and
Technology} 6(2), 025041.
\url{https://doi.org/10.1088/2632-2153/add6ef}

\hypertarget{ref-Jiang2026}{}Jiang, B.
(2026). HindSight: evaluating
LLM-generated research ideas via future
impact.
\href{https://arxiv.org/abs/2603.15164}{arXiv:2603.15164}.

\hypertarget{ref-Kinney2023}{}Kinney,
R., Anastasiades, C., Authur, R.,
Beltagy, I., Bragg, J., Buraczynski, A.,
Cachola, I., Candra, S., et al. (2023).
The Semantic Scholar Open Data Platform.
\href{https://arxiv.org/abs/2301.10140}{arXiv:2301.10140}.

\hypertarget{ref-Krenn2023}{}Krenn, M.,
Buffoni, L., Coutinho, B., Eppel, S.,
Foster, J. G., Gritsevskiy, A., Lee, H.,
Lu, Y., Moutinho, J. P., Sanjabi, N.,
Sonthalia, R., Tran, N. M., Valente, F.,
Xie, Y., Yu, R. and Kopp, M. (2023).
Forecasting the future of artificial
intelligence with machine learning-based
link prediction in an exponentially
growing knowledge network. \emph{Nature
Machine Intelligence} 5(11), 1326--1335.
\url{https://doi.org/10.1038/s42256-023-00735-0}

\hypertarget{ref-Kumar2025}{}Kumar, S.,
Ghosal, T., Goyal, V. and Ekbal, A.
(2025). Can large language models unlock
novel scientific research ideas? In
\emph{Proceedings of EMNLP 2025}.
\href{https://arxiv.org/abs/2409.06185}{arXiv:2409.06185}.

\hypertarget{ref-Lehman2011}{}Lehman, J.
and Stanley, K. O. (2011). Abandoning
objectives: evolution through the search
for novelty alone. \emph{Evolutionary
Computation} 19(2), 189--223.
\url{https://doi.org/10.1162/EVCO_a_00025}

\hypertarget{ref-Lehman2022}{}Lehman,
J., Gordon, J., Jain, S., Ndousse, K.,
Yeh, C. and Stanley, K. O. (2022).
Evolution through large models.
\href{https://arxiv.org/abs/2206.08896}{arXiv:2206.08896}.

\hypertarget{ref-Li2020}{}Li, B., Zhou,
H., He, J., Wang, M., Yang, Y. and Li,
L. (2020). On the sentence embeddings
from pre-trained language models. In
\emph{Proceedings of EMNLP 2020}.
\href{https://arxiv.org/abs/2011.05864}{arXiv:2011.05864}.

\hypertarget{ref-Li2026a}{}Li, F., Tang,
Z., Yu, H., Zhao, Y. and You, J.
(2026a). Can large language models
forecast what researchers study next? In
\emph{Proceedings of EMNLP 2026}.
\href{https://arxiv.org/abs/2609.00747}{arXiv:2609.00747}.

\hypertarget{ref-Li2025}{}Li, L., Xu,
W., Guo, J., Zhao, R., Li, X., Yuan, Y.,
Zhang, B., Jiang, Y., et al.~(2025).
Chain of Ideas: revolutionizing research
via novel idea development with LLM
agents. In \emph{Findings of EMNLP
2025}.
\href{https://arxiv.org/abs/2410.13185}{arXiv:2410.13185}.

\hypertarget{ref-Li2026b}{}Li, Z., Yang,
M., Zou, L., Ma, H., Tian, H., Zhou, S.,
Gong, W., Zhang, K., et al.~(2026b).
GIScholarBench: benchmarking LLM
overconfidence in GIS research.
\href{https://arxiv.org/abs/2606.08036}{arXiv:2606.08036}.

\hypertarget{ref-Liu2026a}{}Liu, Q. E.,
Dubova, M., Conklin, H., Harada, T. and
Griffiths, T. L. (2026a). Assessing the
effect of cross-domain mapping on
creativity in humans and large language
models.
\href{https://arxiv.org/abs/2603.19087}{arXiv:2603.19087}.

\hypertarget{ref-Liu2026b}{}Liu, Z.,
Butler, K., Machlanski, D., Bueno, D.
and Tsaftaris, S. A. (2026b). Large
language models extend known research
but do not create new research
directions.
\href{https://arxiv.org/abs/2610.17439}{arXiv:2610.17439}.

\hypertarget{ref-Lu2024}{}Lu, C., Lu,
C., Lange, R. T., Foerster, J., Clune,
J. and Ha, D. (2024). The AI Scientist:
towards fully automated open-ended
scientific discovery.
\href{https://arxiv.org/abs/2408.06292}{arXiv:2408.06292}.

\hypertarget{ref-Luo2026}{}Luo, J. and
Evans, J. A. (2026). Filling holes in
science draws collective attention, but
most higher-order holes remain
unexplored.
\href{https://arxiv.org/abs/2608.28822}{arXiv:2608.28822}.

\hypertarget{ref-Majumder2019}{}Majumder,
B. P., Li, S., Ni, J. and McAuley, J.
(2019). Generating personalized recipes
from historical user preferences. In
\emph{Proceedings of EMNLP-IJCNLP 2019}.
\href{https://arxiv.org/abs/1909.00105}{arXiv:1909.00105}.

\hypertarget{ref-Malthouse2022}{}Malthouse,
E., Liang, Y., Russell, S. and Hills, T.
T. (2022). The influence of exposure to
randomness on lateral thinking in
divergent, convergent, and creative
search. \emph{Cognition} 218, 104937.
\url{https://doi.org/10.1016/j.cognition.2021.104937}

\hypertarget{ref-Morris2023}{}Morris, J.
X., Kuleshov, V., Shmatikov, V. and
Rush, A. M. (2023). Text embeddings
reveal (almost) as much as text. In
\emph{Proceedings of EMNLP 2023}.
\href{https://arxiv.org/abs/2310.06816}{arXiv:2310.06816}.

\hypertarget{ref-Moura2021}{}de Moura,
L. and Ullrich, S. (2021). The Lean 4
theorem prover and programming language.
In \emph{Automated Deduction, CADE 28},
LNCS 12699, 625--635.
\url{https://doi.org/10.1007/978-3-030-79876-5_37}

\hypertarget{ref-Mouret2015}{}Mouret,
J.-B. and Clune, J. (2015). Illuminating
search spaces by mapping elites.
\href{https://arxiv.org/abs/1504.04909}{arXiv:1504.04909}.

\hypertarget{ref-Mu2018}{}Mu, J. and
Viswanath, P. (2018). All-but-the-top:
simple and effective postprocessing for
word representations. In
\emph{Proceedings of ICLR 2018}.
\href{https://arxiv.org/abs/1702.01417}{arXiv:1702.01417}.

\hypertarget{ref-OpenAI2024}{}OpenAI
(2024). GPT-4o system card.
\href{https://arxiv.org/abs/2410.21276}{arXiv:2410.21276}.

\hypertarget{ref-Padmakumar2024}{}Padmakumar,
V. and He, H. (2024). Does writing with
language models reduce content
diversity? In \emph{Proceedings of ICLR
2024}.
\href{https://arxiv.org/abs/2309.05196}{arXiv:2309.05196}.

\hypertarget{ref-Rzhetsky2015}{}Rzhetsky,
A., Foster, J. G., Foster, I. T. and
Evans, J. A. (2015). Choosing
experiments to accelerate collective
discovery. \emph{Proceedings of the
National Academy of Sciences} 112(47),
14569--14574.
\url{https://doi.org/10.1073/pnas.1509757112}

\hypertarget{ref-Shi2023}{}Shi, F. and
Evans, J. (2023). Surprising
combinations of research contents and
contexts are related to impact and
emerge with scientific outsiders from
distant disciplines. \emph{Nature
Communications} 14, 1641.
\url{https://doi.org/10.1038/s41467-023-36741-4}

\hypertarget{ref-Si2025}{}Si, C., Yang,
D. and Hashimoto, T. (2025). Can LLMs
generate novel research ideas? A
large-scale human study with 100+ NLP
researchers. In \emph{Proceedings of
ICLR 2025}.
\href{https://arxiv.org/abs/2409.04109}{arXiv:2409.04109}.

\hypertarget{ref-Simonsohn2018}{}Simonsohn,
U. (2018). Two lines: a valid
alternative to the invalid testing of
U-shaped relationships with quadratic
regressions. \emph{Advances in Methods
and Practices in Psychological Science}
1(4), 538--555.
\url{https://doi.org/10.1177/2515245918805755}

\hypertarget{ref-Soldaini2024}{}Soldaini,
L., Kinney, R., Bhagia, A., Schwenk, D.,
Atkinson, D., Authur, R., et al.~(2024).
Dolma: an open corpus of three trillion
tokens for language model pretraining
research. In \emph{Proceedings of ACL
2024}.
\href{https://arxiv.org/abs/2402.00159}{arXiv:2402.00159}.

\hypertarget{ref-Son2012}{}Son, C., Suh,
Y., Jeon, J. and Park, Y. (2012).
Development of a GTM-based patent map
for identifying patent vacuums.
\emph{Expert Systems with Applications}
39(3), 2489--2500.
\url{https://doi.org/10.1016/j.eswa.2011.08.101}

\hypertarget{ref-Sourati2023}{}Sourati,
J. and Evans, J. A. (2023). Accelerating
science with human-aware artificial
intelligence. \emph{Nature Human
Behaviour} 7(10), 1682--1696.
\url{https://doi.org/10.1038/s41562-023-01648-z}

\hypertarget{ref-Su2021}{}Su, J., Cao,
J., Liu, W. and Ou, Y. (2021). Whitening
sentence representations for better
semantics and faster retrieval.
\href{https://arxiv.org/abs/2103.15316}{arXiv:2103.15316}.

\hypertarget{ref-Suzgun2023}{}Suzgun,
M., Melas-Kyriazi, L., Sarkar, S. K.,
Kominers, S. D. and Shieber, S. M.
(2023). The Harvard USPTO Patent
Dataset: a large-scale, well-structured,
and multi-purpose corpus of patent
applications. In \emph{Advances in
Neural Information Processing Systems
36, Datasets and Benchmarks Track}.
\href{https://arxiv.org/abs/2207.04043}{arXiv:2207.04043}.

\hypertarget{ref-SyntheticBeef2026}{}SyntheticBeef
(2026). Post on X, 21 October 2026.
\url{https://x.com/SyntheticBeef/status/2112904858718367750}

\hypertarget{ref-Tang2026}{}Tang, Y. and
Yang, Y. (2026). AI research agents
narrow scientific exploration.
\href{https://arxiv.org/abs/2605.27905}{arXiv:2605.27905}.

\hypertarget{ref-Tennenholtz2024a}{}Tennenholtz,
G., Chow, Y., Hsu, C.-W., Jeong, J.,
Shani, L., Tulepbergenov, A.,
Ramachandran, D., Mladenov, M. and
Boutilier, C. (2024a). Demystifying
embedding spaces using large language
models. In \emph{Proceedings of ICLR
2024}.
\href{https://arxiv.org/abs/2310.04475}{arXiv:2310.04475}.

\hypertarget{ref-Tennenholtz2024b}{}Tennenholtz,
G., Chow, Y., Hsu, C.-W., Shani, L.,
Liang, E. and Boutilier, C. (2024b).
Embedding-aligned language models. In
\emph{Advances in Neural Information
Processing Systems 37}.
\href{https://arxiv.org/abs/2406.00024}{arXiv:2406.00024}.

\hypertarget{ref-Uzzi2013}{}Uzzi, B.,
Mukherjee, S., Stringer, M. and Jones,
B. (2013). Atypical combinations and
scientific impact. \emph{Science}
342(6157), 468--472.
\url{https://doi.org/10.1126/science.1240474}

\hypertarget{ref-Valenzuela2015}{}Valenzuela,
M., Ha, V. and Etzioni, O. (2015).
Identifying meaningful citations. In
\emph{Scholarly Big Data: AI
Perspectives, Challenges, and Ideas,
Papers from the 2015 AAAI Workshop},
AAAI Technical Report WS-15-13, 21--26.
\url{https://cdn.aaai.org/ocs/ws/ws0121/10185-46008-1-PB.pdf}

\hypertarget{ref-Wang2026}{}Wang, H.,
Jiang, P., Sun, J., Shi, Z., Yu, H.,
Han, J. and Ji, H. (2026). Learning to
predict future-aligned research
proposals with language models. In
\emph{Findings of EMNLP 2026}.
\href{https://arxiv.org/abs/2603.27146}{arXiv:2603.27146}.

\hypertarget{ref-Wang2024}{}Wang, Q.,
Downey, D., Ji, H. and Hope, T. (2024).
SciMON: scientific inspiration machines
optimized for novelty. In
\emph{Proceedings of ACL 2024}.
\href{https://arxiv.org/abs/2305.14259}{arXiv:2305.14259}.

\hypertarget{ref-Yang2025}{}Yang, Z.,
Liu, W., Gao, B., Xie, T., Li, Y.,
Ouyang, W., Poria, S., Cambria, E. and
Zhou, D. (2025). MOOSE-Chem: large
language models for rediscovering unseen
chemistry scientific hypotheses. In
\emph{Proceedings of ICLR 2025}.
\href{https://arxiv.org/abs/2410.07076}{arXiv:2410.07076}.

\hypertarget{ref-Zhang2024}{}Zhang, J.,
Lehman, J., Stanley, K. and Clune, J.
(2024). OMNI: open-endedness via models
of human notions of interestingness. In
\emph{Proceedings of ICLR 2024}.
\href{https://arxiv.org/abs/2306.01711}{arXiv:2306.01711}.

\hypertarget{ref-Zhang2015}{}Zhang, X.,
Zhao, J. and LeCun, Y. (2015).
Character-level convolutional networks
for text classification. In
\emph{Advances in Neural Information
Processing Systems 28}.
\href{https://arxiv.org/abs/1509.01626}{arXiv:1509.01626}.

\hypertarget{ref-Zhang2025}{}Zhang, Y.,
Li, M., Long, D., Zhang, X., Lin, H.,
Yang, B., Xie, P., Yang, A., et
al.~(2025). Qwen3 Embedding: advancing
text embedding and reranking through
foundation models.
\href{https://arxiv.org/abs/2506.05176}{arXiv:2506.05176}.

\par\endgroup

\clearpage
\appendix
\setappendixprefix
\section{Supplementary material}\label{app:A}

\subsection{Setup of the gaps study}\label{app:A.1}

\begin{table}[!h]
\centering\footnotesize\setlength{\tabcolsep}{4pt}
\caption[]{\textbf{Table A1.} Corpora of the gaps study: past and future splits and the quality signal of each corpus.}\label{tab:A1}
\begin{tabular}{@{}P{\dimexpr 0.300\linewidth-2\tabcolsep\relax}P{\dimexpr 0.270\linewidth-2\tabcolsep\relax}P{\dimexpr 0.190\linewidth-2\tabcolsep\relax}P{\dimexpr 0.240\linewidth-2\tabcolsep\relax}@{}}
\toprule
Corpus & Past (map) & Future (validation) & Quality signal \\
\midrule
Steam games (Steam Games Dataset; Bustos, \hyperlink{ref-Bustos2026}{2026}) & release \textless{} 2024-01-01; 68,222 after removing adult-tagged rows & 2024 onwards; 40,989 & positive review share, items with ≥ 20 reviews \\
arXiv cs.* (arXiv metadata snapshot; Cornell University, \hyperlink{ref-Cornell2026}{2026}), title + abstract & first version \textless{} 2023-01-01; 72,000 & 2023-01-01~to~2026-10-15; 41,000 & none \\
US patent applications (HUPD) & filed 2012--2015; 71,521 & filed 2016--2017; 40,342 & granted vs rejected (pending excluded) \\
\bottomrule
\end{tabular}
\end{table}

Steam items tagged as sexual content,
nudity or NSFW are removed in every
analysis (4,225 rows across both
splits), because a content filter
rejected prompts that showed them as
neighbours. Qwen3-Embedding-0.6B uses
last-token pooling and 512 tokens. For
each future item and for a 10 \%
held-out slice of the past I compute its
isolation (1 − maximum cosine to the
past).

\textbf{Arms.} Targets are built from
the past corpus only, 50 per arm, domain
and generator: REAL (a held-out past
item, itself and near-duplicates with
cos \textgreater{} 0.95 hidden; positive
control), INTERP (normalised midpoint of
two distant past items), GAP (the 50
most isolated of 20,000 normalised draws
from the corpus Gaussian), FAR (the 50
of 20,000 uniform random directions with
the lowest maximum cosine to the
corpus), PLAIN (no target, ``invent a
brand-new item'') and PLAIND (PLAIN with
the titles already proposed shown).

\textbf{Decoder.} Score-only feedback
was chosen over residual-guided feedback
and one-shot best-of-12 in a paired
pilot on 15 targets (best on 11). Steam
used five models as decoders and
generators (gpt-6-luna, run through two
interfaces and counted as two
generators, gpt-6.1-sol, and Claude
Opus, Sonnet and Haiku 5.5); arXiv and
patents used two (gpt-6-luna and Claude
Sonnet 5.5).

\textbf{Judges.} Every item, real or
generated, is rewritten into a neutral
pitch of about 120 words with no name
(on arXiv and patents without reported
results); four judges (gpt-6.1-sol,
Claude Opus, Sonnet and Haiku 5.5) score
each pitch blind on coherence,
feasibility, appeal and novelty, the
last against the five nearest past
items. I rated 80 items blind myself
(Steam 48, arXiv 16, patents 16), 64 of
which also have judge scores; I am not a
domain expert in any of the three
corpora.

\textbf{Pre-registered predictions.} F1,
no future item lands in the far region;
F2, growth is not predicted by hole once
momentum and log past count are
controlled; F3, momentum's AUC for
``top-decile growth cell'' is at least
the hole's, and adding hole and latitude
raises it by less than 0.03 (logistic
regression, 5-fold CV); F4, whether
future items sit nearer the edge or the
centre of the cap than held-out past
items (two-sided Mann-Whitney on
latitude); F5, more isolated future
items are not better received; F6, saved
GAP targets are not filled faster than
density-matched past points. A finding
counts as replicated if direction and
decision match in both embedders. Holm
runs within \{F2, F3, F4, F5\} per
domain and embedder; CIs use 2,000
bootstrap resamples. The F3 bootstrap
resamples cells with replacement and
then cross-validates inside each
resample, so duplicated cells can fall
on both sides of a fold and the
intervals are too narrow.

\subsection{Full table of the future test}\label{app:A.2}

\begin{table}[tbp]
\centering\footnotesize\setlength{\tabcolsep}{4pt}
\caption[]{\textbf{Table A2.} Where the future lands (Qwen3 / BGE-M3).}\label{tab:A2}
\begin{tabular}{@{}P{\dimexpr 0.400\linewidth-2\tabcolsep\relax}P{\dimexpr 0.200\linewidth-2\tabcolsep\relax}P{\dimexpr 0.200\linewidth-2\tabcolsep\relax}P{\dimexpr 0.200\linewidth-2\tabcolsep\relax}@{}}
\toprule
 & Steam & arXiv & Patents \\
\midrule
F1: future items with max cos ≤ 0.15 & 0~/~0 & 0~/~0 & 0~/~0 \\
lowest max-cos of a future item to the past & 0.48~/~0.57 & 0.41~/~0.55 & 0.46~/~0.57 \\
median isolation, future − held-out past & +0.012~/~+0.015 & +0.029~/~+0.018 & +0.004~/~+0.002 \\
F2: partial ρ(growth, hole \textbar{} momentum, size) & +0.20~/~+0.33 & +0.17~/~+0.34 & +0.07~/~+0.07 \\
F3: AUC, momentum alone, scored directly (exploratory) & 0.92~/~0.85 & 0.92~/~0.86 & 0.73~/~0.76 \\
F3: AUC, hole alone, scored directly (exploratory) & 0.45~/~0.54 & 0.49~/~0.53 & 0.50~/~0.48 \\
F3: AUC, momentum alone (cross-validated) & 0.91~/~0.84 & 0.92~/~0.84 & 0.71~/~0.76 \\
F3: AUC, hole alone (cross-validated) & 0.39~/~0.45 & 0.42~/~0.48 & 0.47~/~0.43 \\
F3: AUC gain from adding hole + lat to momentum & −0.006~/~−0.007 & −0.016~/~+0.006 & −0.021~/~+0.031 \\
F4: rank-biserial of latitude, future vs held-out & −0.01~/~−0.12 & −0.17~/~+0.02 & −0.01~/~+0.01 \\
F5: ρ(isolation, quality) & +0.11~/~+0.13 (fails) & n/a & −0.08~/~−0.08 (holds) \\
F6: median percentile of GAP targets (Qwen3 only; exploratory fallback) & 55 & 54 & 39 \\
raw ρ(growth, hole), for comparison with F2 & +0.07~/~+0.26 & −0.07~/~+0.08 & +0.13~/~+0.12 \\
\bottomrule
\end{tabular}
\end{table}

\textbf{F3 intervals.} The
pre-registered bootstrap resampled cells
and then cross-validated, so copies of
one cell could sit on both sides of a
fold. Keeping each resampled cell's
copies in one fold (grouped stratified
5-fold) and z-scoring inside training
folds, the 95 \% intervals of the gain
are −0.107 to +0.009 and −0.102 to
+0.026 (Steam), −0.056 to +0.008 and
−0.029 to +0.033 (arXiv), and −0.141 to
+0.051 and −0.119 to +0.114 (patents),
Qwen3 and BGE-M3; the earlier intervals
ended at 0.012 to 0.037 and at 0.072 and
0.125. \hyperref[tab:A2]{Table~A2} keeps
the registered point estimates
(standardised on all cells); with
in-fold standardisation, the estimand of
the new intervals, they are −0.006 /
−0.007, −0.016 / +0.007 and −0.020 /
+0.031. The decisions, which were
pre-registered on the point estimates,
are unchanged; the arXiv/Qwen3 hole
score reproduces to the third decimal
only (partial ρ 0.173 against 0.175),
for the reason given at the end of this
appendix. Growth and hole share a
denominator; on the numerators alone
(exploratory) the F2 partial ρ is 0.19
to 0.22 (Qwen3) and 0.31 to 0.32
(BGE-M3) in Steam and arXiv, and in
patents 0.09 (CI including 0) and 0.14
(CI touching 0).

\textbf{F4 does not survive
replication.} The future sits slightly
nearer the edge in arXiv with Qwen3
(−0.17) and in Steam with BGE-M3
(−0.12), but neither shift appears in
the other embedder. \textbf{F5 splits by
domain, replicated.} More isolated Steam
games are slightly better reviewed (ρ ≈
+0.11 to +0.13, n = 6,697 after the
adult filter; the pre-registration's
7,081 counts before it); more isolated
patent applications are slightly less
often granted (ρ = −0.08, n = 9,477
decided; 30,865 pending excluded).
\textbf{F6 is inconclusive under the
registered design.} It required 200
held-out controls within ±0.01 of each
GAP target's density, and no target had
them: at most 6 held-out points
qualified per target in Steam, and in
arXiv and patents almost every GAP
target is less dense than any held-out
point. As an exploratory fallback the
script compared each target with its 200
nearest-density held-out points, which
are denser (mean 0.53 to 0.58 against
target medians of 0.41 to 0.52; the
least well matched control is 0.06 to
0.19 away). Against them the targets'
median percentiles are 55, 54 and 39,
none different from 50 (sign tests
\emph{p} = 0.20, 0.20, 0.32), a
comparison biased in an unknown
direction.

\textbf{Validity of the hole score}
(registered follow-up,
\texttt{review\_\allowbreak{}hole\_\allowbreak{}validity.md}).
A gap score is called valid if (a) it
detects holes planted by removing a ball
of items from a cell (AUC ≥ 0.75 with
interval excluding 0.5, in at least 5 of
the 6 corpus × embedder cells), or (b)
it tracks emptiness (rank correlation
with mean kNN density ≤ −0.3 and partial
correlation net of geometry ≤ −0.2, both
with intervals below zero, in at least 5
of 6 cells) and is not dominated by
geometry (R² of position and size ≥ 0.5
while failing the emptiness criterion,
in at least 4 of 6 cells). The
registered score H0 fails both tests at
every resolution and in every space; H1
(kNN density at the centroid), H2 (mean
sparseness of the cell's items) and H3
(share of 500 probe points in the cell
farther than a fixed cosine from every
item) pass. \hyperref[tab:A3]{Table~A3}
summarises them. Correlations of H0 with
size and latitude had been seen while
checking feasibility before the
thresholds were registered. With cells
of at least 100 items (231 to 253 cells
kept) or with Poisson-gamma shrunken
growth, no conclusion changes except the
registered patents F3 point estimate
(\hyperref[sec:4]{Section~4}), and in
Steam every gain interval then ends
below 0.05.

\begin{table}[tbp]
\centering\footnotesize\setlength{\tabcolsep}{4pt}
\caption[]{\textbf{Table A3.} Gap scores at cell level (6 corpus × embedder cells). Planted-hole AUC: range over cells, with the number of cells passing the registered criterion. Correlations: Spearman across the 256 raw cells, range over cells. Growth AUC (directly scored, top decile) and gain over momentum (cross-validated, F3 design): ranges over 128, 256 and 512 raw cells and centred and whitened 256 cells.}\label{tab:A3}
\begin{tabular}{@{}P{\dimexpr 0.220\linewidth-2\tabcolsep\relax}P{\dimexpr 0.200\linewidth-2\tabcolsep\relax}P{\dimexpr 0.140\linewidth-2\tabcolsep\relax}P{\dimexpr 0.160\linewidth-2\tabcolsep\relax}P{\dimexpr 0.130\linewidth-2\tabcolsep\relax}P{\dimexpr 0.150\linewidth-2\tabcolsep\relax}@{}}
\toprule
Score & Planted-hole AUC (cells passing) & ρ with mean density & ρ with centroid cosine to corpus mean & Growth AUC & Gain over momentum \\
\midrule
H0, registered Gaussian hole & 0.45~to~0.59~(0 of 6) & −0.49~to~−0.32 & +0.74~to~+0.80 & 0.34~to~0.57 & −0.098~to~+0.058 \\
H1, centroid kNN density & 0.75~to~0.99~(5 of 6) & −0.95~to~−0.89 & +0.06~to~+0.46 & 0.33~to~0.62 & −0.037~to~+0.049 \\
H2, mean item sparseness & 0.74~to~0.99~(5 of 6) & −1.00 & +0.11~to~+0.54 & 0.32~to~0.65 & −0.035~to~+0.040 \\
H3, uncovered probe share & 0.83~to~0.93~(6 of 6) & −0.96~to~−0.94 & +0.21~to~+0.51 & 0.30~to~0.66 & −0.040~to~+0.055 \\
\bottomrule
\end{tabular}
\end{table}

The reproduction gate of this study
compared the refitted registered
features with the cached ones. k-means
is not bit-reproducible across runs on
the machine used, and an earlier fit in
the same process changes a later one, so
the registered K = 256 was fitted first,
each corpus × embedder ran in its own
process, and the gate was amended
(before the tests were run) to accept a
match either to the cached features or
to the published F2 partial ρ to 10⁻⁶.
arXiv/Qwen3 passed through the second
route with a partition that differs from
the cached one (largest absolute
difference in hole 0.084) and reproduces
the published 0.17472 exactly;
patents/BGE-M3 matched on its second
attempt. The same non-reproducibility,
not floating-point noise in assigning
the Gaussian draws, explains why the
arXiv/Qwen3 hole score had reproduced
only to the third decimal earlier.

\textbf{Item-level tests without cells}
(registered follow-up,
\texttt{review\_\allowbreak{}item\_\allowbreak{}density.md};
thresholds fixed with the F1 medians
known). \emph{A1}, AUC of isolation for
future against held-out past items
(Qwen3 / BGE-M3, 95 \% intervals): Steam
0.559 {[}0.551, 0.566{]} / 0.591
{[}0.584, 0.598{]}; arXiv 0.621
{[}0.614, 0.629{]} / 0.613 {[}0.606,
0.620{]}; patents 0.517 {[}0.509,
0.524{]} / 0.513 {[}0.505, 0.520{]}.
Against held-out items from the momentum
years only: Steam 0.551 / 0.604, arXiv
0.641 / 0.637, patents 0.511 / 0.513.
Low kNN density instead of isolation
gives Steam 0.536 / 0.574, arXiv 0.605 /
0.586, patents 0.514 / 0.504. The rule
was: supports if AUC ≥ 0.60 with lower
bound ≥ 0.55 and Holm \emph{p}
\textless{} 0.05; contradicts if the
upper bound is below 0.55; weak
otherwise when the lower bound exceeds
0.50. Read literally, ``contradicts''
includes intervals reliably above 0.50
but small (Steam/Qwen3 density).
Descriptively, held-out items of the
momentum years against earlier held-out
items give 0.502 to 0.551. \emph{A2},
coverage of 20,000 Gaussian probe points
per corpus and embedder: the rate ratio
(future items per future item over
held-out items per held-out item within
the probe's tenth-neighbour cosine to
the reference set) in the emptiest
decile, divided by that in the 30--70 \%
deciles: Steam 1.057 {[}0.944, 1.174{]}
/ 1.093 {[}0.945, 1.255{]}; arXiv 1.150
{[}1.024, 1.290{]} / 1.097 {[}0.986,
1.221{]}; patents 0.972 {[}0.889,
1.069{]} / 0.921 {[}0.832, 1.023{]}
(Holm \emph{p} 0.12 to 0.69); Spearman
of emptiness with the log rate ratio is
+0.08 and +0.10 in arXiv (Holm \emph{p}
\textless{} 0.006), +0.03 in Steam (not
significant after Holm) and −0.002 and
−0.049 in patents. Supporting required a
ratio of at least 1.25 with lower bound
above 1; contradicting, an upper bound
below 1.10, which only patents meet.
Probes catch few items (median one
held-out and three to seven future
items; 33 to 36 \% have none held out),
and the density-matched check registered
alongside was uninformative, as
anticipated, because almost no held-out
item matches the emptiest probes'
density.

\subsection{Causal anchoring and the cap}\label{app:A.3}

Observationally, decoded FAR items are
much closer to the centroid of their
eight shown neighbours (cosine 0.63 to
0.72 on Steam) than that centroid is to
the target (0.06); on Steam and arXiv
the decoded item is closer to the
centroid than the target is in 100 \% of
FAR items, and on patents every FAR
decode is closer to the centroid than to
its own target. With \(\mathbf{C}\) the
normalised centroid of the shown
embeddings and
\(a_e = \mathbf{e} \cdot \mathbf{C}\),
the identity
\(\mathbf{e} \cdot \mathbf{t} = a_e\,(\mathbf{C} \cdot \mathbf{t}) + \mathbf{r} \cdot \mathbf{t}\),
\(\mathbf{r} = \mathbf{e} - a_e \mathbf{C}\),
separates the anchored part of landing
from the rest; the randomised test asks
whether that anchoring is causal.

\begin{table}[tbp]
\centering\footnotesize\setlength{\tabcolsep}{4pt}
\caption[]{\textbf{Table A4.} Causal anchoring, 20 targets per cell, with the pre-registered Bonferroni intervals (each bound one-sided at 0.05/24, so two-sided 99.6 \%). D = target loss, the decoded text's cosine to its own target with the true neighbours minus with the swapped ones. A = donor attraction, \((\mathbf{e}_\mathrm{swap} - \mathbf{e}_\mathrm{true}) \cdot (\mathbf{C}_\mathrm{donor} - \mathbf{C}_\mathrm{true})\), the move of the decoded embedding \(\mathbf{e}\) towards the centroid of the donor's neighbours and away from the true ones.}\label{tab:A4}
\begin{tabular}{@{}P{\dimexpr 0.085\linewidth-2\tabcolsep\relax}P{\dimexpr 0.120\linewidth-2\tabcolsep\relax}P{\dimexpr 0.070\linewidth-2\tabcolsep\relax}P{\dimexpr 0.255\linewidth-2\tabcolsep\relax}P{\dimexpr 0.200\linewidth-2\tabcolsep\relax}P{\dimexpr 0.270\linewidth-2\tabcolsep\relax}@{}}
\toprule
Domain & Decoder & Arm & D {[}interval{]} & A {[}interval{]} & \(\mathbf{e}\cdot\mathbf{t}\) true / swap / none \\
\midrule
Steam & gpt-6-luna & REAL & +0.265 {[}+0.161,~+0.368{]} & 0.55 {[}0.42,~0.68{]} & 0.691~/~0.426~/~0.383 \\
Steam & Sonnet 5.5 & REAL & +0.311 {[}+0.222,~+0.399{]} & 0.65 {[}0.50,~0.79{]} & 0.765~/~0.454~/~0.465 \\
Steam & gpt-6-luna & GAP & +0.142 {[}+0.085,~+0.198{]} & 0.35 {[}0.28,~0.42{]} & 0.470~/~0.328~/~0.329 \\
Steam & Sonnet 5.5 & GAP & +0.133 {[}+0.084,~+0.182{]} & 0.41 {[}0.32,~0.49{]} & 0.526~/~0.393~/~0.373 \\
Steam & gpt-6-luna & FAR & +0.038 {[}+0.019,~+0.057{]} & 0.30 {[}0.21,~0.39{]} & 0.019~/~−0.019~/~−0.007 \\
Steam & Sonnet 5.5 & FAR & +0.057 {[}+0.041,~+0.072{]} & 0.41 {[}0.30,~0.51{]} & 0.035~/~−0.022~/~−0.019 \\
arXiv & gpt-6-luna & REAL & +0.400 {[}+0.333,~+0.467{]} & 0.81 {[}0.68,~0.93{]} & 0.758~/~0.359~/~0.362 \\
arXiv & Sonnet 5.5 & REAL & +0.479 {[}+0.411,~+0.547{]} & 0.99 {[}0.86,~1.12{]} & 0.809~/~0.330~/~0.377 \\
arXiv & gpt-6-luna & GAP & +0.128 {[}+0.080,~+0.175{]} & 0.46 {[}0.37,~0.55{]} & 0.442~/~0.314~/~0.306 \\
arXiv & Sonnet 5.5 & GAP & +0.184 {[}+0.140,~+0.227{]} & 0.64 {[}0.52,~0.76{]} & 0.486~/~0.302~/~0.283 \\
arXiv & gpt-6-luna & FAR & +0.054 {[}+0.035,~+0.073{]} & 0.47 {[}0.35,~0.60{]} & 0.038~/~−0.016~/~−0.015 \\
arXiv & Sonnet 5.5 & FAR & +0.083 {[}+0.066,~+0.100{]} & 0.64 {[}0.52,~0.77{]} & 0.066~/~−0.017~/~−0.012 \\
\bottomrule
\end{tabular}
\end{table}

\textbf{The quantitative transport law
fails in 7 of 12 cells.} The
pre-registration also froze a
one-coefficient law that predicts where
the decoded text lands from the shown
centroid. It holds within its frozen
error margins in all REAL cells and one
GAP cell, but fails in 7 of 12 cells
(all FAR, three of four GAP): the
direction of anchoring is causal; its
quantitative form is only approximate.
No GAP decode in 300 Steam items (all
Steam generators) and 100 arXiv items is
closer to its target than to the shown
centroid.

\textbf{The cap (raw space).} Two random
directions in \(d = 1024\) have cosine 0
± 0.031, while two Steam games have
median cosine 0.41 (arXiv 0.29); the
corpus mean has norm 0.64 on Steam
(arXiv 0.54, patents 0.51). Let the
in-slab fraction of a vector be the norm
of its projection onto the 232 principal
directions that hold 90 \% of the corpus
energy, \(m(\mathbf{v}) = \|U_k^\top
\mathbf{v}\|\) (a unit vector inside the
slab has \(m = 1\)). Decoded items of
every arm and generator have the in-slab
fraction of a real item (median 0.957
against 0.95 for real games), while FAR
targets have 0.47, the value of a random
direction (\(\sqrt{232/1024} = 0.48\);
as a share of the squared norm, 0.227).
Treating a random direction's scores
against the corpus as Gaussian with
variance (1 − ‖μ‖²)/d, with
\(r = \|\mu\|\), the median corpus
ceiling of the \(k\) most isolated of
\(M\) random directions (here
\(k = 50\), \(M = 20{,}000\)) is
approximately

\[ C_{\mathrm{FAR,median}} \approx \frac{z_N \sqrt{1 - r^2} + r\,\Phi^{-1}\!\left(k/(2M)\right)}{\sqrt d}, \]

with \(z_N\) the Gaussian
expected-maximum approximation for \(N\)
items. The arXiv prediction
(\(N = 72{,}000\), ‖μ‖ = 0.5428) was
0.062, observed 0.058 (error 0.004,
inside the strict criterion of 0.005);
Steam 0.044 against 0.046
(retrospective). On patents the
registered value, 0.0625 ± 0.005, was
not the formula's but an empirical
estimate from a battery of other random
directions on the same corpus; the
observed 0.062 matched it. The
parameter-free formula itself gave 0.068
there, an error of 0.006 that misses the
strict criterion and meets the lenient
one (0.02), as its registration
anticipated. The falsification battery
covered 32 configurations on Steam and
arXiv (all within 0.02, 30 within 0.005,
median absolute error 0.0024) and 16 on
patents (all within 0.02, 9 within
0.005; the formula overpredicts in 15 of
the 16 patents cells). I treat the
formula as a conjecture that has
survived 48 configurations under the
lenient criterion, not as a law: its
residual grows as ‖μ‖ falls.

\textbf{Corrected spaces} (registered
follow-up,
\texttt{review\_\allowbreak{}far\_\allowbreak{}decoding.md}
R1). Under mean-centring (T1),
all-but-the-top (T2), BERT-whitening
with k = 256 (T3) and shrunk whitening
(T4), with FAR and GAP targets rebuilt
in each space by the same rules: the
median pairwise cosine of corpus texts
falls from 0.25--0.59 to −0.012 to
−0.001, the norm of the mean from
0.51--0.77 to at most 0.011, and the
10-occurrence hubness skew from 1.5--6.6
to 0.6--2.4. A FAR target counts as
unreachable when its best past and
future cosines stay below the 1st
percentile of real items'
nearest-neighbour cosine (q01). In 28 of
30 corpus × embedder × transform cells
(22 of the 24 corrected-space cells)
every FAR target is unreachable; the two
exceptions, BERT-whitened BGE-M3 in
Steam (future maximum 0.299 against q01
0.288, one of 50 targets) and arXiv
(0.333 against 0.295, two of 50), are
not confirmed by Qwen3. In T3 and T4 the
GAP and FAR ceilings coincide (for
example Steam/Qwen3 T3 0.228 and 0.227):
after whitening, Gaussian draws are as
far from text as random directions.
Since the pairwise cosine is near zero
in every corrected space while far
targets stay unreachable, the cap does
not explain the result there; in T3 and
T4 the in-slab fraction of FAR targets
(0.91--0.95) equals that of real items,
so the slab does not either, and the FAR
ceilings sit near the median of the
maximum of 20,000 random cosines (for
example 0.104 against 0.129 for
Steam/Qwen3 under T1). Donor attraction
is positive in 12 of 12 decoder cells in
every space; the FAR target loss is
positive in 4 of 4 FAR cells in the raw
space and in 2, 2, 2 and 1 of 4 under T1
to T4.

\subsection{Searching without neighbours}\label{app:A.4}

To ask whether coherent text exists near
a far point at all, the search must not
be anchored. I ran three pre-registered
neighbour-free searches, each gated by a
positive control on hidden REAL items.
The first, an \textbf{LLM score search}
(a black-box LLM writer with score-only
feedback, 44 evaluations per target),
failed its control in every run. A
gradient-based token search with a
language-model fluency penalty (the
second, \textbf{fluent token search})
reached far points only through
high-perplexity token soup
(unconstrained FAR reach 0.22 to 0.23,
collapsing to about 0.04 at λ = 0.2),
and its control was underpowered (arXiv
7 of 10, Steam 4 of 10). The third,
\textbf{seed-and-rewrite}, has an LLM
rewrite the token search's best seeds
into coherent items with the cosine as
its only clue, two models must accept
the result as coherent, and its control
passed on both domains (10 of 10).
\hyperref[tab:A5]{Table~A5} gives the
medians. An earlier version of this
paper described the registered ``≥ q99''
count for GAP targets (19 of 20) as
``closer than any existing item''; the
recount against the ceiling
(\texttt{scripts/\allowbreak{}seed\_\allowbreak{}rewrite\_\allowbreak{}ceiling\_\allowbreak{}count.py},
read-only on the saved run) gives 15 and
14 of 20, and 17 and 15 of 20 within
0.02 below it. For FAR targets the
visible ceiling is exceeded on 16 and 11
of 20, by at most 0.095, and by at least
0.05 over the larger of the visible and
held-out ceilings on 5 and 0.

\begin{table}[tbp]
\centering\footnotesize\setlength{\tabcolsep}{4pt}
\caption[]{\textbf{Table A5.} Seed-and-rewrite, medians over targets (ceiling = best visible corpus item; q99 = 99th percentile of corpus cosines to the target; \(h_\mathrm{coh}\) = best judge-accepted rewrite, maximum over the two rewriters). Targets ≥ q99 is the registered count; targets \textgreater{} ceiling is a post-hoc recount from the saved run (\hyperref[app:A.4]{Appendix~A.4}).}\label{tab:A5}
\begin{tabular}{@{}lllllll@{}}
\toprule
Domain & Arm & ceiling & q99 & \(h_\mathrm{coh}\) & targets ≥ q99 & targets \textgreater{} ceiling \\
\midrule
Steam & REAL & 0.729 & 0.589 & 0.842 & 10/10 & 10/10 \\
Steam & GAP & 0.533 & 0.457 & 0.567 & 19/20 & 15/20 \\
Steam & FAR & 0.042 & −0.002 & 0.069 & 20/20 & 16/20 \\
arXiv & REAL & 0.739 & 0.527 & 0.874 & 10/10 & 10/10 \\
arXiv & GAP & 0.428 & 0.346 & 0.472 & 19/20 & 14/20 \\
arXiv & FAR & 0.055 & 0.010 & 0.057 & 18/20 & 11/20 \\
\bottomrule
\end{tabular}
\end{table}

The registered FAR verdicts use ceiling*
= max(visible corpus ceiling, held-out
ceiling). On arXiv the objection to the
recipe is supported as pre-registered:
90 \% of FAR targets (18 of 20) stay
within 0.03 of ceiling* and none escapes
it by 0.05; against the visible ceiling
alone these are 17 of 20 and 1 of 20
(target k = 14: 0.109 against a visible
ceiling of 0.056 and a held-out one of
0.084). On Steam it is refuted in its
strong form: 5 of 20 FAR targets (25 \%)
get a coherent text at least 0.05 above
ceiling* (the same 5 against the visible
ceiling). That verdict is fragile to the
judge: four of the five escapes carry
invented tokens from the seeds, and
requiring a third model that neither
wrote nor gated any text (not
pre-registered) leaves 3 of 20 (15 \%),
below the 20 \% threshold. The
pre-registered verdict stands as
written. The absolute scale matters
more: the Steam escapes reach cosine
0.09 to 0.14 from their targets, against
a corpus ceiling of 0.04 to 0.05 and a
target at 1.

\textbf{Does the score alone steer?}
(registered follow-up, R3). In the
no-neighbour condition of the anchoring
test, the decoder sees no feedback at
round 0. The change from round 0 to
round 3 in cosine to the own target
minus the mean cosine to the other
targets of the arm is, for FAR targets
with both decoders pooled, −0.0027 (CI
−0.0103 to +0.0058, Holm \emph{p} =
0.75) on Steam and +0.0044 (CI −0.0038
to +0.0133, Holm \emph{p} = 0.31) on
arXiv, below the registered bound of
0.02, unchanged under T1 and T3. For
REAL targets on arXiv it is +0.027 (CI
0.013 to 0.040), for GAP targets +0.024
(CI 0.003 to 0.046) on Steam and +0.025
(CI 0.014 to 0.036) on arXiv, all
Holm-significant, and +0.013 for REAL on
Steam (not significant). The round-0
margin over other targets, from choosing
the best of three candidates, is 0.015
(Steam) and 0.013 (arXiv) for FAR
targets, against a final best margin of
0.035 in both.

\textbf{Trained inversion} (registered
follow-up, R4). No public inverter
exists for Qwen3-Embedding or BGE-M3, so
the test used the vec2text corrector for
gtr-t5-base (Morris et al.,
\hyperlink{ref-Morris2023}{2023}), with
REAL, GAP and FAR targets rebuilt in
that space by the same rules and
32-token texts, on Kaggle GPUs. The
registered validity gates required that
at least 80 \% of real 32-token
truncations pass the two-judge coherence
gate and that hidden real items be
recovered. The inverter recovers hidden
real items (median cosine 0.972 on Steam
and 0.969 on arXiv after 50 correction
steps; round trip on own embeddings BLEU
28.0 and 34.8), but the coherence gate
passed only 12 \% (Steam) and 28 \%
(arXiv) of random real 32-token
truncations, and only 2 of 50 REAL
targets got a gate-passing inversion. By
its registered rule the study is
uninformative in both domains, and I
report no GAP or FAR verdict.
Descriptively, and ungated, FAR
inversions reach a median cosine of 0.10
(Steam) and 0.08 (arXiv), at most 0.18,
as token soup, the inversion analogue of
the fluent search above. The Kaggle
image runs Python 3.13, for which
vec2text's pinned transformers stack has
no wheel; the kernel installs
transformers 4.48.3 and accelerate 1.3.0
before loading the corrector.

\subsection{Non-certifiability}\label{app:A.5}

Fix the encoder, the target
\(\mathbf{t}\) and a proposed bound
\(b\). Let K be the texts with known
coherence labels, P the known coherent
texts, and
\(H_b = \{x : \cos(E(x), \mathbf{t}) > b\}\).
If every known coherent text scores at
most b and some unlabelled admissible
text scores above b, two coherence
predicates that agree on every
observation put the best coherent text
on opposite sides of b
(\texttt{two\_\allowbreak{}worlds}; this
and the next two results are
machine-checked in Lean 4, de Moura and
Ullrich,
\hyperlink{ref-Moura2021}{2021}, in
\texttt{lean/\allowbreak{}LatentGap/\allowbreak{}Impossibility.lean}
of the repository). Over the
unrestricted predicate class, a
transcript certifies b if and only if no
known coherent text is in \(H_b\) and
every member of \(H_b\) has a known
negative label, so any sound certificate
needs at least \(|H_b \setminus K|\)
coherence queries
(\texttt{certifies\_\allowbreak{}iff},
\texttt{query\_\allowbreak{}lower\_\allowbreak{}bound}).
Even with coherence fixed as a known
language-model likelihood threshold,
maximising similarity over coherent
texts encodes Boolean satisfiability
(prose proof, not in Lean). The
statement is close to immediate once
written down. It is the reason
\hyperref[sec:3]{Section~3} states its
far-point result as ``not found'' rather
than ``impossible'': no finite set of
coherence labels certifies a bound
unless every text above it has been
labelled.

\subsection{Combination vs momentum: design and table}\label{app:A.6}

Both of Alien Space's trained models are
replaced by cheap proxies that need no
model calls. Atoms are 200 NMF topics
fitted on TF-IDF of past titles and
abstracts; a paper's atom set is its
three strongest topics; candidates are
all C(200, 3) = 1,313,400 triples.
Scores from past papers only: static
popularity (POP), momentum (M, recent
2021--2022 activity of atoms, pairs and
triple), coherence (COH, mean pairwise
PMI), rarity (RAR, −log(1 + past
count)), NC\(_{0.4}\) = z(COH) + 0.4 ·
z(RAR), and HOLE at the triple's point.
Logistic models are fitted on a
pseudo-window (features to 2020,
outcomes 2021--2022) and applied
unchanged. A triple is a target if it is
the atom set of at least one 2023--2026
paper (24,312 targets, 12,908 strictly
novel). Enrichment counts hits among the
top 11,492 candidates (0.875 \%, the
share Alien Space uses) over uniform
chance; \hyperref[tab:A6]{Table~A6}
gives the results.

\begin{table}[tbp]
\centering\footnotesize\setlength{\tabcolsep}{4pt}
\caption[]{\textbf{Table A6.} Combination vs momentum, arXiv triples (V = 200). Enrichment = hits in the top 0.875 \% over uniform chance.}\label{tab:A6}
\begin{tabular}{@{}P{\dimexpr 0.260\linewidth-2\tabcolsep\relax}P{\dimexpr 0.140\linewidth-2\tabcolsep\relax}P{\dimexpr 0.150\linewidth-2\tabcolsep\relax}P{\dimexpr 0.150\linewidth-2\tabcolsep\relax}P{\dimexpr 0.150\linewidth-2\tabcolsep\relax}P{\dimexpr 0.150\linewidth-2\tabcolsep\relax}@{}}
\toprule
Score & Enrichment, all & AUC, all & Novel, full ranking (a) & Novel, within novel (b) & AUC within novel \\
\midrule
POP (static popularity) & 28.2 & 0.961 & 9.6 & 23.5 & 0.953 \\
M (momentum, unfitted) & 29.1 & 0.969 & 0.0 & 30.9 & 0.962 \\
COH (PMI coherence) = NC\(_0\) & 0.0 & 0.164 & 0.0 & 0.0 & 0.140 \\
NC\(_{0.4}\) (coherence + rarity) & 0.0 & 0.086 & 0.0 & 0.0 & 0.140 \\
HOLE, Qwen3~/~BGE-M3 & 1.43~/~2.30 & 0.622~/~0.693 & 1.87~/~2.36 & 1.87~/~2.39 & 0.618~/~0.696 \\
FIT\_M (fitted momentum) & 34.2 & 0.970 & 8.3 & 31.0 & 0.961 \\
FIT\_MC (+ COH, RAR) & 34.5 & 0.970 & 6.2 & 30.9 & 0.961 \\
FIT\_MCH, Qwen3~/~BGE-M3 (+ HOLE) & 34.5~/~34.6 & 0.970~/~0.970 & 6.4~/~5.9 & 30.6~/~30.9 & 0.961~/~0.961 \\
\bottomrule
\end{tabular}
\end{table}

Column (a) ranks all 1,313,400
candidates and counts strictly novel
targets among the top 0.875 \%; column
(b) ranks only the candidates that never
occurred in the past and counts strictly
novel targets among their top 0.875 \%.

I pre-registered six predictions:
\textbf{H1}, coherence alone (COH)
reaches enrichment above 5× chance on
all targets; \textbf{H2}, unfitted
momentum is at least as good as NC\(_0\)
and NC\(_{0.4}\) in enrichment and AUC
on all targets; \textbf{H3a}, on
strictly novel targets with the full
ranking, adding rarity raises enrichment
(NC\(_{0.4}\) \textgreater{} NC\(_0\));
\textbf{H3b}, within strictly novel
candidates, momentum's AUC is at least
COH's; \textbf{H4}, COH and RAR add less
than 0.02 AUC to fitted momentum;
\textbf{H5}, the embedding hole adds
less than 0.01 AUC, with the same
decision in both embedders. Two failed:
the coherence proxy alone did not exceed
5× chance (H1; it scored 0.0×, because
smoothed PMI rewards pairs of rare
atoms), and adding rarity did not raise
strictly novel hits on the full ranking
(H3a; 0 against 0). The other four held:
momentum beats the combination scores on
all targets (H2) and within strictly
novel candidates (H3b), combination
signals add less than 0.02 AUC to it
(H4), and the embedding hole adds less
than 0.01 (H5). The momentum results
hold, but with a broken coherence proxy
they say nothing about Alien Space's
trained model. A model free to give COH
and RAR any sign gains +0.0002 AUC over
fitted momentum (fitted weights COH
−0.22, RAR −0.07), and the embedding
hole adds −0.0003 (Qwen3) and −0.0000
(BGE-M3) (H5, replicated in both
embedders). For magnitudes only
(different atoms, corpus and holdout),
Alien Space reports 35.9× for coherence
on all targets and 2.35× to 3.24× for
strictly novel triples against uniform
chance; on mine, fitted momentum reaches
34.2× and 8.3×. A uniform-chance
reference makes combination results look
stronger than a popularity reference
would. The pairs arm is saturated (best
possible 2.93×); the V = 100 arm was
dropped under a pre-registered
reduction; NMF stopped at its iteration
cap. The pre-registered bootstrap
resampled future papers and kept the
distinct targets they hit; a resample
keeps on average 17,329 of the 24,312
targets, mostly losing single-paper
triples (72 \% of targets), so it
over-represents popular combinations and
its intervals were off-centre (28 of 55
excluded their own point estimate,
e.g.~momentum's AUC 0.9691 with interval
0.9692 to 0.9701). As a sensitivity
analysis I replaced it with a
paper-cluster bootstrap that weights
each target by its resampled over its
observed paper count, which reproduces
the point estimate exactly. It
conditions on the observed set of
distinct targets and on the fixed
candidate set, so it does not describe
how a new sample of future papers would
change which triples appear; a
target-level bootstrap and DeLong
standard errors give AUC intervals
within about 0.0005 of it. With it, POP
enrichment is 28.2 (27.7 to 28.7), M
29.1 (28.7 to 29.7), fitted momentum
34.2 (33.7 to 34.8) and 8.3 (7.9 to 8.8)
on strictly novel targets, M's AUC
0.9691 (0.9685 to 0.9697); the H4
difference is +0.00015 (0.00012 to
0.00018) and the H5 differences −0.00026
(−0.00033 to −0.00020) in Qwen3 and
−0.00002 (−0.00006 to +0.00002) in
BGE-M3. No decision changes; the
decisions use point estimates, as
pre-registered, and the intervals are
descriptive.

\subsection{Asking without a map, and my ratings against the judges'}\label{app:A.7}

On the LLM judges' novelty (1 to 10,
against the five nearest past items),
PLAIND scored above every map-guided arm
in all three corpora (PLAIND minus
INTERP +1.16 to +3.19 on Steam, +1.07
{[}+0.71, +1.45{]} and +3.01 {[}+2.54,
+3.48{]} on arXiv, +2.09 {[}+1.65,
+2.54{]} and +4.83 {[}+4.46, +5.18{]} on
patents). Plain asking without that
memory collapsed towards one idea per
generator: Sonnet named 45 of its 50
Steam games ``Tidewright'' something,
and median pairwise cosine among a
generator's PLAIN items reached 0.864,
against 0.404 for 50 random corpus games
and 0.35 to 0.49 for the guided arms.
With memory, repetition falls (Steam
0.534 and 0.504), though not to the
guided arms' level. On Steam, GAP was
judged less novel than INTERP (4.91
against 5.52) for all six generators.

This is evidence about judged novelty
and diversity, not about ideas that
work. The judges' novelty seems to
reward specific, unfamiliar wording
against five shown neighbours, while my
own ratings ask whether the concept
already exists. Item-level Spearman
between my novelty and the judges' mean
is weak (two blind Steam batches −0.32
and 0.36, arXiv 0.34, patents 0.13). On
patents I rated every arm lower than the
judges (by 1.2 to 3.0 points) and
answered ``I know this or something that
does the same'' for 3 of 4 FAR and 1 of
3 GAP items, and for none of 5 PLAIND
items. On Steam my PLAIND novelty was
7.0 against 5.75 for INTERP
(Mann-Whitney \emph{p} = 0.105, n = 8 +
8), a direction, not proof.

In an exploratory test on arXiv (future
from 2024-01-01), PLAIND ideas from
gpt-6-luna and Claude Sonnet 5.5 land
nearer later papers than earlier ones
after style is removed (Sonnet 0.58,
gpt-6-luna 0.50; Fisher \emph{p} = 0.002
and 0.020, n = 50 per cell), while
gpt-4o, whose cutoff precedes 2024, does
not (0.38, \emph{p} = 0.20). I read this
as topic-level knowledge of what came
later in generators trained after the
boundary, not foresight. It is also why
the continuation study uses gpt-4o.

\subsection{Continuation study: how real follow-ups move, and the one-shot arms}\label{app:A.8}

\textbf{How real follow-ups move
(pre-registered).} A1, descriptive: the
median step of a real follow-up from its
context is cosine 0.620 (interquartile
range 0.574 to 0.661). I projected each
follow-up's displacement
\(\mathbf{f} - \mathbf{c}\) on three
directions built from the context's 50
nearest past items: \textbf{momentum}
(mean of the recent neighbours minus
mean of the older ones), \textbf{hole}
(from \(\mathbf{c}\) towards the most
isolated of 2,000 draws from the
neighbourhood's Gaussian) and
\textbf{centre} (towards the normalised
corpus mean). \textbf{A2 failed, and its
design was flawed.} I predicted that
follow-ups would align more with
momentum than with the hole; the mean
projection was 0.016 on momentum and
0.336 on the hole (difference −0.320, CI
−0.334 to −0.307). The two directions
are not comparable: the hole direction
runs from \(\mathbf{c}\) to a draw
centred on the mean of \(\mathbf{c}\)'s
own neighbourhood, so any later paper
near \(\mathbf{c}\) projects positively
on it, while the momentum direction is a
difference of two means that does not
start at \(\mathbf{c}\). \textbf{A3}
(two-sided): follow-ups move towards the
centre of the cap (0.427, CI 0.420 to
0.435). \textbf{Matched baseline
(exploratory).} Matching each follow-up
with the later paper outside the sampled
follow-up set whose cosine to
\(\mathbf{c}\) is closest (median
absolute step mismatch 0.0005, so this
match is within support), follow-ups
drift towards the centre less than those
papers (0.427 against 0.471, \emph{p}
\textless{} 10⁻²²) and project slightly
less on the hole direction (0.336
against 0.350, \emph{p} \textless{}
10⁻⁴), while momentum does not separate
them (0.016 against 0.015, \emph{p} =
0.61).

\textbf{One-shot arms (pre-registered).}
Five ideas per context in each of four
arms: CTX (the context's title and
abstract), CTX+NB (plus the eight
nearest past items), CTX+MEM
(sequential, seeing the titles it
already proposed) and CTX+MOM (plus the
eight most recent of the context's 50
nearest past items, ``where this area is
moving''); \hyperref[tab:A7]{Table~A7}
gives the means.

\begin{table}[tbp]
\centering\footnotesize\setlength{\tabcolsep}{4pt}
\caption[]{\textbf{Table A7.} One-shot arms, means over 150 contexts (gpt-4o, Qwen3). Step gap = median idea step minus median follow-up step, per context; diversity = mean pairwise cosine among a context's ideas (lower is more diverse).}\label{tab:A7}
\begin{tabular}{@{}lllll@{}}
\toprule
Arm & step gap & fit & hit & diversity \\
\midrule
CTX & 0.149 & 0.680 & 0.659 & 0.824 \\
CTX+NB & 0.140 & 0.686 & 0.675 & 0.808 \\
CTX+MEM & 0.128 & 0.668 & 0.611 & 0.767 \\
CTX+MOM & 0.138 & 0.685 & 0.617 & 0.803 \\
\bottomrule
\end{tabular}
\end{table}

\textbf{B2 does not survive Holm.}
CTX+NB ideas are slightly closer to the
shown neighbours' centroid than to the
follow-ups' centroid (difference 0.008,
CI −0.001 to 0.016; \emph{p} = 0.023,
Holm 0.068); in BGE-M3 the same
difference, 0.009 (CI 0.004 to 0.013),
survives Holm (\emph{p} \textless{}
0.001), so B2 holds in one embedder
only. \textbf{B3 holds.} Memory of
earlier titles diversifies the ideas
(0.767 against 0.824; Holm \emph{p}
\textless{} 10⁻¹⁹), in all four gpt-4o
cells of the robustness grid and for
OLMo (0.826 → 0.808 in Qwen3), but it
does not lengthen the step much (gap
0.128). \textbf{B4 fails; the point
estimate is reversed.} I predicted that
a verbal hint of where the area is
moving would give the highest hit rate.
Its hit rate was 0.617 against 0.659 for
CTX (difference −0.041, CI −0.083 to
−0.001; McNemar on context majorities 8
against 15), with fit essentially
unchanged (0.685 against 0.680); the
best arm by hit was CTX+NB (0.675). In
BGE-M3 B4 also fails, but the reversal
is not significant (−0.024, CI −0.065 to
+0.019) and the best arm by hit is CTX.
\textbf{Centre drift (exploratory).}
Ideas drift towards the centre much less
than follow-ups (0.13 to 0.15 against
0.427), at a much shorter step, so the
comparison is confounded by step length.

\textbf{Hit under critique (C2, C4;
pre-registered).} C2 fails: GEO's hit
rate drops by 0.11 (CI −0.16 to −0.06).
C4 (two-sided): under the reviewer, hit
drops by 0.05 (CI −0.09 to −0.02).
Against NEUTRAL, GEO loses more hits in
Qwen3 (N4, −0.096, Holm \emph{p} =
0.001). The round-0 hit rate differs
between samples (0.63 to 0.66 in the
discovery sample, 0.54 to 0.60 in the
replication) by more than any arm
effect, which is one more reason no
claim rests on hit.

\textbf{On the hit metric.} Hit compares
an idea's best cosine to at most 50
follow-ups with its best cosine to every
later paper of the same primary category
(excluding the follow-ups), so its level
depends on the number of distractors and
follow-ups per context, and near ties
decide many cases. I report it where it
was pre-registered (C2, C4 and N4,
above) but base no claim on it.

\textbf{Second generator (details).}
OLMo-7B-0424-Instruct replaced the
planned second generator (a deviation
from the pre-registration). It ran with
the prompts copied verbatim and three
forced differences: its 2,048-token
window excludes the prompts that show
neighbours, so it runs only the one-shot
CTX arm, the memory arm and the three
critique arms; its chat template has no
system role, so the system text is
prepended to the user turn; and it
answers as ``Title: \ldots{} Abstract:
\ldots{}'' rather than JSON, so a
tolerant parser reads it. Compute limits
reduced the critique arms to one idea
per context and arm.

\begin{table}[tbp]
\centering\footnotesize\setlength{\tabcolsep}{4pt}
\caption[]{\textbf{Table A8.} Both generators and embedders (exploratory). Step gap: CTX ideas minus real follow-ups, all / S2-flagged subset. Every other column is the \textbf{range over the three critique arms} (and, for step and fit, over both follow-up sets) of the change from round 0 to round 2; the alignment columns use all follow-ups (\hyperref[tab:5]{Table~5}).}\label{tab:A8}
\begin{tabular}{@{}P{\dimexpr 0.160\linewidth-2\tabcolsep\relax}P{\dimexpr 0.150\linewidth-2\tabcolsep\relax}P{\dimexpr 0.155\linewidth-2\tabcolsep\relax}P{\dimexpr 0.180\linewidth-2\tabcolsep\relax}P{\dimexpr 0.175\linewidth-2\tabcolsep\relax}P{\dimexpr 0.180\linewidth-2\tabcolsep\relax}@{}}
\toprule
Generator / embedder & Step gap (all / S2) & Step lengthened & Fit change & Alignment change & Follow-up-specific \\
\midrule
gpt-4o / Qwen3 & 0.149~/~0.129 & 0.029~to~0.087 & −0.041~to~−0.012 & +0.014~to~+0.033 & −0.003~to~+0.015 \\
gpt-4o / BGE-M3 & 0.168~/~0.159 & 0.028~to~0.048 & −0.017~to~−0.007 & +0.015~to~+0.041 & +0.001~to~+0.024 \\
OLMo / Qwen3 & 0.218~/~0.197 & 0.032~to~0.092 & −0.021~to~−0.006 & +0.038~to~+0.077 & +0.024~to~+0.032 \\
OLMo / BGE-M3 & 0.188~/~0.179 & 0.029~to~0.061 & −0.006~to~+0.003 & +0.031~to~+0.075 & +0.012~to~+0.028 \\
\bottomrule
\end{tabular}
\end{table}

Critique lengthens the step in all 24
generator × embedder × follow-up-set ×
arm cells. Fit falls in 23 of 24 cells
(\emph{p} \textless{} 0.05 uncorrected
in 17) and rises in none significantly
(OLMo, BGE-M3, S2-flagged, GEO: +0.003,
\emph{p} = 0.57).

\textbf{OLMo in the discovery sample
(exploratory, with a registered neutral
arm).} With OLMo the step gap is larger
(0.179 to 0.218) and critique lengthens
the step in every cell, as with gpt-4o;
fit falls in 23 of 24 generator ×
embedder × follow-up-set × arm cells and
rises significantly in none
(\hyperref[tab:A8]{Table~A8}). Its
follow-up-specific changes are +0.012 to
+0.032 (\hyperref[tab:5]{Table~5}), the
reviewer's surviving Holm, ORACLE's in
Qwen3 with CI +0.006 to +0.044. A
NEUTRAL arm for OLMo, registered and run
afterwards on the same round-0 ideas,
lengthens OLMo's step (0.042 in Qwen3)
and turns it along \(\mathbf{s}\) by
+0.012 (CI −0.004 to +0.028; +0.019 in
BGE-M3). Against it, GEO still lengthens
the step more (+0.043, CI +0.031 to
+0.055, Holm \emph{p} \textless{} 0.001;
+0.023 in BGE-M3) and raises alignment
(+0.037; +0.032), while the reviewer's
longer step is not detected beyond
revision (+0.010, CI −0.002 to +0.022).
Along \(\mathbf{s}\) no critique arm
differs from NEUTRAL after Holm (GEO
+0.012, reviewer +0.020, ORACLE +0.013
in Qwen3; −0.006 to +0.010 in BGE-M3),
with intervals between −0.024 and
+0.042, so a feedback-specific turn was
neither shown nor bounded on these
contexts; the confirmatory test of
\hyperref[sec:5.3]{Section~5.3}
followed.

\textbf{Tangent confounds} (registered
follow-up,
\texttt{review\_\allowbreak{}tangent\_\allowbreak{}confounds.md}
Part A). \emph{Gap fractions.} With FU
the follow-ups' own leave-one-out
projection on \(\mathbf{s}\) and I0 the
round-0 ideas', G = FU − I0 is 0.136 (CI
0.115 to 0.158) in the gpt-4o
replication (FU 0.254, I0 0.118) and
0.151 in the OLMo test (FU 0.267, I0
0.116). Feedback beyond revision as a
fraction of G is 0.078 (CI 0.001 to
0.158) for GEO, 0.026 (−0.064 to 0.113)
for the reviewer and 0.024 (−0.070 to
0.116) for ORACLE in the replication,
and −0.013, −0.019 and 0.065 in the OLMo
test (upper bounds 0.060, 0.060 and
0.163). The bound ±0.015 is 0.159 /
0.144 of G (Qwen3 / BGE-M3) in the
discovery sample, 0.110 / 0.126 in the
replication, 0.099 / 0.165 in the OLMo
test and 0.342 / 0.254 for gpt-5.4-mini.
\emph{Step.} Within arm, the common
slope of alignment change on step change
is 0.329 (CI 0.176 to 0.478) in the
discovery sample, 0.269 (0.144 to 0.398)
in the replication and 0.148 (0.011 to
0.287) in the OLMo test; for most
critique arms the step explains at least
half of the alignment rise in the
discovery sample and the OLMo test, not
in the replication. Step-adjusted
contrasts along \(\mathbf{s}\) with
NEUTRAL (context fixed effects):
discovery GEO −0.013 (CI −0.024 to
−0.002, Holm \emph{p} = 0.069), reviewer
−0.017 (−0.028 to −0.005, Holm 0.024),
ORACLE −0.019 (−0.032 to −0.006, Holm
0.024); replication +0.008, +0.002 and
+0.004 (all insufficient). \emph{Genre.}
With g the unit mean of ideas minus real
abstracts (‖mean difference‖ 0.11 to
0.33 across cells), the four contrast
readings that change are discovery Qwen3
ORACLE − NEUTRAL alignment (now
Holm-significant, −0.020), OLMo Qwen3
GEO and reviewer (insufficient) and
replication BGE-M3 E4 (from detected to
insufficient); the readings that depend
on how genre is defined are those that
change with a genre direction built from
ideas minus contexts: it additionally
removes the Holm significance of
discovery Qwen3 ORACLE's N6, and makes
two gpt-5.4-mini contrasts in BGE-M3
Holm-significant, the reviewer's
alignment against NEUTRAL (N5) and GEO's
follow-up-specific change against
NEUTRAL (N6). A direction built from the
follow-ups reproduces all four primary
changes; one that leaves each context
out reproduces three of them, keeping
OLMo Qwen3 GEO at ``no feedback-specific
turn of 0.015 or more'' (−0.003, CI
−0.015 to +0.008).
\hyperref[tab:A9]{Table~A9} gives the
fractions for every sample and embedder.
\emph{Echo} is reported in
\hyperref[sec:5.1]{Section~5.1};
critique lowers it (GEO, −0.015 in
3-gram containment). \emph{Shared
follow-ups.} Distinct follow-ups shared
by at least two contexts: 24 of 2,859
(discovery), 8 of 2,410 (replication),
48 of 4,085 (OLMo test); a bootstrap
over connected components changes one
interval, replication Qwen3 E4 (+0.0001
to +0.0213), and no Holm decision.
\emph{Citation counts.} Replication
contexts have a median of 39 all-venue
citations and the OLMo contexts 57,
against 11 in their frame; the step gap
rises with citation count in the OLMo
test only.

\Needspace*{15\baselineskip}
\begingroup\footnotesize\setlength{\tabcolsep}{4pt}\setlength{\LTcapwidth}{\linewidth}
\begin{longtable}{@{}P{\dimexpr 0.240\linewidth-2\tabcolsep\relax}P{\dimexpr 0.160\linewidth-2\tabcolsep\relax}P{\dimexpr 0.130\linewidth-2\tabcolsep\relax}P{\dimexpr 0.210\linewidth-2\tabcolsep\relax}P{\dimexpr 0.260\linewidth-2\tabcolsep\relax}@{}}
\caption[]{\textbf{Table A9.} Turns along \(\mathbf{s}\) as fractions of the gap G between the follow-ups' own leave-one-out projection on \(\mathbf{s}\) and the round-0 ideas' projection (registered follow-up), with bootstrap 95 \% intervals. \(\varphi\): the arm's own change from round 0 to round 2 over G. \(\psi\): the arm's change minus NEUTRAL's over G (for ORACLE-H, minus ORACLE's, on the vectors that hold both arms). In the discovery samples Reviewer and ORACLE do not share round-0 ideas with NEUTRAL; the gpt-4o discovery ORACLE-H arm has no BGE-M3 vectors.}\label{tab:A9}\\
\toprule
Sample, embedder & G {[}95 \% CI{]}; ±0.015 / G & Arm & \(\varphi\) (own change / G) & \(\psi\) (beyond NEUTRAL / G) \\
\midrule
\endfirsthead
\multicolumn{5}{@{}l}{\emph{Table A9, continued}}\\
\toprule
Sample, embedder & G {[}95 \% CI{]}; ±0.015 / G & Arm & \(\varphi\) (own change / G) & \(\psi\) (beyond NEUTRAL / G) \\
\midrule
\endhead
\bottomrule
\endlastfoot
Discovery, gpt-4o, Qwen3 & 0.094 {[}0.073, 0.116{]}; 0.16 & GEO & +0.16 {[}+0.06,~+0.28{]} & −0.04 {[}−0.15,~+0.07{]} \\
Discovery, gpt-4o, Qwen3 &  & Reviewer & +0.07 {[}−0.02,~+0.16{]} & −0.13 {[}−0.27,~−0.02{]} \\
Discovery, gpt-4o, Qwen3 &  & ORACLE & −0.03 {[}−0.14,~+0.07{]} & −0.23 {[}−0.39,~−0.09{]} \\
Discovery, gpt-4o, Qwen3 &  & NEUTRAL & +0.20 {[}+0.10,~+0.32{]} & --- \\
Discovery, gpt-4o, Qwen3 &  & ORACLE-H & +0.07 {[}−0.02,~+0.18{]} & +0.10 {[}−0.03,~+0.25{]} (minus ORACLE) \\
Discovery, OLMo, Qwen3 & 0.128 {[}0.105, 0.150{]}; 0.12 & GEO & +0.19 {[}+0.04,~+0.34{]} & +0.10 {[}−0.05,~+0.25{]} \\
Discovery, OLMo, Qwen3 &  & Reviewer & +0.25 {[}+0.12,~+0.39{]} & +0.16 {[}−0.00,~+0.34{]} \\
Discovery, OLMo, Qwen3 &  & ORACLE & +0.19 {[}+0.06,~+0.33{]} & +0.10 {[}−0.08,~+0.29{]} \\
Discovery, OLMo, Qwen3 &  & NEUTRAL & +0.09 {[}−0.04,~+0.22{]} & --- \\
Replication, gpt-4o, Qwen3 & 0.136 {[}0.115, 0.158{]}; 0.11 & GEO & +0.14 {[}+0.07,~+0.22{]} & +0.08 {[}+0.00,~+0.16{]} \\
Replication, gpt-4o, Qwen3 &  & Reviewer & +0.09 {[}+0.03,~+0.15{]} & +0.03 {[}−0.06,~+0.11{]} \\
Replication, gpt-4o, Qwen3 &  & ORACLE & +0.09 {[}+0.02,~+0.15{]} & +0.02 {[}−0.07,~+0.12{]} \\
Replication, gpt-4o, Qwen3 &  & NEUTRAL & +0.06 {[}−0.01,~+0.14{]} & --- \\
OLMo test, Qwen3 & 0.151 {[}0.122, 0.181{]}; 0.10 & GEO & +0.03 {[}−0.06,~+0.11{]} & −0.01 {[}−0.09,~+0.06{]} \\
OLMo test, Qwen3 &  & Reviewer & +0.02 {[}−0.05,~+0.09{]} & −0.02 {[}−0.10,~+0.06{]} \\
OLMo test, Qwen3 &  & ORACLE & +0.11 {[}+0.03,~+0.19{]} & +0.07 {[}−0.03,~+0.16{]} \\
OLMo test, Qwen3 &  & NEUTRAL & +0.04 {[}−0.03,~+0.11{]} & --- \\
gpt-5.4-mini, Qwen3 & 0.044 {[}0.027, 0.061{]}; 0.34 & GEO & −0.06 {[}−0.42,~+0.20{]} & +0.10 {[}−0.19,~+0.36{]} \\
gpt-5.4-mini, Qwen3 &  & Reviewer & −0.10 {[}−0.44,~+0.13{]} & +0.06 {[}−0.22,~+0.35{]} \\
gpt-5.4-mini, Qwen3 &  & ORACLE & −0.13 {[}−0.44,~+0.09{]} & +0.03 {[}−0.24,~+0.31{]} \\
gpt-5.4-mini, Qwen3 &  & NEUTRAL & −0.16 {[}−0.45,~+0.02{]} & --- \\
gpt-5.4-mini, Qwen3 &  & ORACLE-H & +0.02 {[}−0.21,~+0.23{]} & +0.15 {[}−0.12,~+0.48{]} (minus ORACLE) \\
Discovery, gpt-4o, BGE-M3 & 0.104 {[}0.089, 0.119{]}; 0.14 & GEO & +0.23 {[}+0.13,~+0.34{]} & +0.10 {[}+0.00,~+0.19{]} \\
Discovery, gpt-4o, BGE-M3 &  & Reviewer & +0.06 {[}−0.02,~+0.13{]} & −0.08 {[}−0.19,~+0.03{]} \\
Discovery, gpt-4o, BGE-M3 &  & ORACLE & +0.01 {[}−0.08,~+0.09{]} & −0.13 {[}−0.24,~−0.02{]} \\
Discovery, gpt-4o, BGE-M3 &  & NEUTRAL & +0.14 {[}+0.05,~+0.22{]} & --- \\
Discovery, OLMo, BGE-M3 & 0.120 {[}0.104, 0.136{]}; 0.12 & GEO & +0.24 {[}+0.10,~+0.38{]} & +0.08 {[}−0.06,~+0.22{]} \\
Discovery, OLMo, BGE-M3 &  & Reviewer & +0.15 {[}+0.04,~+0.28{]} & −0.00 {[}−0.15,~+0.16{]} \\
Discovery, OLMo, BGE-M3 &  & ORACLE & +0.10 {[}−0.01,~+0.22{]} & −0.05 {[}−0.21,~+0.10{]} \\
Discovery, OLMo, BGE-M3 &  & NEUTRAL & +0.15 {[}+0.04,~+0.26{]} & --- \\
Replication, gpt-4o, BGE-M3 & 0.119 {[}0.104, 0.135{]}; 0.13 & GEO & +0.17 {[}+0.09,~+0.26{]} & +0.10 {[}+0.01,~+0.18{]} \\
Replication, gpt-4o, BGE-M3 &  & Reviewer & +0.08 {[}+0.01,~+0.16{]} & +0.01 {[}−0.08,~+0.10{]} \\
Replication, gpt-4o, BGE-M3 &  & ORACLE & +0.03 {[}−0.03,~+0.09{]} & −0.04 {[}−0.14,~+0.05{]} \\
Replication, gpt-4o, BGE-M3 &  & NEUTRAL & +0.08 {[}+0.00,~+0.15{]} & --- \\
OLMo test, BGE-M3 & 0.091 {[}0.067, 0.116{]}; 0.16 & GEO & +0.07 {[}−0.05,~+0.19{]} & −0.00 {[}−0.12,~+0.10{]} \\
OLMo test, BGE-M3 &  & Reviewer & +0.07 {[}−0.04,~+0.17{]} & −0.00 {[}−0.12,~+0.11{]} \\
OLMo test, BGE-M3 &  & ORACLE & +0.08 {[}−0.02,~+0.20{]} & +0.01 {[}−0.12,~+0.15{]} \\
OLMo test, BGE-M3 &  & NEUTRAL & +0.07 {[}−0.04,~+0.19{]} & --- \\
gpt-5.4-mini, BGE-M3 & 0.059 {[}0.044, 0.075{]}; 0.25 & GEO & +0.27 {[}+0.09,~+0.46{]} & +0.27 {[}+0.10,~+0.44{]} \\
gpt-5.4-mini, BGE-M3 &  & Reviewer & +0.15 {[}−0.01,~+0.29{]} & +0.14 {[}−0.04,~+0.32{]} \\
gpt-5.4-mini, BGE-M3 &  & ORACLE & −0.09 {[}−0.24,~+0.03{]} & −0.10 {[}−0.27,~+0.06{]} \\
gpt-5.4-mini, BGE-M3 &  & NEUTRAL & +0.01 {[}−0.13,~+0.14{]} & --- \\
gpt-5.4-mini, BGE-M3 &  & ORACLE-H & −0.05 {[}−0.21,~+0.09{]} & +0.04 {[}−0.12,~+0.21{]} (minus ORACLE) \\
\end{longtable}
\endgroup

\textbf{Rewrite and paraphrase nulls}
(registered follow-up, Part B of
\texttt{review\_\allowbreak{}tangent\_\allowbreak{}confounds.md}).
gpt-4o rewrote all 2,883 discovery and
2,419 replication follow-ups as
proposals, with 450 paraphrases of
contexts and 450 rewrites of nearby
non-citing later papers per sample; OLMo
did the same on Kaggle for 1,197
continuations and sampled follow-ups.
One rewrite per sample needed a retry
and none was dropped for gpt-4o; OLMo's
context paraphrases lost 92 of 450 to
parse failures. Rewrites keep their
content at a median cosine of 0.86 to
the original (Qwen3; 22 \% below 0.80)
and paraphrases at 0.94. The registered
P2 rule compares, at matched step, the
ideas with the rewritten nearby papers
(difference along \(\mathbf{s}\):
+0.036, CI 0.021 to 0.052, discovery;
+0.014, CI −0.000 to +0.028,
replication; −0.006, CI −0.021 to
+0.010, OLMo; Qwen3) and asks whether a
dose-response fitted on the null texts
alone predicts the observed alignment
rise; it does for most critique arms in
both gpt-4o samples (discovery GEO 0.057
predicted against 0.033 observed,
reviewer 0.041 against 0.030, ORACLE
0.021 against 0.014), so every cell
reads ``not separable from a step effect
with these controls''.

\textbf{OLMo test by number of
continuations} (registered follow-up,
\texttt{review\_\allowbreak{}olmo\_\allowbreak{}reliability.md}).
83 contexts have two consensus
continuations, 44 three or four and 23
five or more. The bound ±0.015 is 0.066,
0.052 and 0.041 of the continuations'
leave-one-out projection in these
strata. The contrasts do not differ
between contexts with two continuations
and the rest (reviewer −0.022, CI −0.045
to +0.001, Holm \emph{p} = 0.17;
differences below about 0.032 have less
than 80 \% power). Split-half
reliability of \(\mathbf{s}\) on the 46
contexts with at least four
continuations is a mean split-half
correlation of 0.435 and λ = 0.298 (CI
0.247 to 0.357), giving a reliability of
0.37 for two continuations; with 20,000
resamples the Monte Carlo error of the
interval ends is about 10⁻⁴. In BGE-M3
the corrected bound is about 0.016.
\hyperref[tab:A10]{Tables~A10} and
\hyperref[tab:A11]{A11} give the
registered per-embedder reporting by
stratum.

\FloatBarrier

\begin{table}[tbp]
\centering\footnotesize\setlength{\tabcolsep}{4pt}
\caption[]{\textbf{Table A10.} OLMo test by number of consensus continuations k (registered follow-up): number of contexts n, the continuations' mean leave-one-out projection on \(\mathbf{s}\) (L), the bound ±0.015 as a fraction of it, and each critique arm's change along \(\mathbf{s}\) minus NEUTRAL's, with 95 \% t intervals.}\label{tab:A10}
\begin{tabular}{@{}P{\dimexpr 0.130\linewidth-2\tabcolsep\relax}P{\dimexpr 0.050\linewidth-2\tabcolsep\relax}P{\dimexpr 0.060\linewidth-2\tabcolsep\relax}P{\dimexpr 0.080\linewidth-2\tabcolsep\relax}P{\dimexpr 0.227\linewidth-2\tabcolsep\relax}P{\dimexpr 0.227\linewidth-2\tabcolsep\relax}P{\dimexpr 0.226\linewidth-2\tabcolsep\relax}@{}}
\toprule
Embedder, k & n & L & 0.015 / L & GEO − NEUTRAL & Reviewer − NEUTRAL & ORACLE − NEUTRAL \\
\midrule
Qwen3, 2 & 83 & 0.228 & 0.066 & −0.005 {[}−0.020,~+0.009{]} & −0.013 {[}−0.030,~+0.004{]} & +0.008 {[}−0.012,~+0.028{]} \\
Qwen3, 3--4 & 44 & 0.290 & 0.052 & −0.007 {[}−0.028,~+0.014{]} & +0.000 {[}−0.014,~+0.014{]} & +0.005 {[}−0.020,~+0.031{]} \\
Qwen3, 5 or more & 23 & 0.362 & 0.041 & +0.020 {[}−0.015,~+0.055{]} & +0.027 {[}−0.010,~+0.064{]} & +0.025 {[}−0.008,~+0.058{]} \\
Qwen3, all & 150 & 0.267 & 0.056 & −0.002 {[}−0.013,~+0.009{]} & −0.003 {[}−0.015,~+0.009{]} & +0.010 {[}−0.004,~+0.024{]} \\
BGE-M3, 2 & 83 & 0.246 & 0.061 & −0.006 {[}−0.018,~+0.006{]} & −0.004 {[}−0.016,~+0.009{]} & +0.004 {[}−0.010,~+0.017{]} \\
BGE-M3, 3--4 & 44 & 0.295 & 0.051 & −0.001 {[}−0.023,~+0.020{]} & +0.007 {[}−0.015,~+0.029{]} & +0.004 {[}−0.023,~+0.032{]} \\
BGE-M3, 5 or more & 23 & 0.346 & 0.043 & +0.022 {[}−0.004,~+0.047{]} & −0.003 {[}−0.033,~+0.027{]} & −0.014 {[}−0.049,~+0.022{]} \\
BGE-M3, all & 150 & 0.276 & 0.054 & −0.000 {[}−0.010,~+0.009{]} & −0.000 {[}−0.011,~+0.010{]} & +0.001 {[}−0.011,~+0.013{]} \\
\bottomrule
\end{tabular}
\end{table}

\begin{table}[tbp]
\centering\footnotesize\setlength{\tabcolsep}{4pt}
\caption[]{\textbf{Table A11.} OLMo test by number of consensus continuations k: each arm's own change along \(\mathbf{s}\) from round 0 to round 2, with 95 \% t intervals; in the rows for all contexts, Holm over the four arms (Wilcoxon).}\label{tab:A11}
\begin{tabular}{@{}P{\dimexpr 0.180\linewidth-2\tabcolsep\relax}P{\dimexpr 0.205\linewidth-2\tabcolsep\relax}P{\dimexpr 0.205\linewidth-2\tabcolsep\relax}P{\dimexpr 0.205\linewidth-2\tabcolsep\relax}P{\dimexpr 0.205\linewidth-2\tabcolsep\relax}@{}}
\toprule
Embedder, k & GEO & Reviewer & ORACLE & NEUTRAL \\
\midrule
Qwen3, 2 & +0.013 {[}−0.003, +0.030{]} & +0.006 {[}−0.008, +0.020{]} & +0.027 {[}+0.010, +0.043{]} & +0.019 {[}+0.004, +0.034{]} \\
Qwen3, 3--4 & −0.017 {[}−0.037, +0.003{]} & −0.010 {[}−0.029, +0.009{]} & −0.005 {[}−0.026, +0.016{]} & −0.010 {[}−0.026, +0.006{]} \\
Qwen3, 5 or more & +0.011 {[}−0.029, +0.051{]} & +0.018 {[}−0.016, +0.052{]} & +0.016 {[}−0.007, +0.039{]} & −0.009 {[}−0.031, +0.013{]} \\
Qwen3, all (Holm over four) & +0.004 {[}−0.008, +0.016{]}, Holm 1 & +0.003 {[}−0.008, +0.014{]}, Holm 1 & +0.016 {[}+0.004, +0.027{]}, Holm 0.076 & +0.006 {[}−0.004, +0.016{]}, Holm 1 \\
BGE-M3, 2 & +0.007 {[}−0.007, +0.022{]} & +0.010 {[}−0.003, +0.022{]} & +0.017 {[}+0.005, +0.029{]} & +0.013 {[}+0.001, +0.025{]} \\
BGE-M3, 3--4 & −0.013 {[}−0.032, +0.006{]} & −0.005 {[}−0.022, +0.013{]} & −0.007 {[}−0.026, +0.012{]} & −0.011 {[}−0.033, +0.010{]} \\
BGE-M3, 5 or more & +0.038 {[}+0.002, +0.074{]} & +0.013 {[}−0.016, +0.043{]} & +0.002 {[}−0.028, +0.033{]} & +0.016 {[}−0.007, +0.040{]} \\
BGE-M3, all (Holm over four) & +0.006 {[}−0.005, +0.017{]}, Holm 0.52 & +0.006 {[}−0.004, +0.016{]}, Holm 0.41 & +0.008 {[}−0.002, +0.017{]}, Holm 0.21 & +0.006 {[}−0.003, +0.016{]}, Holm 0.41 \\
\bottomrule
\end{tabular}
\end{table}

\textbf{Step gap by follow-up year
(exploratory).} Dated by their first
arXiv version (computed with
\texttt{review\_\allowbreak{}olmo\_\allowbreak{}2023\_\allowbreak{}followups.md}),
the discovery follow-ups split into
1,161 from 2024, 1,006 from 2025 and 716
from 2026, and the gap of gpt-4o's
one-shot ideas is 0.153 (CI 0.139 to
0.166), 0.155 (0.141 to 0.170) and 0.155
(0.140 to 0.170) in Qwen3 and 0.164,
0.170 and 0.177 in BGE-M3.
\texttt{scripts/\allowbreak{}step\_\allowbreak{}gap\_\allowbreak{}by\_\allowbreak{}year.py}
dates them instead by their Semantic
Scholar publication date (1,177, 999 and
707) and gives 0.153, 0.154 and 0.155
(Qwen3). With that dating, within
contexts, a follow-up's step changes by
−0.0004 per year (CI −0.0057 to +0.0048)
in Qwen3 and −0.0070 (−0.0098 to
−0.0038) in BGE-M3, where later
follow-ups are slightly farther from the
context; contexts from 2021 and 2022
give 0.155 and 0.144 (Qwen3).

\textbf{OLMo against citing papers from
2023} (registered follow-up,
\texttt{review\_\allowbreak{}olmo\_\allowbreak{}2023\_\allowbreak{}followups.md}).
For each discovery context I fetched
every citing paper whose first arXiv
version falls between 1 April and 31
December 2023, after OLMo's data cutoff
(1,659 papers, at most 50 per context:
1,204 papers in 143 contexts), and
embedded them with the follow-ups'
recipe. The registered rule calls the
short step not a lag artefact in an
embedder if OLMo's gap against them is
positive with its interval excluding
zero and, on contexts with both, at
least half its 2024 gap. Both cells
pass: 0.215 (CI 0.200 to 0.231) and
0.184 (0.175 to 0.194) against 2023
papers; on the 139 paired contexts the
ratio to 2024 is 0.975 (CI 0.913 to
1.042) in Qwen3 and 1.015 (0.976 to
1.056) in BGE-M3. By first-version year
the OLMo gap is 0.215, 0.222, 0.224 and
0.221 for 2023 to 2026 in Qwen3 (0.184,
0.183, 0.190, 0.195 in BGE-M3). Within
contexts, a follow-up's step falls by
0.0041 per year (CI 0.0004 to 0.0079) in
Qwen3 and 0.0062 (0.0041 to 0.0083) in
BGE-M3 over 2023--2026: later citing
papers sit slightly farther out, in the
direction a lag would predict, but by
about a fortieth of the gap per year.
gpt-4o's gap against the 2023 papers is
also unchanged (0.147 against 0.153),
but its October 2023 cutoff overlaps the
window, so that comparison is
descriptive.

\textbf{Embedding check.} The
pre-registered runs embedded ideas and
follow-ups with a different runtime from
the corpus; on the 6,033 texts embedded
both ways the two agree to cosine
0.99999 or better. The robustness grid
uses the corpus runtime throughout and
reproduces the pre-registered Qwen3
numbers to the third decimal.

\subsection{Why a distance-matched baseline is not identified}\label{app:A.9}

A natural baseline for direction matches
every idea, at every round, with the
later arXiv paper of my corpus (2023
onwards, not a follow-up of the context;
with replacement) whose cosine to the
context is closest to the idea's step,
and takes \textbf{excess fit} = idea fit
minus matched paper fit. It is
identified only if, for most ideas, such
a paper exists at about the idea's
distance. It does not: the share of
ideas closer to their context than any
later paper of the pool is, at round 0 /
round 2, 0.69 / 0.40 (gpt-4o, Qwen3),
0.96 / 0.79 (gpt-4o, BGE-M3), 0.92 /
0.74 (OLMo, Qwen3) and 0.97 / 0.84
(OLMo, BGE-M3). For those ideas the
matched paper is the pool's closest
paper, which is further out, so excess
fit at round 0 is biased by the step
mismatch, and its change across rounds
partly measures the ideas moving into
support. I therefore report no
excess-fit result; the radial-invariant
measure of
\hyperref[sec:5.3]{Section~5.3} replaces
it. The distance-matched baseline for
the real follow-ups in
\hyperref[app:A.8]{A.8} is not affected,
because follow-ups are themselves later
papers and the match there is within
0.0005.

\section{Pre-registrations}\label{app:B}

Each document is in the repository
(\texttt{docs/\allowbreak{}prereg/\allowbreak{}},
except the two blocks of
\texttt{docs/\allowbreak{}design.md}),
was committed before the data it tests
were generated or inspected (except
\texttt{neutral\_\allowbreak{}revision.md},
see its row), and is identified in
\hyperref[tab:A12]{Table~A12} by the
commit that first added it; results were
added afterwards, in separate sections
of the same document or in result files.
Where a document was revised before its
test run, the revision has its own
commit and date. Commit timestamps are
set locally and the public repository is
a later anonymised copy, so the dates
below are not independently anchored.
Smoke calls run before a test run are
listed with what happened to their
outputs: most were run on targets
excluded from the test set or not
analysed, while the inspiration format
check and the neutral control's plumbing
test used the frozen protocol and are
kept in the data. Two groups are
registered follow-ups, written after the
original results were known, and are
labelled as such in the text: the
fifteen strengthening studies
(\texttt{strengthen\_\allowbreak{}*}),
written on 5 and 6 November 2026 (two
freeze analyses of saved outputs, two
simulations from the real design, and
eleven freeze new labels or
generations); and the thirteen review
studies
(\texttt{review\_\allowbreak{}*}): seven
committed together on 14 November 2026
before any of them was run, in answer to
a first review of the preprint by an AI
agent, and six committed on 15 November
2026, each before its own data were
generated or analysed, in answer to a
second one; their analysis scripts were
committed with the results, not with the
registrations. Every completed study is
listed.

\Needspace*{15\baselineskip}
\begingroup\footnotesize\setlength{\tabcolsep}{4pt}\setlength{\LTcapwidth}{\linewidth}
\begin{longtable}{@{}P{\dimexpr 0.240\linewidth-2\tabcolsep\relax}P{\dimexpr 0.180\linewidth-2\tabcolsep\relax}P{\dimexpr 0.270\linewidth-2\tabcolsep\relax}P{\dimexpr 0.230\linewidth-2\tabcolsep\relax}P{\dimexpr 0.080\linewidth-2\tabcolsep\relax}@{}}
\caption[]{\textbf{Table A12.} Pre-registration documents: the commit that first added each document, what it tests, the smoke calls run before its test run, and the sections that report it.}\label{tab:A12}\\
\toprule
Document & Commit, date & Tests & Smoke calls before the test run & Section \\
\midrule
\endfirsthead
\multicolumn{5}{@{}l}{\emph{Table A12, continued}}\\
\toprule
Document & Commit, date & Tests & Smoke calls before the test run & Section \\
\midrule
\endhead
\bottomrule
\endlastfoot
arXiv FAR ceiling in \texttt{design.md} & \texttt{2aab2ad}, 2026-10-24 & parameter-free corpus ceiling of far targets & none & \hyperref[sec:3]{3}, \hyperref[app:A.3]{\hyperref[app:A.3]{A.3}} \\
\texttt{memory\_\allowbreak{}probe.md} & \texttt{f4e61f6}, 2026-10-25 & anticipation by generators with known cutoffs & none & \hyperref[app:A.7]{\hyperref[app:A.7]{A.7}} \\
\texttt{coherent\_\allowbreak{}search.md} & \texttt{15eeead}, 2026-10-26 (rev. 2 the same day) & neighbour-free LLM search with score feedback & rev. 1 and rev. 2 smokes on target k = 0, excluded & \hyperref[sec:3]{3}, \hyperref[app:A.4]{\hyperref[app:A.4]{A.4}} \\
patents block of \texttt{design.md} & \texttt{05c997a}, 2026-10-26 & the gaps study on patents, including the FAR ceiling & none & \hyperref[sec:3]{3}, \hyperref[app:A.3]{\hyperref[app:A.3]{A.3}} \\
\texttt{fluent\_\allowbreak{}search.md} & \texttt{edfcb5c}, 2026-10-27; rev. 2 \texttt{c643773}, 2026-10-28 & gradient token search with a fluency penalty & 9 + 4 Kaggle runs on target k = 0, excluded & \hyperref[app:A.4]{\hyperref[app:A.4]{A.4}} \\
\texttt{fluent\_\allowbreak{}search.md}, rev. 3 & \texttt{3235967}, 2026-10-28 & seed-and-rewrite, tuned on REAL k = 11--20 & tuning targets disjoint from the test set & \hyperref[sec:3]{3}, \hyperref[app:A.4]{\hyperref[app:A.4]{A.4}} \\
\texttt{anchoring\_\allowbreak{}causal.md} & \texttt{38804b5}, 2026-10-28 & randomised swap of the shown neighbours & target k = 0, excluded & \hyperref[sec:3]{3}, \hyperref[app:A.3]{\hyperref[app:A.3]{A.3}} \\
\texttt{where\_\allowbreak{}future\_\allowbreak{}lands.md} & \texttt{8aec940}, 2026-10-30 & F1--F6: where the future lands, hole vs momentum & none & \hyperref[sec:4]{4}, \hyperref[app:A.2]{\hyperref[app:A.2]{A.2}} \\
\texttt{combination\_\allowbreak{}vs\_\allowbreak{}momentum.md} & \texttt{14a904b}, 2026-10-30 & combination signals vs momentum on topic triples (H1--H5) & timing run only & \hyperref[sec:4]{4}, \hyperref[app:A.6]{\hyperref[app:A.6]{A.6}} \\
\texttt{continue\_\allowbreak{}from\_\allowbreak{}context.md} & \texttt{127b678}, 2026-10-30 & how follow-ups move; one-shot continuation arms & none & \hyperref[sec:5.1]{5.1}, \hyperref[app:A.8]{\hyperref[app:A.8]{A.8}} \\
\texttt{iterative\_\allowbreak{}critique.md} & \texttt{9b77c3a}, 2026-10-30 & GEO, Reviewer and ORACLE critique (C1--C5) & none & \hyperref[sec:5.2]{5.2} \\
\texttt{modern\_\allowbreak{}and\_\allowbreak{}oracle\_\allowbreak{}history.md} & \texttt{b649b12}, 2026-11-01 & gpt-5.4-mini; ORACLE with history (M1--M4, H1--H2) & 9 ORACLE-H calls on context 0 and 3 gpt-5.4-mini calls on an unrelated prompt, outputs not analysed & \hyperref[sec:5.4]{5.4} \\
\texttt{inspiration\_\allowbreak{}from\_\allowbreak{}anything.md} & draft \texttt{5d79dee}, 2026-11-03; frozen \texttt{7ea8ae8}, 2026-11-03 & distance of inspiration material (P1--P5) & 6 format calls, kept in the run & \hyperref[sec:6]{6} \\
\texttt{neutral\_\allowbreak{}revision.md} & 2026-11-04; by my record written before the first call, but committed together with its results, so not independently timestamped & content-free revision control for the critique arms (N1--N6) & a 2-context plumbing test, kept in the data & \hyperref[sec:5.2]{5.2}, \hyperref[sec:5.3]{5.3} \\
\texttt{strengthen\_\allowbreak{}inspiration\_\allowbreak{}shape.md} & \texttt{f2bdd71}, 2026-11-05 & selection-aware bootstrap of the two-lines test; bound on a middle optimum & none (saved outputs) & \hyperref[sec:6]{6}, \hyperref[app:C.2]{C.2} \\
\texttt{strengthen\_\allowbreak{}cli\_\allowbreak{}powered.md} & \texttt{565e848}, 2026-11-05 & stress test at 5 ideas per call & none & \hyperref[app:C.3]{C.3} \\
\texttt{strengthen\_\allowbreak{}continuations.md} & \texttt{ae4f914}, 2026-11-05 & blind labelling of 400 + 331 follow-up pairs; step and direction against continuations & none & \hyperref[sec:5]{5}, \hyperref[sec:5.1]{5.1}, \hyperref[sec:8]{8} \\
\texttt{strengthen\_\allowbreak{}bge\_\allowbreak{}replication.md} & \texttt{a4041b6}, 2026-11-05 & BGE-M3 replication of causal anchoring, M1--M4, H1--H2, A1--A3, B2, B4, C2, C4 & none (re-embedding saved texts) & \hyperref[sec:3]{3}, \hyperref[sec:5.4]{5.4}, \hyperref[app:A.8]{\hyperref[app:A.8]{A.8}}, \hyperref[app:C.1]{C.1} \\
\texttt{strengthen\_\allowbreak{}continuations\_\allowbreak{}full.md} & \texttt{ff86a5b}, 2026-11-05 & blind labelling of the remaining 2,152 follow-up pairs; step and direction against all continuations & none & \hyperref[sec:5]{5}, \hyperref[sec:5.1]{5.1}, \hyperref[sec:8]{8} \\
\texttt{strengthen\_\allowbreak{}shape\_\allowbreak{}coverage.md} & \texttt{9211060}, 2026-11-05 & coverage simulation of the shape bound & none (simulation) & \hyperref[app:C.2]{C.2} \\
\texttt{strengthen\_\allowbreak{}cli\_\allowbreak{}callsize.md} & \texttt{7476b90}, 2026-11-05 & stress test with call size randomised within one run & none & \hyperref[app:C.3]{C.3} \\
\texttt{strengthen\_\allowbreak{}cli\_\allowbreak{}sessions.md} & \texttt{046a75e}, 2026-11-05; deviation \texttt{b444d49}, 2026-11-06 & call size and no-material baseline over eight sessions (S1, S2) & none & \hyperref[app:C.3]{C.3} \\
\texttt{strengthen\_\allowbreak{}human\_\allowbreak{}labels.md} & \texttt{cebcf48}, 2026-11-05 & blind author check of 40 LLM-labelled follow-up pairs & none & \hyperref[sec:5.1]{5.1}, \hyperref[sec:8]{8} \\
\texttt{strengthen\_\allowbreak{}shape\_\allowbreak{}band.md} & \texttt{011be07}, 2026-11-05 & simultaneous (sup-t) bound on a middle optimum, with coverage simulation & none (simulation; observed bound computed first) & \hyperref[sec:6]{6}, \hyperref[app:C.2]{C.2} \\
\texttt{strengthen\_\allowbreak{}olmo\_\allowbreak{}neutral.md} & \texttt{56d6a1e}, 2026-11-05 & content-free revision control for OLMo (O1--O6) & none & \hyperref[app:A.8]{\hyperref[app:A.8]{A.8}} \\
\texttt{strengthen\_\allowbreak{}modern\_\allowbreak{}neutral.md} & \texttt{67db500}, 2026-11-06 & content-free revision control for gpt-5.4-mini~(N1--N6) & none & \hyperref[sec:5.4]{5.4} \\
\texttt{strengthen\_\allowbreak{}anchoring\_\allowbreak{}none.md} & \texttt{67db500}, 2026-11-06 & past game under instruction against no material, 1,000 new ideas per arm (A1) & none & \hyperref[sec:6]{6} \\
\texttt{strengthen\_\allowbreak{}critique\_\allowbreak{}replication.md} & \texttt{67db500}, 2026-11-06~(17:01); clarification of the test rule \texttt{3eb74e5}, 2026-11-06~(19:17), after the three critique arms had been generated and during the last minute of the NEUTRAL arm, before any analysis & confirmatory replication of the gpt-4o critique study on 150 new contexts (R1, R2, E1--E6) & none & \hyperref[sec:5.2]{5.2}, \hyperref[sec:5.3]{5.3} \\
\texttt{strengthen\_\allowbreak{}olmo\_\allowbreak{}direction.md} & \texttt{25e7a0a}, 2026-11-05 & OLMo critique arms against content-free revision along the continuations' direction, 150 new contexts in two waves with a width-only stopping rule; 17 contexts are shared with the gpt-4o replication, not 0 as first recorded & label dry runs (batch counts only); wave 1's first kernel run died of GPU memory in round 1 and was rerun from scratch with the same inputs and seed, its outputs not analysed & \hyperref[sec:5.3]{5.3} \\
\texttt{review\_\allowbreak{}sample\_\allowbreak{}overlap.md} & \texttt{949d453}, 2026-11-14 & true overlap of the gpt-4o replication and the OLMo test (17 contexts); registered readings on the 133 non-overlapping contexts & none (saved outputs) & \hyperref[sec:5.3]{5.3} \\
\texttt{review\_\allowbreak{}olmo\_\allowbreak{}reliability.md} & \texttt{949d453}, 2026-11-14 & reliability and scale of the OLMo direction (Part A); Holm over the 12 detection tests and pooling of the gpt-4o replication (Part B, a post hoc correction of a registered test) & none (saved outputs) & \hyperref[sec:5.3]{5.3}, \hyperref[app:A.8]{\hyperref[app:A.8]{A.8}} \\
\texttt{review\_\allowbreak{}tangent\_\allowbreak{}confounds.md} & \texttt{949d453}, 2026-11-14 & gap fractions, step dose-response, genre direction, prompt echo, shared follow-ups, citation counts (Part A); rewrite and paraphrase nulls (Part B, run after Part A as registered; registered in its own part of the same document, not in the separate file it announced) & none (Part A on saved outputs; Part B 7,126 gpt-4o calls and an OLMo Kaggle run) & \hyperref[sec:5]{5}, \hyperref[sec:5.1]{5.1}, \hyperref[sec:5.3]{5.3}, \hyperref[app:A.8]{\hyperref[app:A.8]{A.8}} \\
\texttt{review\_\allowbreak{}posctrl.md} & \texttt{949d453}, 2026-11-14 & positive control revealing a held-out follow-up's title (HINT), fixed-target oracle (ORACLE-C), argmax switching, OLMo positive control & none recorded (1,800 gpt-4o calls; the first OLMo kernel stalled and was re-pushed unchanged) & \hyperref[sec:5.3]{5.3} \\
\texttt{review\_\allowbreak{}hole\_\allowbreak{}validity.md} & \texttt{949d453}, 2026-11-14; reproduction gate amended \texttt{329ee6c}--\texttt{277a52d}, 2026-11-14, before the tests & validity of the registered hole score; three alternative gap scores; 128--512 cells, centred and whitened space, minimum cell size, shrunken growth, 12- and 24-month horizons & none (saved embeddings) & \hyperref[sec:4]{4}, \hyperref[app:A.2]{\hyperref[app:A.2]{A.2}} \\
\texttt{review\_\allowbreak{}far\_\allowbreak{}decoding.md} & \texttt{949d453}, 2026-11-14 & far targets after centring and whitening (R1), independent target sets (R2), the score channel alone (R3), trained inversion in gtr-t5-base~(R4) & none recorded & \hyperref[sec:3]{3}, \hyperref[app:A.3]{\hyperref[app:A.3]{A.3}}, \hyperref[app:A.4]{\hyperref[app:A.4]{A.4}} \\
\texttt{review\_\allowbreak{}anticipation\_\allowbreak{}control.md} & \texttt{949d453}, 2026-11-14 & validation of the anticipation margin with held-out and rewritten future games; dirty future set; isolation range per source & none (391 gpt-4o calls; 10 rewrites blocked by the content filter and dropped) & \hyperref[sec:6]{6}, \hyperref[app:C.2]{C.2} \\
\texttt{review\_\allowbreak{}item\_\allowbreak{}density.md} & \texttt{75503dd}, 2026-11-15 & item-level isolation of new items against held-out past items (A1); coverage of the emptiest probe points (A2); no cells & none (saved embeddings) & \hyperref[sec:4]{4}, \hyperref[app:A.2]{\hyperref[app:A.2]{A.2}} \\
\texttt{review\_\allowbreak{}f5\_\allowbreak{}controls.md} & \texttt{4aac831}, 2026-11-15 & F5 with controls, review-count strata and symmetric tails & none (saved embeddings) & \hyperref[sec:4]{4} \\
\texttt{review\_\allowbreak{}olmo\_\allowbreak{}2023\_\allowbreak{}followups.md} & \texttt{52432d0}, 2026-11-15 & OLMo's step gap against citing papers first posted April--December 2023 & none (free APIs and a Kaggle embedding run) & \hyperref[sec:5.1]{5.1}, \hyperref[app:A.8]{\hyperref[app:A.8]{A.8}} \\
\texttt{review\_\allowbreak{}prompt\_\allowbreak{}sensitivity.md} & \texttt{f062ef1}, 2026-11-15 & the one-shot step under two other user-prompt wordings, discovery sample & 5 gpt-4o calls shared with \texttt{review\_\allowbreak{}rewrite\_\allowbreak{}nocontext.md}, excluded from every analysis & \hyperref[sec:5.1]{5.1}, \hyperref[sec:8]{8} \\
\texttt{review\_\allowbreak{}rewrite\_\allowbreak{}nocontext.md} & \texttt{e130f42}, 2026-11-15 & the short step against follow-ups rewritten as proposals without the context, discovery sample & the 5 smoke calls above & \hyperref[sec:5.1]{5.1}, \hyperref[sec:8]{8} \\
\texttt{review\_\allowbreak{}leakfree\_\allowbreak{}encoder.md} & \texttt{685219d}, 2026-11-15 & Section-4 tests and gap-score validity in a past-only TF-IDF/LSA space (three corpora) and in all-mpnet-base-v2 (Steam, arXiv) & none (a first mpnet run failed at start-up before embedding anything) & \hyperref[sec:4]{4}, \hyperref[sec:8]{8} \\
\end{longtable}
\endgroup

\section{Replication summary and checks of the inspiration study}\label{app:C}

\subsection{Replication across embedders}\label{app:C.1}

\hyperref[tab:A13]{Table~A13} says, for
each confirmatory result, whether the
second embedder was run and whether the
result held there.

\Needspace*{15\baselineskip}
\begingroup\footnotesize\setlength{\tabcolsep}{4pt}\setlength{\LTcapwidth}{\linewidth}
\begin{longtable}{@{}P{\dimexpr 0.330\linewidth-2\tabcolsep\relax}P{\dimexpr 0.090\linewidth-2\tabcolsep\relax}P{\dimexpr 0.580\linewidth-2\tabcolsep\relax}@{}}
\caption[]{\textbf{Table A13.} Replication of confirmatory results in BGE-M3.}\label{tab:A13}\\
\toprule
Result & Section & BGE-M3 \\
\midrule
\endfirsthead
\multicolumn{3}{@{}l}{\emph{Table A13, continued}}\\
\toprule
Result & Section & BGE-M3 \\
\midrule
\endhead
\bottomrule
\endlastfoot
Causal anchoring, 6 independent target sets & \hyperref[sec:3]{3} & replicated: donor attraction in all 12 decoder cells, also in centred and whitened spaces; target loss in the 4 REAL cells (not defined for GAP and FAR targets) \\
F1, nothing in the far region & \hyperref[sec:4]{4} & replicated \\
F2 fails: partial association with growth & \hyperref[sec:4]{4} & replicated (Steam, arXiv); patents not significant in either; with valid gap scores in 12 of 15 configurations \\
F3, hole adds nothing to momentum & \hyperref[sec:4]{4} & replicated in Steam and arXiv; patents/BGE-M3 +0.031, inconclusive \\
F4, latitude shift of the future & \hyperref[sec:4]{4} & not replicated \\
F5, isolation and quality & \hyperref[sec:4]{4} & replicated (fails in Steam, holds in patents) \\
F6, GAP targets not filled faster & \hyperref[sec:4]{4} & not run (inconclusive in Qwen3) \\
F5 with controls (registered follow-up) & \hyperref[sec:4]{4} & replicated: survives controls in Steam and patents in both embedders \\
Item-level isolation and coverage of empty points (registered follow-up) & \hyperref[sec:4]{4} & same verdict in both embedders: arXiv isolation shift supported, patents contradicted, Steam weak; preferential filling of empty points not shown in any domain (Steam and arXiv inconclusive, patents contradicting) \\
H5, hole adds nothing to fitted momentum & \hyperref[sec:4]{4}, \hyperref[app:A.6]{\hyperref[app:A.6]{A.6}} & replicated; H1--H4 do not use an embedder \\
A1--A3, how real follow-ups move & \hyperref[sec:5]{5}, \hyperref[app:A.8]{\hyperref[app:A.8]{A.8}} & replicated, including A2's failure \\
B1, the short step & \hyperref[sec:5.1]{5.1} & replicated, both follow-up sets, both generators, and against rewritten follow-ups \\
B3, memory raises diversity & \hyperref[app:A.8]{\hyperref[app:A.8]{A.8}} & replicated \\
B2, B4, shown neighbours and momentum hint & \hyperref[app:A.8]{\hyperref[app:A.8]{A.8}} & B2 holds in BGE-M3 (Holm \emph{p}~\textless{} 0.001), not in Qwen3; B4 fails in both, its reversal significant in Qwen3 only \\
C1, C3, the step lengthens under critique; C5 fails & \hyperref[sec:5.2]{5.2} & replicated (\hyperref[tab:A8]{Table~A8}) \\
C2, C4, hit changes & \hyperref[sec:5.2]{5.2}, \hyperref[app:A.8]{\hyperref[app:A.8]{A.8}} & replicated \\
N1, N2, critique lengthens the step beyond neutral revision & \hyperref[sec:5.2]{5.2} & replicated \\
N3--N6, critique against neutral revision & \hyperref[sec:5.2]{5.2} & N3: the reviewer's extra fit loss in both, GEO's in Qwen3 only; N4: GEO in Qwen3 only; N5: no significant contrast in either; N6: ORACLE negative in Qwen3 only \\
N1--N6 for gpt-5.4-mini (registered follow-up) & \hyperref[sec:5.4]{5.4} & N1, N2 replicated; N5: GEO above and ORACLE below NEUTRAL in both; N6: no contrast after Holm in either (GEO +0.016, uncorrected \emph{p}~0.02, in BGE-M3) \\
R1, R2, E1--E6, gpt-4o critique replication (registered follow-up) & \hyperref[sec:5.3]{5.3} & R1, R2 replicated; after Holm over 12, E1 detected in both, E2 in Qwen3 only, E3 within ±0.015 in BGE-M3 only, E4 about +0.011 in both and not significant, E5, E6 insufficient in both \\
OLMo critique against neutral revision along the continuations' direction (registered follow-up) & \hyperref[sec:5.3]{5.3} & GEO and reviewer within ±0.015 in both on the registered estimand; insufficient in Qwen3 after step adjustment, genre removal or dropping the 17 shared contexts; the positive control fails in both \\
Positive control, revealing a follow-up's title (registered follow-up) & \hyperref[sec:5.3]{5.3} & PC1 and PC3 replicated; PC2 (content beats distance feedback) in Qwen3 only \\
Short step against rewritten follow-ups and continuations (registered follow-up) & \hyperref[sec:5.1]{5.1} & replicated \\
M1--M4, H1--H2, 2026 generator and ORACLE-H & \hyperref[sec:5.4]{5.4} & M1--M3 replicated and M4 fails in both; H1 fails in Qwen3 but holds in BGE-M3~(+0.0045, Holm \emph{p}~0.027); H2's oracle nulls within ±0.015 in both \\
Short step against true continuations (registered follow-up) & \hyperref[sec:5.1]{5.1} & holds for gpt-4o in Qwen3 over all 47 contexts with a continuation (0.048), in BGE-M3~(0.086) and for OLMo in both \\
Short step with two other user-prompt wordings (registered follow-up) & \hyperref[sec:5.1]{5.1} & replicated (0.164 and 0.159, 97 \% and 95 \% of the original gap) \\
OLMo's short step against citing papers from 2023 (registered follow-up) & \hyperref[sec:5.1]{5.1}, \hyperref[app:A.8]{\hyperref[app:A.8]{A.8}} & replicated (0.184 against 0.183 for 2024) \\
Short step against follow-ups rewritten without the context (registered follow-up) & \hyperref[sec:5.1]{5.1} & replicated (0.167); the rewritten papers' step is unchanged in BGE-M3 but 0.082 farther out in Qwen3 \\
\hyperref[sec:4]{Section~4} in spaces without post-split encoder training (registered follow-up) & \hyperref[sec:4]{4} & not applicable: run in LSA (three corpora) and all-mpnet-base-v2 (Steam, arXiv) instead of the two main embedders; the registered conclusions held in both, with Steam inconclusive in LSA and patents weakly informative \\
P1, no positive distance slope & \hyperref[sec:6]{6} & fails in Qwen3, holds in BGE-M3 \\
P2, source over quintile & \hyperref[sec:6]{6} & fails with the frozen quintiles; holds with BGE-M3's own \\
P3, P4, manipulation check and uptake & \hyperref[sec:6]{6} & replicated \\
P5, no cell beats both baselines & \hyperref[sec:6]{6} & fails in Qwen3, holds in BGE-M3 \\
Past game under instruction worse than no material (registered follow-up) & \hyperref[sec:6]{6} & replicated (−0.0040, \emph{p}~= 9~×~10⁻⁸) \\
Margin validation with rewritten future games (registered follow-up) & \hyperref[sec:6]{6} & rewrites above no material in Qwen3 only (outcome F overall) \\
\end{longtable}
\endgroup

\subsection{Design and full results of the inspiration study, the margin's validation, and the shape bound}\label{app:C.2}

\textbf{Design.} I built a pool of
49,531 items from nine sources, dated
before 2024: past Steam games, board
games (BoardGameGeek), film plots,
Wikipedia articles describing a sport or
game and its rules, arXiv papers,
patents, news items (AG News; Zhang et
al., \hyperlink{ref-Zhang2015}{2015}),
recipes (Food.com; Majumder et al.,
\hyperlink{ref-Majumder2019}{2019}) and
random Wikipedia articles. A stimulus's
distance is its isolation from the Steam
past, 1 − its maximum cosine to any past
Steam game. Within each source I cut the
items into distance quintiles and drew
40 per source × quintile (1,800
stimuli). gpt-4o proposed one Steam game
per call under two framings of the same
stimulus: INCIDENTAL (``something you
happened to read today'') and INSTRUCTED
(the same, plus ``use it as
inspiration''). Two baselines need no
stimulus: 200 ideas with no material
(NONE), and 200 real 2023 games drawn
from the highest-momentum cells of the
past map (MOMENTUM,
\hyperref[sec:4]{Section~4}). The ground
truth is a stricter future than
\hyperref[sec:4]{Section~4}'s: the
32,322 Steam games released from 2024
with an appid above 2,680,000, a proxy
for a store page created after the
cutoff; the MOMENTUM games are past
items and are not in it. The proxy can
be checked only from release dates,
which bound page creation from above: no
game above the threshold in the future
set was released before 2024 and, apart
from two apparent date errors, no game
above it was released before 18 November
2023. I therefore call the set
contamination-reduced, not
contamination-free. The pool sampled the
whole pre-2024 Steam split rather than
stopping at the cutoff: 10 of the 200
Steam stimuli and 63 of the 200 MOMENTUM
games were released between October and
December 2023. An idea's anticipation is
its \textbf{margin}: its maximum cosine
to those games minus its maximum cosine
to a random sample of past games of the
same size (averaged over five samples),
so that a positive margin means it sits
nearer to what came next than to what
existed. Its \textbf{uptake} is its
cosine to its own stimulus minus its
mean cosine to five other stimuli of the
same source and quintile.

Before the test run (frozen at commit
\texttt{7ea8ae8}; a 6-call format check
came first and is kept in the data) I
pre-registered five predictions, tested
per framing with stimuli as the unit and
Holm over P1--P4: \textbf{P1}, the
within-source slope of margin on
distance rank is not positive, with the
competing shapes (down; up; an inverted
U, by the two-lines test of Simonsohn,
\hyperlink{ref-Simonsohn2018}{2018}) all
reported; \textbf{P2}, source explains
more variance in margin than distance
quintile; \textbf{P3}, isolation rises
with distance (the manipulation check);
\textbf{P4}, instruction raises uptake;
\textbf{P5}, no source × quintile cell
beats both baselines (Holm over 45
cells). P1 and P3 rank distance within
each embedder; P2 and P5 use the primary
space's quintiles in both, as frozen.
\hyperref[tab:10]{Table~10} gives the
outcomes.

Without the 10 late-2023 Steam stimuli
no outcome changes (Qwen3 P1 +0.0037 and
+0.0052, Holm \emph{p} 0.037 and 0.009).

\textbf{No sweet spot was detected.} The
pre-registered two-lines test finds no
inverted U in either embedder or framing
(\hyperref[fig:5]{Figure~5}). That is a
failure to detect, not a bound: its
breakpoint is chosen from the data, so
the segment intervals it yields are not
confidence bounds on every possible
middle optimum. A registered bootstrap
of the whole procedure (4,000 resamples
of stimuli within source, the breakpoint
re-selected in each), written after
these data had been seen, finds the fall
after the breakpoint compatible with
zero in every embedder and framing (95
\% intervals from −0.007 to +0.008),
with the breakpoint not identified. A
simultaneous sup-t bound on the peak of
a spline common to all sources, whose
coverage I checked by simulation from
the real design (below), puts any broad
middle optimum at most 0.005 to 0.010
margin units above the better end (0.25
to 0.40 of a per-idea standard deviation
with 4 df; 0.38 to 0.58 with 8 df, which
also covers humps at least 0.15 wide).
The shapes are not all flat: a quadratic
term is significantly negative in Qwen3
under instruction and in BGE-M3 under
the incidental framing, and in BGE-M3
the margin rises over the nearest fifth
of distance (+0.0068, CI +0.0030 to
+0.0107, incidental, the registered
exploratory check). The primary result
is that P1 fails: in Qwen3 a farther
stimulus gives a slightly higher margin,
significantly after Holm on its
one-sided test (+0.004 to +0.005 from
the nearest to the farthest stimulus of
a source), while in BGE-M3 the slope is
not distinguishable from zero. Rescored
against a reference set that matches the
held-out games' format (tagged games
only; exploratory) the slopes keep their
sign (Qwen3 +0.0027 and +0.0032, BGE-M3
+0.0007 and −0.0012) and lose
significance. In an exploratory
regression that adds the idea's own
isolation, the Qwen3 slope falls to
+0.0015 and +0.0027; isolation is
measured after generation and shares the
past-similarity term of the margin, so
this attenuation may be partly
mechanical.

The design also tests distance only
within narrow ranges: 7 of the 9 sources
span less than half of the pooled range
of isolation in each embedder
(\hyperref[tab:A14]{Table~A14}), the
nine ranges together cover it, and
between-source distance is confounded
with content, so nothing here speaks to
an optimum outside or across those
ranges. The choice of future set moves
the baseline: against the 8,667
post-2024 games excluded from the strict
set, whose store page may predate the
cutoff, ideas with no material score
0.007 to 0.009 higher than against
size-matched subsets of the strict set,
about twice the largest effect in
\hyperref[tab:10]{Table~10}, and the gap
is not explained by tag format; the
within-set contrasts use one set
throughout.

What distance reliably does is the
manipulation it is: farther material
yields more isolated ideas (P3), and
when the model is told to build on it,
less coherent ones. For coherence, a
blind judge (gpt-6.1-sol) rated a
stratified sample of 920 ideas (10 per
source × quintile × framing, and 20
without material) from 1 to 5, with
quadratic-weighted κ = 0.62 against a
second judge (Claude Opus 5.5) on 100 of
them; the within-source slope is −0.27
under instruction (\emph{p} = 0.014) and
nil under the incidental framing
(exploratory). Being told to use the
material does bring the idea closer to
it (P4).

\textbf{Content effects only partly
replicate.} In Qwen3, source explains
about ten times more variance in margin
than distance (P2) and six cells beat
both baselines (P5 fails: four quintiles
of recipes, which become cooking and
crafting games, a genre the post-cutoff
catalogue grew into). In BGE-M3, P5
holds under every binning, and P2 fails
under the frozen Qwen3 quintiles (Holm
\emph{p} 0.11 and 0.17) but holds with
BGE-M3's own quintiles of the drawn
stimuli (0.046 and 0.009), so the source
effect is there in both spaces while the
recipe cells are not
(\hyperref[fig:5]{Figure~5}). The
content effect that recurs, on this
margin, is anchoring under instruction.
Told to build on a past game, gpt-4o
gives the lowest margin of any source in
Qwen3 (−0.005, 0.013 below the other
sources, Mann-Whitney \emph{p} = 2 ×
10⁻⁸) and the second lowest in BGE-M3,
while its uptake doubles from 0.11 under
the incidental framing to 0.23. I
pre-registered that one comparison with
new ideas: 1,000 past-game stimuli not
used before, under instruction, against
1,000 calls with no material. On this
margin a past game is worse than no
material, by 0.0052 (CI 0.0032 to
0.0073; one-sided Welch \emph{p} = 6 ×
10⁻⁷) in Qwen3 and by 0.0040 (CI 0.0025
to 0.0055; \emph{p} = 9 × 10⁻⁸) in
BGE-M3, mostly through the past games
nearest the rest of the catalogue. It is
a within-measure contrast: both arms are
ideas written by the same model in the
same format, which is what the
validation above leaves the margin able
to compare.

\textbf{Positive control} (registered
follow-up,
\texttt{review\_\allowbreak{}anticipation\_\allowbreak{}control.md}).
200 held-out post-cutoff games (as
listed) are scored against the rest of
the future set, against 200 ideas with
no material, 200 MOMENTUM games and 200
real 2023 games. Margins (Qwen3 /
BGE-M3): held-out future games +0.0214 /
+0.0159, no material +0.0001 / −0.0006,
MOMENTUM −0.0005 / −0.0010, 2023 games
−0.0073 / −0.0037. Held-out minus no
material: +0.0213 (CI 0.0135 to 0.0296;
signal-to-noise 0.81 per-idea SD) and
+0.0165 (CI 0.0103 to 0.0226; 1.02);
over 100 random splits, means 0.0190 and
0.0131. Restricted to tagged games, so
that format is matched on that feature:
+0.0175 and +0.0095. Rewritten by gpt-4o
in the idea format, with 191 future and
199 past rewrites after the content
filter (9 + 1 dropped as registered):
rewritten future games minus no material
+0.0088 (CI 0.0037 to 0.0139,
signal-to-noise 0.34, Holm \emph{p} =
0.0016) and +0.0024 (CI −0.0011 to
+0.0060, 0.15, Holm \emph{p} = 0.094);
rewritten future minus rewritten 2023
games +0.0093 (Holm 0.0038) and +0.0033
(Holm 0.086); the paired cost of
rewriting −0.010 and −0.013. By the
registered labels Qwen3 gives W
(rewrites above no material after Holm,
with low signal-to-noise) and BGE-M3
gives F (as-is games above no material,
rewrites not), and the weaker embedder
governs: F. The broad-optimum bound is
1.08 (Qwen3) and 2.67 (BGE-M3) times the
rewritten games' margin, so the nulls
are uninformative relative to the
measure. \emph{Deviation and
sensitivity:} the ten blocked rewrites
were also blocked when the same gpt-4o
was reached through OpenRouter's Azure
route and were then written by the Codex
CLI, a different model; with those ten
in place both embedders give W (BGE-M3
+0.0032, Holm \emph{p} = 0.040), but the
nine future ones average a margin of
0.0206 in BGE-M3, about eight times the
0.0026 of all 200 future rewrites and
11.6 times the 0.0018 of the 191
retained ones, so that run is not a
neutral substitute. \emph{Future set:}
against the 8,667 post-2024 games
excluded from the strict set,
no-material ideas score 0.0069 (Qwen3)
and 0.0095 (BGE-M3) higher than against
size-matched subsets of the strict set;
the gap does not shrink when tag format
is matched (reading ``not format''), and
the excluded games were released earlier
(median 2024-10-02 against 2025-04-18),
consistent with their being nearer the
cutoff and partly visible before it.

\begin{table}[tbp]
\centering\footnotesize\setlength{\tabcolsep}{4pt}
\caption[]{\textbf{Table A14.} Range of stimulus isolation per source (5th to 95th percentile of the 200 drawn stimuli) and its ratio to the pooled range of all 1,800 stimuli (Qwen3 0.245 to 0.652; BGE-M3 0.190 to 0.538). Sources with a ratio below 0.5 are range-restricted.}\label{tab:A14}
\begin{tabular}{@{}P{\dimexpr 0.260\linewidth-2\tabcolsep\relax}P{\dimexpr 0.240\linewidth-2\tabcolsep\relax}P{\dimexpr 0.130\linewidth-2\tabcolsep\relax}P{\dimexpr 0.240\linewidth-2\tabcolsep\relax}P{\dimexpr 0.130\linewidth-2\tabcolsep\relax}@{}}
\toprule
Source & Qwen3 range & Ratio & BGE-M3 range & Ratio \\
\midrule
Steam games & 0.129~to~0.372 & 0.60 & 0.097~to~0.276 & 0.51 \\
Board games & 0.260~to~0.484 & 0.55 & 0.235~to~0.381 & 0.42 \\
Film plots & 0.441~to~0.583 & 0.35 & 0.352~to~0.460 & 0.31 \\
Sports rules & 0.282~to~0.482 & 0.49 & 0.250~to~0.430 & 0.52 \\
arXiv papers & 0.464~to~0.667 & 0.50 & 0.370~to~0.495 & 0.36 \\
Patents & 0.469~to~0.660 & 0.47 & 0.379~to~0.502 & 0.35 \\
News & 0.500~to~0.659 & 0.39 & 0.418~to~0.565 & 0.42 \\
Recipes & 0.530~to~0.670 & 0.34 & 0.410~to~0.491 & 0.23 \\
Random Wikipedia & 0.491~to~0.662 & 0.42 & 0.443~to~0.574 & 0.38 \\
\bottomrule
\end{tabular}
\end{table}

The arXiv ratio in Qwen3 rounds to 0.50
and is below 0.5 before rounding. In raw
isolation units the exploratory P1 slope
is +0.0022 (CI 0.0002 to 0.0042) and
+0.0027 (CI 0.0006 to 0.0048) per 0.1 in
Qwen3 (incidental, instructed) and
+0.0018 (CI 0.0001 to 0.0036) and
−0.00001 in BGE-M3.

\textbf{The shape bound} (registered
follow-ups, written after the
\hyperref[sec:6]{Section~6} data had
been seen). A spline common to all
sources (4 df), maximised over distance
ranks 0.1 to 0.9, gives the size of a
middle peak above the better end. Its
nominal bootstrap percentile is not a
calibrated bound: in a registered
simulation from the real design it
covered the peak of the fitted spline
itself in as few as 83 \% of datasets. I
replaced it by a simultaneous bound: a
one-sided sup-t band over the 162
differences between the curve at each
grid rank and at either end, projected
onto the peak. In 1,000 simulated
datasets per scenario and cell (14
scenarios, 4 cells) this bound covers
the peak of the fitted 4-df spline in at
least 98 \% of datasets in every
scenario, and the true peak of flat,
linear, broad and source-varying curves
in at least 99 \%. It does not cover
narrow humps (width 0.15 at half
maximum), which a 4-df curve flattens
(coverage of their true peak 12 to 97
\%). With 8 df the same bound covers the
fitted and the true peak of every
simulated hump other than the spike (all
at least 0.15 wide) in at least 95 \% of
datasets, the registered condition for
quoting it for such humps; no variant
bounds a spike of width 0.05. Observed,
the bound is 0.005 to 0.010 margin units
above the better end with 4 df (0.25 to
0.40 of a per-idea standard deviation)
and 0.009 to 0.010 with 8 df (0.38 to
0.58). Under the simulated shapes and
noise model, a broad middle optimum
larger than the first, or a hump of the
simulated kinds at least 0.15 wide
larger than the second, is excluded at
95 \%. The bootstrap's lower percentile,
read as a test for a peak, is positive
in 2 to 4 \% of datasets with no peak,
and for a simulated broad hump of half a
per-idea standard deviation in every
dataset; observed, it is below zero in
all four cells. The interval for the
fall at the selected breakpoint is
conservative in every simulated scenario
(coverage at least 0.978), though with a
narrow hump it does not bound the fall
at other breakpoints. In BGE-M3 the
margin rises over the nearest fifth of
distance under both framings (+0.0068,
CI +0.0030 to +0.0107, incidental;
+0.0052, CI +0.0015 to +0.0091,
instructed, examined in addition), while
its change beyond it is compatible with
zero but not shown to be small (+0.0004,
CI −0.003 to +0.004; −0.0014, CI −0.005
to +0.002).

\textbf{Anchoring details.} In the first
run a past game under instruction was
not significantly worse than no material
(Holm over the nine sources, \emph{p} ≥
0.47), which is why the comparison was
registered with new ideas. In the
confirmatory run the effect is
concentrated in the past games nearest
the rest of the catalogue (Qwen3: −0.006
to −0.009 in quintiles 1 to 3, −0.001 in
quintile 5). Shown a past game
incidentally in the first run, the model
does no worse than with other material
(fourth worst of nine in Qwen3, best in
BGE-M3). In BGE-M3 the instructed past
game is 0.002 below the other sources
(\emph{p} = 0.04).

\subsection{Stress test with later generators}\label{app:C.3}

Claude Opus 5.5 and gpt-6.1-sol, run as
command-line agents told to answer
without tools (Claude with its file,
shell and web tools disallowed; Codex in
a read-only sandbox in an empty
directory, with no trace kept to confirm
abstention), each generated 450
instructed ideas (10 stimuli per cell)
and 100 with no material. Both were
trained after the future games appeared,
so I read them only against their own
no-material arm. A first run wrote 25
ideas per call, which left two calls per
source and no call-level test able to
reach significance (a past game scored
−0.010 for Claude and −0.008 for
gpt-6.1-sol below no material).

A registered second run on the same
stimuli with 5 ideas per call (90
stimulus calls and 20 no-material calls
per generator, in random order; all 110
completed in the first pass) shrinks a
past game's gap to no material to
−0.0035 for both generators, not
detected (exact Mann-Whitney on call
means, \emph{p} = 0.29 for gpt-6.1-sol
and 0.31 for Claude, Holm 0.57 for both;
95 \% cluster-bootstrap intervals −0.011
to +0.004 and −0.010 to +0.003); a past
game still ranks among the three worst
of the nine sources for both. The
distance slopes stay null with 90 call
clusters (gpt-6.1-sol +0.0035, CI
−0.0025 to +0.0095; Claude −0.0020, CI
−0.0086 to +0.0046), but distance does
not make these generators' ideas
detectably more isolated either (P3 does
not replicate in either run), so the
null slopes say little about distance as
such.

The no-material baseline moved between
the runs (Claude's from −0.0003 to
−0.0062), so a registered third run
randomised call size within one run: 400
no-material ideas and all 200 past-game
stimuli, each at 5 and at 25 per call
(144 calls per generator in one random
order, all completed at the first
attempt). Call size does not move the
baseline detectably (25 minus 5 per
call, cluster-bootstrap CIs: gpt-6.1-sol
−0.0028, CI −0.0062 to +0.0008; Claude
+0.0020, CI −0.0014 to +0.0054; CR1 t
test, Holm 0.25 for both), nor the past
game's gap (interaction +0.0039 and
−0.0023, Holm 0.65). For Claude the
interval excludes the earlier shift's
point estimate (+0.0058), but that
shift, from 4 and 20 calls, is itself
uncertain (CI +0.0008 to +0.0110), and a
joint call bootstrap puts the current
effect minus the earlier shift at
−0.0039 (CI −0.0100 to +0.0022): call
size is neither detected nor excluded as
its cause. gpt-6.1-sol's baseline at 5
per call moved by 0.006 between the two
runs that used that size.

A fourth registered run repeated the
no-material arm at both sizes in eight
sessions planned two hours apart (per
generator and session, 16 calls of 5 and
4 of 25 in random order; an overnight
connection failure lost three sessions,
re-run at the end as recorded before
analysis, but two of them started about
a minute after the previous session's
end rather than two hours later).
Averaged over sessions, 25 per call
lowers gpt-6.1-sol's baseline by 0.005
(cluster-bootstrap CI −0.007 to −0.003,
CR1 Holm \emph{p} \textless{} 0.001),
with no detected difference in that
effect between sessions; for Claude it
is not detected (+0.0013, CI −0.0012 to
+0.0038, Holm 0.32). Call size does not
explain gpt-6.1-sol's move at a fixed
size, and for Claude the session effect
minus the earlier shift is −0.0046
(joint call bootstrap, implemented after
S1 to apply the registered reading rule;
CI −0.0104 to +0.0011), so call size is
again neither detected nor excluded as
the cause of that shift. No difference
between the baseline's session means
beyond call-level noise was detected
(joint Wald \emph{p} = 0.27 and 0.77,
Holm 0.55 and 0.77; standard deviation
across sessions 0.0030, CI 0.0022 to
0.0054, and 0.0019, CI 0.0015 to
0.0048).

Of the earlier baselines, gpt-6.1-sol's
second run at 5 per call (−0.0002) lies
below its eight session means at that
size (0.0018 to 0.0102) and Claude's
first run at 25 per call (−0.0003) above
its eight at that size (−0.0083 to
−0.0031); the others lie within their
ranges (descriptive). The interaction
also changes the mix of stimuli in a
call (one quintile per call at 5, all
five at 25), so it tests that batching
protocol rather than call size alone.

In the third run a past game again
scores below no material (secondary, not
confirmatory): for gpt-6.1-sol at both
sizes (−0.011, CI −0.016 to −0.006, at 5
per call; −0.007, CI −0.013 to −0.002,
at 25) and for Claude at 25 per call
(−0.005, CI −0.009 to −0.001) but not at
5 (−0.0025, CI −0.007 to +0.002). In the
second run Claude's recipe ideas
anticipate better than its no-material
ideas (+0.016, Holm \emph{p} = 0.003
over the nine sources), which the first
run did not show.

\end{document}